%% file: tabdata_ieee.tex
\documentclass[lettersize,journal]{IEEEtran}
\usepackage[dvipsnames]{xcolor}
\usepackage{amsmath,amsfonts}
\usepackage{algorithmic}
\usepackage{algorithm}
\usepackage{array,bm}
\usepackage[caption=false,font=normalsize,labelfont=sf,textfont=sf]{subfig}
\usepackage{textcomp}
\usepackage{stfloats}
\usepackage{url}
\usepackage{verbatim}
\usepackage{graphicx}
\usepackage{tabularx,booktabs,multirow}
\usepackage{amsfonts}       % blackboard math symbols
\usepackage{nicefrac}       % compact symbols for 1/2, etc.
\usepackage{microtype}      % microtypography
\usepackage{xcolor}         % colors
\usepackage{svg}            % include svgs
\usepackage{caption}        % set caption font
\usepackage{soul}
\usepackage{dsfont}         % double bar letters
\usepackage{makecell}       % line breaks in cells
\usepackage{tcolorbox}
\usepackage{cite}
\usepackage{tikz}
\usepackage{adjustbox}
\newcommand{\ieeeversion}[0]{}
\newcommand{\citep}[1]{\cite{#1}}
\newcommand{\added}[1]{\textcolor{black}{#1}}
\newcommand{\revised}[1]{\textcolor{black}{#1}}
\usepackage{hyperref}

\hypersetup{
colorlinks = true,
linkcolor = ForestGreen,
anchorcolor = blue,
citecolor = Blue,
filecolor = cyan,
menucolor = ForestGreen,
runcolor = cyan,
urlcolor = ForestGreen}

\begin{document}

\title{Deep Neural Networks and Tabular Data: A Survey}
%~\IEEEmembership{Member,~IEEE,}
\author{Vadim Borisov, Tobias Leemann, Kathrin Seßler, Johannes Haug,\\ Martin Pawelczyk and Gjergji Kasneci% <-this % stops a space
\thanks{All authors are with the Data Science and Analytics Research (DSAR) group at the University of Tübingen, 72070 Tübingen, Germany. Gjergji Kasneci is also affiliated with Schufa Holding AG, 65201 Wiesbaden, Germany. \\
Corresponding authors: \\
\href{mailto:vadim.borisov@uni-tuebingen.de}{vadim.borisov@uni-tuebingen.de}  \\
\href{mailto:tobias.leemann@uni-tuebingen.de}{tobias.leemann@uni-tuebingen.de}  \\
© 2022 IEEE. Personal use of this material is permitted. Permission from IEEE must be obtained for all other uses, in any current or future media, including reprinting/republishing this material for advertising or promotional purposes, creating new collective works, for resale or redistribution to servers or lists, or reuse of any copyrighted component of this work in other works.
}% <-this % stops a space
%\thanks{Manuscript received April 19, 2021; revised August 16, 2021.}
}

\markboth{Submitted to the IEEE, %~Vol.~0, No.~0,
June~2022}%
{Shell \MakeLowercase{\textit{et al.}}: A Sample Article Using IEEEtran.cls for IEEE Journals}


\maketitle

\begin{abstract}
Heterogeneous tabular data are the most commonly used form of data and are essential for numerous critical and computationally demanding applications.
On homogeneous data sets, deep neural networks have repeatedly shown excellent performance and have therefore been widely adopted. 
However, their \added{adaptation} to tabular data for inference or data generation tasks remains highly challenging.
\added{To facilitate further progress in the field}, this work provides an overview of state-of-the-art deep learning methods for tabular data.
We categorize \added{these methods} into three groups: data transformations, specialized architectures, and regularization models.
\added{For each of these groups,} our work offers a comprehensive overview of the main approaches. 
Moreover, we discuss deep learning approaches for generating tabular data, and we also provide an overview over strategies for explaining deep models on tabular data. 
Thus, our first contribution is to address the main research streams and existing methodologies in the mentioned areas, while highlighting relevant challenges and open research questions. 
\added{Our second contribution} is to provide an empirical comparison of traditional machine learning methods with \added{eleven} deep learning approaches across \added{five} popular real-world tabular data sets of different sizes and with different learning objectives. 
Our results, which we have made \href{https://github.com/kathrinse/TabSurvey}{publicly available} as competitive benchmarks, indicate that algorithms based on gradient-boosted tree ensembles still mostly outperform deep learning models \added{on supervised learning tasks, suggesting that the research progress on competitive deep learning models for tabular data is stagnating}.
To the best of our knowledge, this is the first in-depth overview of deep learning approaches for tabular data; as such, this
work can serve as a valuable starting point to guide researchers and practitioners interested in deep learning with tabular data. 
\end{abstract}

\begin{IEEEkeywords}
Deep neural networks, Tabular data, Heterogeneous data, Discrete data, Tabular data generation, Probabilistic modeling, Interpretability, Benchmark, Survey 
\end{IEEEkeywords}

\input{content.tex}

\bibliographystyle{IEEEtran}
\bibliography{cas-refs}

% \newpage

 

% \begin{IEEEbiography}[{\includegraphics[width=1in,height=1.25in,clip,keepaspectratio]{images/fig1}}]{Michael Shell}
% \end{IEEEbiography}



\vfill

\end{document}

%% file: content.tex
\section{Introduction}
\label{sec:intro}

\revised{Ever-increasing computational resources and the availability of large, labelled data sets have accelerated the success of deep neural networks\cite{schmidhuber2015deep, deeplearningbook}. In particular, architectures based on convolutions, recurrent mechanisms \cite{hochreiter1997long}, or transformers \cite{vaswani2017attention} have led to unprecedented performance in a multitude of domains. Although deep learning methods perform outstandingly well for classification or data generation tasks on homogeneous data (e.g., image, audio, and text data), tabular data still pose a challenge to deep learning models~\cite{arik2019tabnet, popov2019neural, Shwartz-Ziv2021, elsayed2021we}.
Tabular data -- in contrast to image or language data -- are heterogeneous, leading to dense numerical and sparse categorical features. Furthermore, the correlation among the features is weaker than the one introduced through spatial or semantic relationships in image or speech data.
Hence, it is necessary to discover and exploit relations without relying on spatial information~\cite{somepalli2021saint}. 
Therefore, Kadra et~al.\ called  tabular data sets the last \textit{“unconquered castle”} for deep neural network models~\cite{Kadra2021reg}.}

%Variables can be correlated or independent, and the features have no positional information.  
%We discuss these and more reason in more details in Section \ref{sec:discussion}.

Heterogeneous data are the most commonly used form of data~\citep{Shwartz-Ziv2021}, and it is ubiquitous in many crucial applications, such as medical diagnosis based on patient history~\citep{ulmer2020trust, somani2021deep, borisov2021robust}, predictive analytics for financial applications (e.g., risk analysis, estimation of creditworthiness, the recommendation of investment strategies, and portfolio management)~\citep{clements2020sequential}, click-through rate (CTR) prediction~\citep{guo2017deepfm}, user recommendation systems~\citep{zhang2019deep}, customer churn prediction~\citep{ahmed2017survey, tang2020customer}, cybersecurity~\citep{buczak2015survey}, fraud detection~\citep{Cartella2021}, identity protection~\citep{liu2021machine}, psychology~\citep{urban2021deep}, delay estimations~\citep{shoman2020deep}, anomaly detection~\citep{pang2021deep}, and so forth. In all these applications, a boost in predictive performance and robustness may have considerable benefits for both end users and companies that provide such solutions. Simultaneously, this requires handling many data-related pitfalls, such as noise, impreciseness, different attribute types and value ranges, or \revised{the missing value problem} and privacy issues.

Meanwhile, deep neural networks offer multiple advantages over traditional machine learning methods.
First, these methods are highly flexible \citep{sahoo2017online}, allow for efficient and iterative training, and are particularly valuable for AutoML~\citep{he2021automl, artzi2021classification, shi2021multimodal, fakoor2020fastdad, gijsbers2019open, yin2020tabert}.
Second, tabular data generation is possible using deep neural networks and can, for instance, help mitigate class imbalance problems~\citep{wang2019generative}. 
Third, neural networks can be deployed for multimodal learning problems where tabular data can be one of many input modalities~\citep{baltruvsaitis2018multimodal,lichtenwalter2021deep, shi2021multimodal, polsterl2021combining, soares2021predicting}, for tabular data distillation~\citep{medvedev2020new, li2020tnt}, for federated learning \citep{roschewitz2021ifedavg}, and in many more scenarios. 

\added{Successful deployments of data-driven applications require solving several tasks, among which we identified three \textit{core challenges}: (1) \textit{inference} (2) \textit{data generation}, and (3) \textit{interpretability}. The most crucial task is inference which is concerned with making predictions based on past observations. While a powerful predictive model is critical for all the applications mentioned in the previous paragraph, the interplay between tabular data and deep neural networks goes beyond simple inference tasks. Before a predictive model can even be trained, the training data usually needs to be preprocessed. This is where data generation plays a crucial role, as one of the standard deployment steps involves the imputation of missing values~\citep{sanchez2020improving, gondara2018mida, camino2020working} and the rebalancing of the data set ~\citep{engelmann2021conditional, darabi2021synthesising} (i.e., equalizing sample sizes for different classes). Furthermore, it might be simply impossible to use the actual data due to privacy concerns, e.g., in financial or medical applications~\cite{kamthe2021copula,Choi2017medGAN}. Thus, to tackle the data preprocessing and privacy challenges, probabilistic tabular data \textit{generation} is essential.}
\added{Finally, with stricter data protection laws such as  California Consumer Privacy Act (CCPA) \cite{ccpa2021} and the European General Data Protection Regulation (EU GDPR)  \cite{regulation2016gdpr}, which both mandate a right to explanations for automated decision systems (e.g., in the form or recourse \cite{voigt2017eu}), \textit{interpretability} is becoming a key aspect for predictive models used for tabular data \cite{sahakyan2021explainable,GRISCI2021111}. During deployment, interpretability methods also serve as a valuable tool for model debugging and auditing \cite{bhatt2020explainable}}.

\added{Evidently, apart from the the core challenges of inference, generation, and interpretability, there are several other important subfields, such as working with data streams, distribution shifts, as well as privacy and fairness considerations that should not be neglected. Nevertheless, to navigate the vast body of literature, we focus on the identified core problems and thoroughly review the state of the art in this work. We will briefly discuss the remaining topics at the end of this survey.}

\added{Beyond reviewing current literature, we think that an exhaustive comparison between existing deep learning approaches for heterogeneous tabular data is necessary to put reported results into context. The variety of benchmarking data sets and the different setups often prevent comparison of results across papers. Additionally, important aspects of deep learning models, such as training and inference time, model size, and interpretability, are usually not discussed. We aim to bridge this gap by providing a comparison of the surveyed inference approaches with classical -- yet very strong -- baselines such as XGBoost~\cite{XGBOOST}. We open-source our code, allowing researchers to reproduce and extend our findings.}

%As the data collection step, especially for heterogeneous data, is costly and time-consuming, there are many approaches for synthetic tabular data generation. However, modelling the probability distribution of rows in tabular data and generating realistic synthetic data are challenging since heterogeneous tabular data typically includes a mix of discrete and continuous variables. Continuous variables may have multiple modes, whereas discrete columns are often imbalanced. All these pitfalls in combination with missing, noisy, or unbounded values make the issue of tabular data generation quite complex, even for modern deep generative architectures.

% interpretable tab data and neural network 
%Another important aspect is the interpretation of deep neural networks on tabular data~\citep{GRISCI2021111}. Many popular approaches for the interpretation of deep neural networks stem from the computer vision domain, where a pixel group is highlighted, creating a so-called saliency map. Nevertheless, for tabular data sets, highlighting a variable relationship is also essential. Many existing methods -- especially those based on attention mechanisms \citep{vaswani2017attention} -- offer a highlighting of relationships by design and their attention maps can be easily visualized.
In summary, the aims of this survey are to provide:
\begin{enumerate}
\item a thorough review of existing scientific literature on deep learning for tabular data;
\item a taxonomic categorization of the available approaches for classification and regression tasks on heterogeneous tabular data;
\item a presentation of the state of the art and promising paths towards tabular data generation;
\item an overview of existing explanation approaches for deep models for tabular data;
\item an extensive empirical comparison of traditional machine learning methods and deep learning models on multiple real-world heterogeneous tabular data sets;
\item a discussion on the main reasons for the limited success of deep learning on tabular data;
\item a list of open challenges related to deep learning for tabular data.
\end{enumerate}

%Thus, data science practitioners and researchers will be able to quickly identify promising starting points and guidance for the use cases or research questions at hand. 

%The remainder of the survey is organized as follows:

Accordingly, this survey is structured as follows:
We discuss related works in Section~\ref{sec:related_work}. 
To introduce the reader to the field, in Section~\ref{sec:maintabulardata}, we provide definitions of the key terms, a brief outline of the domain's history, and propose a unified taxonomy of current approaches to deep learning with tabular data.
Section~\ref{sec:neuralnetworks_for_tabdata} covers the main methods for modelling tabular data using deep neural networks.
Section~\ref{sec:tab_data_generation} presents an overview on tabular data generation using deep neural networks. 
An overview of explanation mechanisms for deep models for tabular data is presented in Section~\ref{sec:explanation}. 
In Section~\ref{sec:experiments_main}, we provide an extensive empirical comparison of machine and deep learning methods on real-world data, that also involves model size, runtime, and interpretability.
In Section~\ref{sec:discussion}, we summarize the state of the field and give future perspectives.
Finally, we outline several open research questions before concluding in Section~\ref{sec:conclusion}.

{
  \hypersetup{linkcolor=black,  linktoc=all}
  \tableofcontents
}

%\textit{To help improve the survey, please do not hesitate to send corrections and suggestions to corresponding authors. }

%%%%%%%%%%%%%%%%%%%%%%%%%%%%%
%%%%%%%%%%%%%%%%%%%%%%%%%%%%%
%%%%%%%%%%%%%%%%%%%%%%%%%%%%%
%%%%%%%%%%%%%%%%%%%%%%%%%%%%%

\section{Related Work}
\label{sec:related_work}

To the best of our knowledge, there is no study dedicated exclusively to the application of deep neural networks to tabular data, spanning the areas of supervised and unsupervised learning, data synthesis, \revised{and interpretability.} 
Prior works cover some of these aspects, but none of them systematically discusses the existing approaches in the broadness of this survey. 
%We also could not find any work that reviews state-of-the-art approaches for synthesizing tabular data using deep neural networks.

However, there are some works that cover parts of the domain. 
There is a comprehensive analysis of common approaches for categorical data encoding as a preprocessing step for deep neural networks by Hancock \& Khoshgoftaar \cite{hancock2020survey}. 
The authors compared existing methods for categorical data encoding on various tabular data sets and different deep learning architectures. 
We also discuss the key categorical data encoding methods in Section \ref{sec:data_encoding}.

A recent survey by Sahakyan et al. \cite{sahakyan2021explainable} summarizes explanation techniques in the context of tabular data. 
Hence, we do not provide a detailed discussion of explainable machine learning for tabular data in this paper. 
However, for the sake of completeness, we present some of the most relevant works in Section \ref{sec:explanation} and highlight open challenges in this area.

Gorishniy et al. \cite{gorishniy2021revisiting} empirically evaluated a large number of state-of-the-art deep learning approaches for tabular data on a wide range of data sets. 
The authors demonstrated that a tuned deep neural network model with a ResNet-like architecture~\citep{he2016deep} shows comparable performance to some state-of-the-art deep learning approaches for tabular data. 

Recently, Shwartz-Ziv \& Armon \cite{Shwartz-Ziv2021} published a study on several different deep models for tabular data including TabNet~\citep{arik2019tabnet}, NODE~\citep{popov2019neural}, Net-DNF~\citep{katzir2021netdnf}.
Additionally, they compared deep learning approaches to gradient boosting decision tree algorithms regarding accuracy, training effort, inference efficiency, and hyperparameter optimization time. 
They observed that deep models had the best results on their chosen data sets, however, not one single deep model could outperform all the others in general.
The deep models were challenged by gradient boosting decision trees, leading the authors to conclude that efficient tabular data modelling using deep neural networks is still an open research problem. %With our survey, we aim to provide the background necessary for future work on this question. 
\added{In the face of this evidence, we aim to integrate the necessary background for future research on the inference problem and on the intertwined challenges of generation and explainability into a single work.}

% A quantitative study by \citet{Gupta2021} analyzed the robustness of neural networks also considering different state-of-the-art regularization techniques. 
% To fool a prediction model, tabular data can be corrupted, and adversarial examples can be produced. 
% Using this data, it is possible to mislead fraud detection models as well as humans as shown by~\citet{Cartella2021}. 
% Before using deep learning models for tabular data in a critical environment, one should be aware of the possible susceptibility to attacks.

\begin{figure}
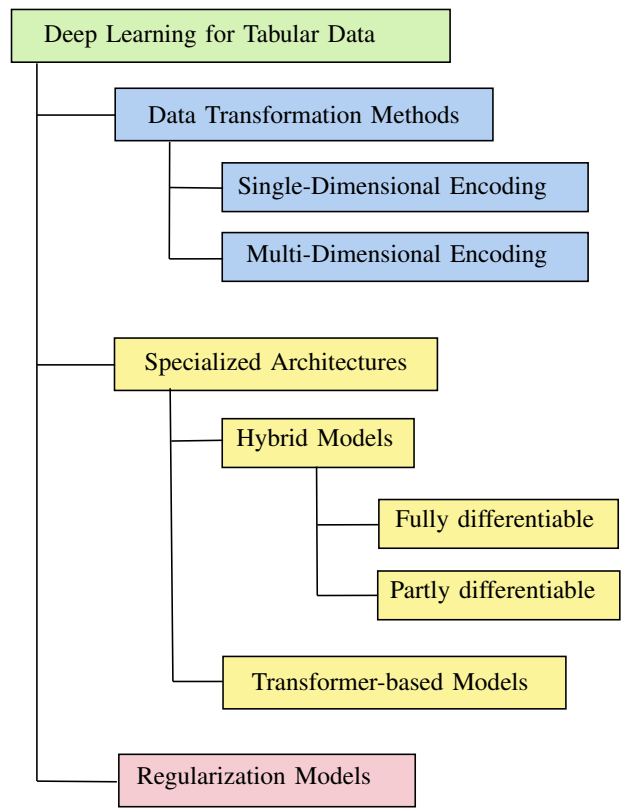

    \centering
    \include{images/taxonomy}
    \caption{Unified taxonomy of deep neural network models for heterogeneous tabular data.
    }
    \label{fig:overview}
\end{figure}

%%%%%%%%%%%%%%%%%%%%%%%%%%%%%
%%%%%%%%%%%%%%%%%%%%%%%%%%%%%

\section{Tabular Data and Deep Neural Networks}
\label{sec:maintabulardata}

\subsection{Definitions} % Terms defined: Deep Neural Network, heterogeneous data / homogeneous
\label{sec:definitions}
In this section, we give definitions for central terms used in this work. 
We also provide pointers to the original works for more detailed explanations of the methods. 

The key concept in this survey is a \textit{(deep) neural network}. 
Unless stated otherwise we use this concept as a synonym for feed-forward networks, as described by~\cite{deeplearningbook}, and name the concrete model whenever we deviate from this concept. 
A deep neural network deﬁnes a mapping $\hat{f}$,
\begin{equation}
    y = f(x) \approx \hat{f}(x; W),
\end{equation}
that learns the value of the model parameters $W$ (i.e., the ``weights'' of a neural network) that results in the best approximation of the true underlying and unknown function $f$. \added{In this case, $x$ is a multi-dimensional data sample (i.e., $x\in \mathbb{R}^n$) with corresponding target $y$ (where typically, $y\in \mathbb{R}^k$ for $k$ classes and $y\in \mathbb{R}$ for regression tasks) from a data set of tuples $\left\{\left(x_i, y_i\right)\right\}_{i\in\mathcal{I}}$.}
The network is called feed-forward if the input information flows in one direction to the output without any feedback connections.

Throughout this survey we focus on \textit{heterogeneous data} which usually contains a variety of attribute types --
These include both continuous and discrete attributes of different type (e.g., binary values, ordinal values, high-cardinality categorical values).
This is fundamentally different from \textit{homogeneous} data modalities, such as images, audio, or text data where only a single feature type is present.

\added{\textit{Categorical variables} are an attribute type of particular importance.} According to Lane's definition~\cite{statistics2003Lane}, categorical variables are qualitative values. 
They ``do not imply a numerical ordering", unlike quantitative values, which are ``measured in terms of numbers''. Usually, a categorical variable can take one out of a limited set of values.
Examples of typical categorical variables include \emph{gender}, \emph{user\_id}, \emph{product\_type} and \emph{topic}.

\textit{Tabular data}, \added{sometimes also called \textit{structured data} \cite{ryan2020deep}}, is a subcategory of the heterogeneous data format that is usually presented in a table~\citep{cvitkovic2020deep} with data points as rows and features as columns. 
% Hence, for the scope of this work, we can define a tabular data set as a table which contains $n$ numerical columns $\{N_1, \ldots, N_n\}$ and $c$ categorical columns $\{C_1, \ldots, C_c\}$. 
% All columns are random variables following a joint distribution $\mathbb{P}(N_{1:n}, C_{1:c})$. 
\added{In summary, for the scope of this work, we refer to a data set with a fixed number of features that are either continuous or categorical as tabular.} 
Each data point can be understood as a row in the table, or -- taking a probabilistic view -- as a sample from the unknown joint distribution. 
\added{An illustrative example of five rows of a heterogeneous, tabular data is provided in Table \ref{tab:tabular_example}}.

%This data format is widely used in practical machine learning and data science. 

\begin{table}[]
\centering
{
% Setting the horizontal and vertical space
\renewcommand{\arraystretch}{0.8}
\setlength{\tabcolsep}{8pt}
    \begin{tcolorbox}[boxsep=-1mm]
    \begin{tabular}{rcccccc@{}}  %{m{4cm}cl} %{@{}l@{}r@{}r@{}r@{}r@{}r@{}}}
        \toprule
        Age & Education & Occupation & Sex & Income \\
        \midrule
        
        39 & Bachelors & Adm-clerical  & Male &  $\le$50K\\
        50 & Bachelors & Exec-managerial & Male & $>$50K\\
        38 & HS-grad & Handlers-cleaners & Male &  $\le$50K\\
        53 & 11th & Handlers-cleaners	 & Male &  $\le$50K\\
        28 & Bachelors & Prof-specialty	& Female &  $>$50K\\
        \bottomrule
    \end{tabular}
    \end{tcolorbox}
    }
    
    \caption{\added{An example of a heterogeneous tabular data set. Here we show five samples with selected variables from the Adult data set~\cite{Dua:2019}. Section \ref{sec:dataset} provides further details on this data set.}}
    \label{tab:tabular_example}
    
\end{table}
%Lastly, for the data generation section we have to define the most popular neural network architectures called generative adversarial network (GAN)~\citep{goodfellow2014generative} and variational autoencoder (VAE)~\citep{Kingma2014VAE}.

\subsection{A Brief History of Deep Learning on Tabular Data}
\revised{Tabular data are one of the oldest forms of data to be statistically analysed. Before digital collection of text, images, and sound was possible, almost all data were tabular \added{~\cite{miles1881sunstroke, fisher1936use, jdanov2019human}}}. \revised{Therefore, it was the target of early machine learning research \cite{fix1951discriminatory}}. However, deep neural networks became popular in the digital age and were further developed with a focus on homogeneous data. In recent years, various supervised, self-supervised, and semi-supervised deep learning approaches have been proposed that explicitly address the issue of tabular data modelling again. Early works mostly focused on data transformation techniques for preprocessing~\citep{giles1992learning, horne1995experimental, willenborg1996statistical}, which are still important today \citep{hancock2020survey}.

A huge stimulus was the rise of e-commerce, which demanded novel solutions, especially in advertising \citep{richardson2007predicting, guo2017deepfm}.
These tasks required fast and accurate estimation on heterogeneous data sets with many categorical variables, for which the traditional machine learning approaches are not well suited (e.g., categorical features that have high cardinality can lead to very sparse high-dimensional feature vectors and non-robust models).
As a result, researchers and data scientists started looking for more flexible solutions, e.g., those based on deep neural networks, that can capture complex non-linear dependencies in the data. 

In particular, the click-through rate prediction problem has received a lot of attention~\citep{guo2017deepfm, ke2019deepgbm, wang2021masknet}. 
A large variety of approaches were proposed, most of them relying on specialized neural network architectures for heterogeneous tabular data.
%The most important methods for click-through rate estimation are included in our survey.

% The click-through rate (CTR) estimation problem became a massive stimulus for the development of the field of deep learning on tabular data since the majority of CTR data sets (CITE) are tabular data sets with multiple categorical variables (e.g., city, gender, preferences). Additionally, these data sets are large, complex, and hard to use in an online fashion. Therefore, researchers and data scientists have started looking for a way to utilize deep learning models. Since DNNs offer multiple advantages over decision tree-based algorithms - scalability and batch learning (no need to keep the whole data set in memory, which is the case of most decision tree-based machine learning algorithms). 

A more recent line of research, sparked by Shavitt \& Segal \cite{shavitt2018regularization}, evolved based on the idea that regularization may improve the performance of deep neural networks on tabular data \citep{Kadra2021reg}. 
This has led to an intensification of research on regularization approaches.

% In the recent years, attention-based approaches such as transformers dominate the research. It could be explained by tremendous success of BERT \citep{devlin2018bert} and GPT-3 \citep{brown2020language} models on text data, and visual transformers (ViTs) on visual data . 

Due to the tremendous success of attention-based approaches such as transformers on textual \citep{brown2020language} and visual data \citep{dosovitskiy2021visiontransformer, khan2021transformers}, researchers have recently also started applying attention-based methods and self-supervised learning techniques to tabular data. 
After the introduction of transformer architectures to the field of tabular data \citep{arik2019tabnet}, a lot of research effort has focused on transformer architectures that can be successfully applied to very large tabular data sets.

% \subsection{data sets}

% However, it is hard to track the research of in XX century in this area since the term \textit{tabular data}, in a way how we understand it nowadays, it used since the last decade.

% Before the "deep learning era" most of the data set were tabular (according to our definition), because the data collection/gathering and saving were extremely expensive tasks. 

% Iris flower data set collected in 1936 by sir Ronald Fisher, is an example of the tabular data, however there are no categorical variables. A more recent example and more coplex data set is the German Credit data data set collected in 1994 by Hans Hofmann. Interestinly, it was "corrected" in the recent years \cite{groemping2019south} 

% However, it is hard to track the research of the XX century in this area since the term \textit{tabular data}, in a way how we understand it nowadays, it used since the last decade. 

% The most significant computational disadvantage of GBDTs is the need to store (almost) the entire data set in memory, therefore researches and data scientists ... 

% Nowadays, we see many approaches utilize the attention-based architecture for tabular learning using DNNs... 

\subsection{Challenges of Learning With Tabular Data}
As we have mentioned in Section \ref{sec:related_work}, deep neural networks %are usually inferior to 
\added{often perform less favourably compared to} more traditional machine learning methods (e.g., tree-based methods) when dealing with tabular data. 
However, it is often unclear why deep learning cannot achieve the same level of predictive quality as in other domains such as image classification and natural language processing. 
In the following, we identify and discuss four possible reasons:
\begin{enumerate}%[noitemsep]
    \item \added{\textbf{Low-Quality Training Data}:} 
    %\textbf{Inappropriate Training Data:} 
    Data quality is a common issue with real-world tabular data sets. They often include missing values~\citep{sanchez2020improving}, extreme data (outliers)~\citep{pang2021deep}, erroneous or inconsistent data \citep{karr2006data}, and have small overall size relative to the high-dimensional feature vectors generated from the data \citep{Xu2018TGAN}. Also, due to the expensive nature of data collection, tabular data are frequently class-imbalanced. \added{ These challenges affect all machine learning algorithms; however, most of the modern decision tree-based algorithms can handle missing values or different/extreme variable ranges internally by looking for appropriate  approximations and split values \cite{XGBOOST, ke2017lightgbm, prokhorenkova2018catboost}}.%, and in comparison to deep neural networks these methods less sensitive to inconsistent data.}
    
    \item \textbf{Missing or Complex Irregular Spatial Dependencies:} There is often no spatial correlation between the variables in tabular data sets \citep{Zhu2021igtd}, or the dependencies between features are rather complex and irregular. When working with tabular data, the structure and relationships between its features have to be learned from scratch. Thus, the \textit{inductive biases} used in popular models for homogeneous data, such as convolutional neural networks, are unsuitable for modelling this data type \citep{katzir2021netdnf, rahaman2019spectral, mitchell2017spatial}.  %Therefore, different architectures may be required, which is investigated in \ref{sec:archi_based}.
    
    \item \revised{\textbf{Dependency on Preprocessing:} A key advantage of deep learning on homogeneous data is that it includes an implicit representation learning step \cite{GoodfellowDLBook2016}, so only a minimal amount of preprocessing or explicit feature construction is required. However, for tabular data and deep neural networks the performance may strongly depend on the selected preprocessing strategy \cite{gorishniy2022embeddings}. Handling the categorical features remains particularly challenging \citep{hancock2020survey} and can easily lead to a very sparse feature matrix (e.g., by using a one-hot encoding scheme) or introduce a synthetic ordering of previously unordered values (e.g., by using an ordinal encoding scheme). \added{Lastly, preprocessing methods for deep neural networks may lead to information loss, leading to a reduction in predictive performance~\citep{fitkov2012evaluating}.}}
    %However, as categorical features may be very sparse (a problem known as \textit{curse of dimensionality}), }
    %\Citet{hancock2020survey} have analyzed different embedding techniques for categorical variables. 
    %Dealing with categorical features is also one of the main aspects we discuss in Section \ref{sec:neuralnetworks_for_tabdata}.   %subdividing them into three broad categories: deterministic, algorithmic, and automatic. Some of the methods discussed in this paper can handle the categorical data (only transformed into a numerical representation); others include a more advanced embedding to address this challenge. One possibility to encode the data are using the hashing trick. \citep{Kang2020DHE} describe different ways to use hashing, also in combination with a learnable encoding using neural networks.
    %Applications that work with homogeneous data have effectively used data augmentation \citep{perez2017effectiveness}, transfer learning \citep{tan2018survey} and test-time augmentation \citep{shanmugam2020and}. For heterogeneous tabular data, these techniques are often difficult to apply. However, some frameworks for learning with tabular data, such as VIME \citep{yoon2020vime} and SAINT \citep{somepalli2021saint}, use data augmentation strategies in the embedding space.

    \item \added{\textbf{Importance of Single Features}: While typically changing the class of an image requires a coordinated change in many features, i.e., pixels, the smallest possible change of a categorical (or binary) feature can entirely flip a prediction on tabular data \cite{shavitt2018regularization}.}
    %This is usually less problematic for homogeneous (continuous) data sets.}
    %Deep neural networks can be extremely fragile to tiny perturbations of the input data \citep{szegedy2013intriguing, levy2020not}. 
    \revised{In contrast to deep neural networks, decision-tree algorithms can handle varying feature importance exceptionally well by selecting a single feature and appropriate threshold (i.e., splitting) values  and ``ignoring" the rest of the data sample. Shavitt \& Segal \cite{shavitt2018regularization} have argued that individual weight regularization may mitigate this challenge and motivated more work in this direction \cite{Kadra2021reg}.} %As a consequence of their extremeensitivity, artificial neural network models have high curvature decision boundaries \citep{poole2016exponential, achille2019information}, whereas decision tree-based algorithms can learn hyperplane-like boundaries. In order to reduce the input-related sensitivity of deep neural networks, some approaches propose to apply strong regularization to learning parameters \citep{shavitt2018regularization, Kadra2021reg}. We discuss these methods in Section \ref{sec:reg_models} in more detail. 
    %Finally, deep neural networks usually have an excessive number of hyperparameters \citep{deeplearningbook}, where traditional machine learning approaches on tabular data (e.g., decision-tree-based models, support-vector machines \citep{cortes1995support}) typically have significantly fewer hyperparameters. As a consequence, the tuning time for deep learning models is vastly higher than that of decision tree-based approaches. In addition, neural networks tend to be sensitive to the choice of hyperparameters (e.g. learning rate, number of hidden layers or activation function). In general, fewer hyperparameters are preferable to reduce the risk of non-robust predictions.
    % \item Asymmetrical manifold is required for categorical data (Heterogeneous manifold / non-euclidean space) 
\end{enumerate}
\added{With these four fundamental challenges in mind, we continue by organizing and discussing the strategies developed to address them. We start by developing a suitable taxonomy.}

\subsection{Unified Taxonomy}
In this section, we introduce a taxonomy of approaches that allows for a unified view of the field. 
%We observed that the works we encountered while preparing this survey fall into three main categories:
We divide the works from the deep learning with tabular data literature into three main categories:
\textit{data transformation methods}, \textit{specialized architectures}, and \textit{regularization models}. 
In Fig.~\ref{fig:overview}, we provide an overview of our taxonomy of deep learning methods for tabular data.

\textbf{Data transformation methods.} The methods in the first group transform categorical and numerical data.
This is usually done to enable deep neural network models to better extract the information signal. 
Methods from this group do not require new architectures or adaptations of the existing data processing pipeline. 
Nevertheless, the transformation step comes at the cost of an increased preprocessing time. 
This might be an issue for high-load systems~\citep{baylor2017tfx}, particularly in the presence of categorical variables with high cardinality and growing data set size. 
We can further subdivide this area into \textit{Single-Dimensional Encodings} and \textit{Multi-Dimensional Encodings}.
The former encodings are employed to transform each feature independently while the latter encoding methods map an entire record to another representation.

\textbf{Specialized architectures.} The biggest share of works investigates specialized architectures and suggests that a different deep neural network architecture is required for tabular data.
Two types of architectures are of particular importance: \textit{hybrid models} fuse classical machine learning approaches (e.g., decision trees) with neural networks, while \textit{transformer-based models} rely on attention mechanisms.

\textbf{Regularization models.} Lastly, the group of regularization models claims that one of the main reasons for the moderate performance of deep learning models on tabular data is their extreme non-linearity and model complexity.
Therefore, strong regularization schemes are proposed as a solution.
They are mainly implemented in the form of special-purpose loss functions.

We believe our taxonomy may help practitioners find the methods of choice that can be easily integrated into their existing tool chain. 
For instance, applying data transformations can result in performance improvements while maintaining the current model architecture.
Conversely, using specialized architectures, the data preprocessing pipeline can be kept intact.

%  - How we selected the works 
%  - How we came out with the current taxonomy 
%  - differences 

%%%%%%%%%%%%%%%%%%%%%%%%%%%%%
%%%%%%%%%%%%%%%%%%%%%%%%%%%%%

%%%%%%%%%%%%%%%%%%%%%%%%%%%%%%%%%%%%%%%%%%%%%%%%
%%%%%%%%%%%%%%%%%%%%%%%%%%%%%%%%%%%%%%%%%%%%%%%%
%%%%%%%%%%%%%%%%%%%%%%%%%%%%%%%%%%%%%%%%%%%%%%%%
%%%%%%%%%%%%%%%%%%%%%%%%%%%%%%%%%%%%%%%%%%%%%%%%
%%%%%%%%%%%%%%%%%%%%%%%%%%%%%%%%%%%%%%%%%%%%%%%%
%%%%%%%%%%%%%%%%%%%%%%%%%%%%%%%%%%%%%%%%%%%%%%%%
%%%%%%%%%%%%%%%%%%%%%%%%%%%%%%%%%%%%%%%%%%%%%%%%
\section{Deep Neural Networks for Tabular Data}
\label{sec:neuralnetworks_for_tabdata}

% In the last years many papers considering this problem were published, all trying to find a deep learning algorithm which outperforms the decision tree-based algorithms.

In this section, we discuss the use of deep neural networks on tabular data for classification and regression tasks according to the taxonomy presented in the previous section.
We provide an overview of existing deep learning approaches in this area of research in Table~\ref{tab:overview_methods} and examine the three methodological categories in detail: data transformation methods (see Subsection \ref{sec:data_transformation}), architecture-based methods (see Subsection \ref{sec:archi_based}), and regularization-based models (see Subsection \ref{sec:reg_models}).

\setlength{\tabcolsep}{10pt}
\renewcommand{\arraystretch}{1.5}
\newcommand{\wref}[2]{#1 #2}
\begin{table*}[!h] 
\centering
\begin{tabularx}{0.9\textwidth}{m{0.5cm} ccp{8cm}}
\toprule
    & \makecell{Method} & \revised{Interpretability} & Key Characteristics\\
    \midrule
    \multirow{4}{*}{\rotatebox[origin=c]{90}{Encoding~~}} & \wref{SuperTML} {\cite{sun2019supertml}} &  & Transform tabular data into images for CNNs \\
    
    & \wref{VIME}{\cite{yoon2020vime}} &  & Self-supervised learning and contextual embedding\\
    
    & \wref{IGTD}{\cite{Zhu2021igtd}} &  & Transform tabular data into images for CNNs\\
    
    & \wref{SCARF}{\cite{bahri2021scarf}} & & Self-supervised contrastive learning  \\
    \midrule
    
    \multirow{12}{*}{\rotatebox[origin=c]{90}{Architectures, Hybrid~~~~}} & \wref{Wide\&Deep}{\cite{cheng2016wide}}  &  & Embedding layer for categorical features \\
    
    & \wref{DeepFM}{\cite{guo2017deepfm}} &  & Factorization machine for categorical data\\

    & \wref{SDT}{\cite{frosst2017distilling}} & \checkmark & Distill neural network into interpretable decision tree\\
        
    & \wref{xDeepFM}{\cite{lian2018xdeepfm}} & &  Compressed interaction network \\

    & \wref{TabNN}{\cite{Ke2019tabnn}} & & DNNs based on feature groups distilled from GBDT \\

    & \wref{DeepGBM}{\cite{ke2019deepgbm}} & & Two DNNs, distill knowlegde from decision tree \\
    
    & \wref{NODE}{\cite{popov2019neural}} &  & Differentiable oblivious decision trees ensemble  \\

    & \wref{NON}{\cite{luo2020NON}} &  & Network-on-network model \\

    & \wref{DNN2LR}{\cite{liu2020dnn2lr}} & & Calculate cross feature wields with DNNs for LR \\ 

    & \wref{Net-DNF}{\cite{katzir2021netdnf}} & & Structure based on disjunctive normal form \\

    & \wref{Boost-GNN}{\cite{Ivanov2021bgnn}} & & GNN on top decision trees from the GBDT algorithm \\
    
    & \wref{SDTR}{\cite{Luo2021sdtr}} & & Hierarchical differentiable neural regression model \\
    \midrule
    
    \multirow{4}{*}{\rotatebox[origin=c]{90}{\makecell{Architectures,~~~~~\\Transformer~~~~~}}} & \wref{TabNet}{\cite{arik2019tabnet}} & \checkmark & Sequential attention structure \\ 
    
    & \wref{TabTransformer}{\cite{Huang2020tabtrans}} & \checkmark & Transformer network for categorical data\\
    
    & \wref{SAINT}{\cite{somepalli2021saint}} &\checkmark & Attention over both rows and columns \\

    & \wref{ARM-Net}{\cite{cai2021arm}} & &  Adaptive relational modelling with multi-headgated attention network \\
    
    & \wref{Non-Param. Transformer}{\cite{kossen2021selfattention}} & & Process the entire dataset at once, use attention between data points\\
    \midrule
    
    \multirow{2}{*}{\rotatebox[origin=c]{90}{Regul.}}& \wref{RLN}{\cite{shavitt2018regularization}}& \checkmark & Hyperparameters regularization scheme\\
    
    & \wref{Regularized DNNs}{\cite{Kadra2021reg}} & & A "cocktail" of regularization techniques\\
    \bottomrule
    \end{tabularx}
    \caption{Overview of deep learning approaches for tabular data. We organize them in categories ordered chronologically inside the groups. The ``Interpretability'' column indicates whether the approach offers some form interpretability for the model's decisions. The key characteristics of every model are summarized \added{in the last column}.
    %We report the discussed methods in the chronological order.
    }
    \label{tab:overview_methods}
\end{table*}

\subsection{Data Transformation Methods}
\label{sec:data_transformation}

%In this subsection, we cover methods following the data transformation approach. 

Most traditional approaches for deep neural networks on tabular data fall into this group. Interestingly, data preprocessing plays a relatively minor role in computer vision, even though the field is currently dominated by deep learning solutions~\citep{deeplearningbook}. 
There are many different possibilities to transform tabular data, and each may have a different impact on the learning results~\citep{hancock2020survey}.

% It can be done by multiple ways, the \ref{sec:data_encoding} 
% sophisticated preprocessing 

%%%%%%%%%%%%%%%%%%%%%%%%%%%%%%%%%%%%%%%%%%%%%%%%
\subsubsection{Single-Dimensional Encoding}
\label{sec:data_encoding}

One of the critical obstacles for deep learning with tabular data are categorical variables. 
Since neural networks only accept real number vectors as inputs, these values must be transformed before a model can use them. 
Therefore, the first class of methods attempts to encode categorical variables in a way suitable for deep learning models. 

Approaches in this group~\cite{hancock2020survey} are divided into \textit{deterministic} techniques, which can be used before training the model, and more complicated \textit{automatic} techniques that are part of the model architecture.
There are many ways for deterministic data encoding; hence we restrict ourselves to the most common ones without the claim of completeness.

The simplest data encoding technique might be ordinal or label encoding. 
Every category is just mapped to a discrete numeric value, e.g., $\left\{  \texttt{Apple},~\texttt{Banana} \right\}$ are encoded as $\left\{ 0, 1\right\}$.
One drawback of this method may be that it introduces an artificial order to previously unordered categories.
Another straightforward method that does not induce any order is the one-hot encoding.
One additional column for each unique category is added to the data. 
Only the column corresponding to the observed category is assigned the value one, with the other values being zero. 
In our example, \texttt{Apple} could be encoded as \texttt{(1,0)} and \texttt{Banana} as \texttt{(0,1)}. 
In the presence of a diverse set of categories in the data, this method can lead to high-dimensional sparse feature vectors and exacerbate the ``curse of dimensionality'' problem.

Binary encoding limits the number of new columns by transforming the qualitative data into a numerical representation (as the label encoding does) and using the binary format of the number. 
Again the digits, are split into different columns, but there are only $\log(c)$ new columns if $c$ is the number of unique categorical values. 
If we extend our example to three fruits, e.g., \{\texttt{Apple}, \texttt{Banana}, \texttt{Pear}\}, we only need two columns to represent them: \texttt{(01)}, \texttt{(10)}, \texttt{(11)}.

One approach that needs no extra columns and does not include any artificial order is the so-called leave-one-out encoding. 
It is based on the target encoding technique proposed in the work by~\cite{MicciBarreca2001}, where every category is replaced with the mean of the target variable of that category. 
The leave-one-out encoding excludes the current row when computing the mean of the target variable to avoid overfitting. 
This approach is also used in the CatBoost framework \citep{prokhorenkova2018catboost}, a state-of-the-art machine learning library for heterogeneous tabular data based on the gradient boosting algorithm \citep{friedman2002stochastic}. 

A different strategy is hash-based encoding. 
Every category is transformed into a fixed-size value via a deterministic hash function. 
The output size is not directly dependent on the number of input categories but can be chosen manually.

% One approach was introduced by~\citep{yoon2020vime}. In contrast to the other models, they do not use a supervised model but propose a self-and semi-supervised deep learning framework for tabular data called \textbf{VIME}. Considering that there is often much more unlabelled than label data, they use this in the first step to train an encoder self-supervised by using two pretext tasks, mask vector and feature vector estimation. The structure of the self-supervised learning part can be found in figure \ref{fig:vime_self}. They solve these tasks; the encoder learns how to construct an informative homogeneous representation of the raw input data. The encoder and the pretext tasks are no complicated architectures but simple multi-layer perceptrons. A predictive model (also an MLP) is trained in the semi-supervised step, using the labelled and unlabelled data transformed by the encoder. For the latter, a novel data augmentation method is used, corrupting one (unlabelled) data point multiple times with different masks. On the predictions from all augmented samples from one original data point, a consistency loss $\mathcal{L}_u$ can be computed. Combined with a supervised loss $\mathcal{L}_s$ from the labelled data, the predictive model minimizes the final loss $\mathcal{L}_{final} = \mathcal{L}_s + \beta \cdot \mathcal{L}_u$. VIME exceeds other state-of-the-art self-, semi- or supervised baselines.

%%%%%%%%%%%%%%%%%%%%%%%%%%%%%%%%%%%%%%%%%%%%%%%%
\subsubsection{Multi-Dimensional Encoding}
\label{sec:data_covnersion}

A first automatic encoding strategy is the VIME approach~\citep{yoon2020vime}.
The authors propose a self- and semi-supervised deep learning framework for tabular data that trains an encoder in a self-supervised fashion by using two pretext tasks. 
Those tasks that are independent from the concrete downstream task which the predictor has to solve. 
The first task of VIME is called mask vector estimation; its goal is to determine which values in a sample are corrupted. The second task, i.e., feature vector estimation, is to recover the original values of the sample. The encoder itself is a simple multilayer perceptron. 
This automatic encoding makes use of the fact that there is often much more unlabelled than labelled data. The encoder learns how to construct an informative homogeneous representation of the raw input data. 
In the semi-supervised step, a predictive model, which is also a deep neural network model, is trained using the labelled and unlabelled data transformed by the encoder.
For the encoder, a novel data augmentation method is used, corrupting an unlabelled data point multiple times with different masks. 
On the predictions from all augmented samples from one original data point, a consistency loss $\mathcal{L}_u$ can be computed that rewards similar outputs.
Combined with a supervised loss $\mathcal{L}_s$ from the labelled data, the predictive model minimizes the final loss $\mathcal{L} = \mathcal{L}_s + \beta \cdot \mathcal{L}_u.$
To summarize, the VIME network trains an encoder, which is responsible to transform the categorical and numerical features into a new homogeneous and informative representation. This transformed feature vector is used as an input to the predictive model. 
For the encoder itself, the categorical data can be transformed by a simple one-hot-encoding and binary encoding.
%Instead of encoding the data to apply deep neural networks, the values can be converted into a different format to make use of other deep neural networks.

Another stream of research aims at transforming the tabular input into a more homogeneous format.
Since the revival of deep learning, convolutional neural networks have shown tremendous success in computer vision tasks. Therefore, the work by~\cite{sun2019supertml} proposed the SuperTML method, which is a data conversion technique to transform tabular data into an image data format (2-d matrices), i.e., black-and-white images.

The image generator for tabular data (IGTD) by~\cite{Zhu2021igtd} follows an idea similar to SuperTML. 
The IGTD framework converts tabular data into images to make use of classical convolutional architectures. 
As convolutional neural networks rely on spatial dependencies, the transformation into images is optimized by minimizing the difference between the feature distance ranking of the tabular data and the pixel distance ranking of the generated image.
Every feature corresponds to one pixel, which leads to compact images with similar features close at neighbouring pixels. Thus, IGDTs can be used in the absence of domain knowledge.
The authors show relatively solid results for data with strong feature relationships but the method may fail if the features are independent or feature similarities can not characterize the relationships.
In their experiments, the authors used only gene expression profiles and molecular descriptors of drugs as data. 
This kind of data may lead to a favourable inductive bias, so the general viability of the approach remains unclear.

% \begin{figure}[t]
%     \centering
%     \includegraphics[width=\columnwidth]{images/methods/NODE.png}
%     \caption{Structure of one Oblivious Decision Tree (ODT) inside the NODE layer~\citep{popov2019neural}. The NODE architecture contains multiple connected layers, each one with several ODTs.}
%     \label{fig:NODE}
% \end{figure}

\subsection{Specialized Architectures}
\label{sec:archi_based}

Specialized architectures form the largest group of approaches for deep tabular data learning. Hence, in this group, the focus is on the development and investigation of novel deep neural network architectures designed specifically for heterogeneous tabular data. Guided by the types of available models, we divide this group into two sub-groups: Hybrid models (presented in \ref{sec:hybrid_models}) and transformer-based models (discussed in \ref{sec:transformer_based}).

%%%%%%%%%%%%%%%%%%%%%%%%%%%%%%%%%%%%%%%%%%%%%%%%
\subsubsection{Hybrid Models}
\label{sec:hybrid_models}

Most approaches for deep neural networks on tabular data are hybrid models. They transform the data and fuse successful classical machine learning approaches, often decision trees, with neural networks. We distinguish between fully differentiable models, that can be differentiated with respect to all their parameters and partly differentiable models.

% \begin{figure}
%     \centering
%     \includegraphics[width=0.85\columnwidth]{images/methods/DeepGBM.PNG}
%     \caption{An overview over the DeepGBM architecture~\citep{ke2019deepgbm}, a combination of two NNs, one handling the spare categorical features, one processing the dense numerical ones.}
%     \label{fig:DeepGBM}
% \end{figure}

\textbf{Fully differentiable Models.}
The fully differentiable models in this category offer a valuable property: They permit end-to-end deep learning for training and inference by means of gradient descent optimizers. Thus, they allow for highly efficient implementations in modern deep learning frameworks that exploit GPU or TPU acceleration throughout the code. 

Popov et al. \cite{popov2019neural} propose an ensemble of differentiable oblivious decision trees \citep{langley1994oblivious} -- also known as the NODE framework for deep learning on tabular data. 
Oblivious decision trees use the same splitting function for all nodes on the same level and can therefore be easily parallelized. NODE is inspired by the successful CatBoost \citep{prokhorenkova2018catboost} framework.
To make the whole architecture fully differentiable and benefit from end-to-end optimization, NODE utilizes the entmax transformation \citep{peters2019sparse} and soft splits. 
In the original experiments, the NODE framework outperforms XGBoost and other GBDT models on many data sets. As NODE is based on decision tree ensembles, there is no preprocessing or transformation of the categorical data necessary. 
Decision trees are known to handle discrete features well. 
In the official implementation, strings are converted to integers using the leave-one-out encoding scheme.
The NODE framework is widely used and provides a sound implementation that can be readily deployed.

Frosst \& Hinton \cite{frosst2017distilling} contribut another model relying on soft decision trees (SDT) to make neural networks more interpretable. 
They investigated training a deep neural network first, before using a mixture of its outputs and the ground truth labels to train the SDT model in a second step.
This also allows for semi-supervised learning with unlabelled samples that are labelled by the deep neural network and used to train a more robust decision tree along with the labelled data. 
The authors showed that training a neural model first increases accuracy over SDTs that are directly learned from the data. 
However, their distilled trees still exhibit a performance gap to the neural networks that were fitted in the initial step. 
Nevertheless, the model itself shows a clear relationship among different classes in a hierarchical fashion. 
It groups different categorical values based on the common patterns, e.g., the digits 8 and 9 from the MNIST data set \citep{lecun-mnisthandwrittendigit-2010}. 
To summarize, the proposed method allows for high interpretability and efficient inference, at the cost of slightly reduced accuracy.

Follow-up work~\citep{Luo2021sdtr} extends this line of research to heterogeneous tabular data and regression tasks and presents the soft decision tree regressor (SDTR) framework. 
The SDTR is a neural network which imitates a binary decision tree. Therefore, all neurons, like nodes in a tree, get the same input from the data instead of the output from previous layers.
In the case of deep networks, the SDTR could not beat other state-of-the-art models, but it has shown promising results in a low-memory setting, where single tree models and shallow architectures were compared.

%As the soft decision trees are differentiable, SDTRs can be included in end-to-end learning pipelines. 
%SDTRs use neural networks to imitate binary decision trees, which leads to interpretability worse than GBDTs but better than standard NNs. In low-memory, single tree settings, SDTRs show their best performance compared to the other state-of-the-art models.

\revised{Katzir et al. \cite{katzir2021netdnf} follow a related idea.
Their Net-DNF builds on the observation that} every decision tree is merely a form of a Boolean formula, more precisely a disjunctive normal form. 
They use this inductive bias to design the architecture of a neural network, which is able to imitate the characteristics of the gradient boosting decision trees algorithm. 
The resulting Net-DNF was tested for classification tasks on data sets with no missing values, where it showed results that are comparable to those of XGBoost \citep{XGBOOST}. 
However, the authors did not mention how to handle high-cardinality categorical data, as the used data sets contained mostly numerical and few binary features.

%The Net-DNF method is another decision tree inspired approach was proposed by~\cite{katzir2021netdnf}, taking into account that every decision tree is merely a form of a boolean normal form. The structure of their Net-DNF tries to imitate the characteristics of the gradient boosting decision trees algorithm. The proposed deep neural network architecture only needs four layers in total, and therefore relatively few hyperparameters. The experiments on multiple data sets show that the Net-DNF outperforms FCNs in almost every task but is slightly worse than the gradient boosting decision tree algorithm under fine-tuned settings. Also, they only consider classification tasks using data sets with no missing values that do not require preprocessing.  Also the Net-DNF paper does not talk about the handling of categorical data. It seems that in their experiments only data sets with numerical features where used.

Linear models (e.g., linear and logistic regression) provide global interpretability but are inferior to complex deep neural networks. Usually, handcrafted feature engineering is required to improve the accuracy of linear models.
Liu et al. \cite{liu2020dnn2lr} use a deep neural network to combine the features in a possibly non-linear way; the resulting combination then serves as input to the linear model. This enhances the simple model while still providing interpretability.

%To overcome this issue, DNN2LR~\citep{liu2020dnn2lr} uses DNNs to calculate feasible cross features. A linear model can select the valuable features based on this candidate set, including cross-feature combinations. In their experiments, the empowered DNN2LR model can outperform other complex DNN models while still providing interpretability.

The work by Cheng et al. \cite{cheng2016wide} proposes a hybrid architecture that consists of linear and deep neural network models -- Wide\&Deep. A linear model that takes single features and a wide selection of hand-crafted logical expressions on features as an input is enhanced by a deep neural network to improve the generalization capabilities. Additionally, Wide\&Deep learns an $n$-dimensional embedding vector for each categorical feature. All embeddings are concatenated resulting in a dense vector used as input to the neural network. The final prediction can be understood as a sum of both models. A similar work by Guo et al. \cite{guo2016entity} proposes an embedding using deep neural networks for categorical variables. 

Another contribution to the realm of Wide\&Deep models is DeepFM \cite{guo2017deepfm}. The authors demonstrate that it is possible to replace the hand-crafted feature transformations with learned Factorization Machines (FMs) \citep{rendle2010factorization}, leading to an improvement of the overall performance. The FM is an extension of a linear model designed to capture interactions between features within high-dimensional and sparse data efficiently.
Similar to the original Wide\&Deep model, DeepFM also relies on the same embedding vectors for its “wide” and “deep” parts. In contrast to the original Wide\&Deep model, however, DeepFM alleviates the need for manual feature engineering.

Lastly, Network-on-Network (NON)~\citep{luo2020NON} is a classification model for tabular data, which focuses on capturing the intra-feature information efficiently. It consists of three components: a field-wise network consisting of one unique deep neural network for every column to capture the column-specific information, an across-field-network, which chooses the optimal operations based on the data set, and an operation fusion network, connecting the chosen operations allowing for non-linearities. As the optimal operations for the specific data are selected, the performance is considerably better than that of other deep learning models. However, the authors did not include decision trees in their baselines, the current state-of-the-art models on tabular data. Also, training as many neural networks as columns and selecting the operations on the fly may lead to a long computation time.

%%%%%%%%%%%%%%%%%%%%%%%%%%%%%%%%%%%%%%%%%%%%%%%%
%%%%%%%%%%%%%%%%%%%%%%%%%%%%%%%%%%%%%%%%%%%%%%%%
\textbf{Partly differentiable Models.} 
This subgroup of hybrid models aims at combining non-differentiable approaches with deep neural networks. Models from this group usually utilize decision trees for the non-differentiable part. 

% The method proposed in~\cite{moosmann2007fast} shows how the data can be encoded using the Random Forests (RF) algorithm by accessing leaf indices in the decision trees. The same idea was presented in~\cite{FACEBOOK}, where instead of the random forest algorithm, trees from the gradient boosting decision tree (GBDT) are used for the categorical data encoding. This work empirically demonstrates that the boosted decision trees are a powerful and convenient way to implement non-linear and categorical feature transformations for heterogeneous data. The DeepGBM framework~\cite{DEEPGBM} further evolved the idea of distilling knowledge from decision trees leaf index by encoding them using a neural network for online learning tasks. This approach recived a lot of attention, but the leaf indices' from a decision tree embedding do not fully represent the whole decision tree structure. Hence, each boosted tree is treated as a new \textit{meta categorical feature}.

The DeepGBM model~\citep{ke2019deepgbm} combines the flexibility of deep neural networks with the preprocessing capabilities of gradient boosting decision trees. 
DeepGBM consists of two neural networks -- CatNN and GBDT2NN. While CatNN is specialized to handle sparse categorical features, GBDT2NN is specialized to deal with dense numerical features.  

In the preprocessing step for the CatNN network, the categorical data are transformed via an ordinal encoding (to convert the potential strings into integers), and the numerical features are discretized, as this network is specialized for categorical data.
The GBDT2NN network distills the knowledge about the underlying data set from a model based on gradient boosting decision trees by accessing the leaf indices of the decision trees. This embedding based on decision tree leaves was first proposed by \cite{moosmann2006fast} for the random forest algorithm. Later, the same knowledge distillation strategy has been adopted for gradient boosting decision trees \citep{FACEBOOK}. 

Using the proposed combination of two deep neural networks, DeepGBM has a strong learning capacity for both categorical and numerical features. 
\revised{Distinctively, the authors implemented and tested DeepGBM's online prediction performance, which is significantly higher than that of gradient boosting decision trees.} 
On the downside, the leaf indices can be seen as meta categorical features since these numbers cannot be directly compared.
Also, it is not clear how other data-related issues, such as missing values, different scaling of numeric features, and noise influence the predictions produced by the models.

% \begin{figure}[t]
%     \centering
%     \includegraphics[width=\columnwidth]{images/methods/tabnet_encoder.pdf}
%     \caption{The TabNet encoder is composed of a feature transformer, an attentive transformer and feature masking. Next to the network output also feature attributes for interpretability are provided~\citep{arik2019tabnet}.}
%     \label{fig:tabnet_encoder}
% \end{figure}

% \begin{figure}
%     \centering
%     \includegraphics[width=0.7\columnwidth]{images/methods/TabTransformer.png}
%     \caption{The architecture of the TabTransformer by~\cite{Huang2020tabtrans}. The categorical data are transformed into a robust contextual embedding, before it is concatenated with the numerical data and put into a MLP.}
%     \label{fig:TabTransformer}
% \end{figure}

The TabNN architecture, introduced by~\cite{Ke2019tabnn}, is based on two principles: explicitly leveraging expressive feature combinations and reducing model complexity. 
It distills the knowledge from gradient boosting decision trees to retrieve feature groups; it clusters them and then constructs the neural network based on those feature combinations. 
Also, structural knowledge from the trees is transferred to provide an effective initialization. 
However, the construction of the network already takes different extensive computation steps of which one is only a heuristic to avoid an NP-hard problem. \revised{Overall, considering the construction challenges and that an implementation of TabNN was not provided, the practical use of the network seems limited.}

In similar spirit to DeepGBM and TabNN, the work from~\cite{Ivanov2021bgnn} proposes using gradient boosting decision trees for the data prepossessing step. 
The authors show that a decision tree structure has the form of a directed graph. 
Thus, the proposed framework exploits the topology information from the decision trees using graph neural networks \citep{scarselli2008graph}. The resulting architecture is coined Boosted Graph Neural Network (BGNN). In multiple experiments, BGNN demonstrates that the proposed architecture is superior to existing solid competitors in terms of predictive performance and training time. 

% \begin{figure*}
%     \centering
%     \includegraphics[width=0.6\textwidth]{images/methods/armnet-overview-cropped.pdf}
%     \caption{Overview on ARM-Net architecture proposed by \cite{cai2021arm}. It utilizes the exponential transformation with multi-head gated attention network over embedding modules.}
%     \label{fig:armnet}
% \end{figure*}

% \begin{figure}
%     \centering
%     \includegraphics[width=0.54\columnwidth]{images/methods/SAINT_a.PNG}
%     \caption{SAINT~\citep{somepalli2021saint} consists of self-attention blocks followed by intersample attention blocks. After an embedding layer, the data are transformed by the SAINT model, before the predictions are done by a MLP.}
%     \label{fig:SAINT}
% \end{figure}

%%%%%%%%%%%%%%%%%%%%%%%%%%%%%%%%%%%%%%%%%%%%%%%%
\subsubsection{Transformer-based Models}
\label{sec:transformer_based}

Transformer-based approaches form another subgroup of model-based deep neural methods for tabular data. 
Inspired by the recent surge of interest in transformer-based methods and their successes on text and visual data \citep{wang2019language,khan2021transformers}, researchers and practitioners have proposed multiple approaches using deep attention mechanisms \citep{vaswani2017attention} for heterogeneous tabular data.

TabNet \citep{arik2019tabnet} is one of the first transformer-based models for tabular data. Like a decision tree, the TabNet architecture comprises multiple subnetworks that are processed in a sequential hierarchical manner. According to \cite{arik2019tabnet}, each subnetwork corresponds to one  \textit{decision step}. To train TabNet, each decision step (subnetwork) receives the current data batch as input. TabNet aggregates the outputs of all decision steps to obtain the final prediction.
At each decision step, TabNet first applies a sparse feature mask \citep{Martins2016sparsemax} to perform soft instance-wise feature selection. The authors claim that the feature selection can save valuable resources, as the network may focus on the most important features. The feature mask of a decision step is trained using attentive information from the previous decision step. To this end, a \textit{feature transformer} module decides which features should be passed to the next decision step and which features should be used to obtain the output at the current decision step. Some layers of the feature transformers are shared across all decision steps.
The obtained feature masks correspond to local feature weights and can also be combined into a global importance score. Accordingly, TabNet is one of the few deep neural networks that offers different levels of interpretability by design. Indeed, experiments show that each decision step of TabNet tends to focus on a particular subdomain of the learning problem (i.e., one particular subset of features). This behaviour is similar to convolutional neural networks.
TabNet also provides a decoder module that is able to preprocess input data (e.g., replace missing values) in an unsupervised way. Accordingly, TabNet can be used in a two-stage self-supervised learning procedure, which improves the overall predictive quality. \added{One of the popular Python \cite{van1995python} frameworks for tabular data provides an efficient implementation of TabNet \cite{joseph2021pytorch}.}  
Recently, TabNet has also been investigated in the context of fair machine learning \citep{boughorbel2021fairness,mehrabi2021survey}.

Attention-based architectures offer mechanisms for interpretability, which is an essential advantage over many hybrid models. \added{Figure \ref{fig:tabnet_kernelshap} shows attentions maps of the TabNet model and KernelSHAP explanation framework on the Adult data set \cite{Dua:2019}.}

% The historically first model based on the attention mechanism is TabNet~\citep{arik2019tabnet}. It utilizes sequential attention to choose  the most salient features in each step instance-wise. This leads to very efficient training and makes all decisions interpretable, both locally and globally. For the sparse instance-wise feature selection a sparsemax normalization is used, introduced by~\citep{Martins2016sparsemax}. Each decision step contains a split, deciding what information is the output and which is input for the next step. The masks values correspond to the importance of the features in the decision and can later be used for interpretability. As it can be seen in figure \ref{fig:tabnet_encoder}, next to the normal network output also the feature attributes derived from the masks are provided. If a large amount of unlabelled data are available TabNet is able to use this for a unsupervised pre-training and the labelled data later for a supervised fine-tuning. TabNet is also one of the very few deep learning architectures for tabular data which provides interpretability at all, but a large amount of data are needed to provide a reliable performance.
% The TabNet implementation is designed to work with categorical data. Therefore only strings need to be transformed as integer, e.g. via a ordinal- or binary encoding.
% The work~\citep{boughorbel2021fairness} provides a fairness analysis~\citep{mehrabi2021survey} of the TabNet framework.

Another supervised and semi-supervised approach is introduced by Huang et al.~\cite{Huang2020tabtrans}. Their TabTransformer architecture uses self-attention-based transformers to map the categorical features to a contextual embedding. This embedding is more robust to missing or noisy data and enables interpretability. The embedded categorical features are then together with the numerical ones fed into a simple multilayer perceptron. If, in addition, there is an extra amount of unlabelled data, unsupervised pre-training can improve the results, using masked language modelling or replace token detection. Extensive experiments show that TabTransformer matches the performance of tree-based ensemble techniques, showing success also when dealing with missing or noisy data. 
The TabTransformer network puts a significant focus on the categorical features. It transforms the embedding of those features into a contextual embedding which is then used as input for the multilayer perceptron. This embedding is implemented by different multi-head attention-based transformers, which are optimized during training. 
% The architecture of the TabTransformer model is depicted in Fig. \ref{fig:TabTransformer}.

ARM-net \cite{cai2021arm} is an adaptive neural network for relation modelling tailored to tabular data. The key idea of the ARM-net framework is to model feature interactions with combined features (feature crosses) selectively and dynamically by first transforming the input features into exponential space and then determining the interaction order and interaction weights adaptively for each feature cross. Furthermore, the authors propose a novel sparse attention mechanism to generate the interaction weights given the input data dynamically. Thus, users can explicitly model feature crosses of arbitrary orders with noisy features filtered selectively. 

% The architecture of the ARM-net is depicted in Fig. \ref{fig:armnet}. 

SAINT (Self-Attention and Intersample Attention Transformer)~\citep{somepalli2021saint} is a hybrid attention approach, combining self-attention~\citep{vaswani2017attention} with inter-sample attention over multiple rows. When handling missing or noisy data, this mechanism allows the model to borrow the corresponding information from similar samples, which improves the model's robustness. The technique is reminiscent of nearest-neighbour classification. In addition, all features are embedded into a combined dense latent vector, enhancing existing correlations between values from one data point. To exploit the presence of unlabelled data, a self-supervised contrastive pre-training can further improve the results, minimizing the distance between two views of the same sample and maximizing the distance between different ones.
Like the VIME framework (Section \ref{sec:data_encoding}), SAINT uses CutMix \citep{yun2019cutmix} to augment samples in the input space and uses mixup \citep{zhang2017mixup} in the embedding space.

%Figure \ref{fig:SAINTexp} shows attention maps on the MNIST \citep{lecun-mnisthandwrittendigit-2010} and volkert \citep{gijsbers2019open} data sets, which can be found using different attention mechanisms.

\added{Finally, even some new learning paradigms are being proposed. For instance, the Non-Parametric Transformer (NPT) \cite{kossen2021selfattention} does not construct a mapping from individual inputs to outputs but uses the entire data set at once. By using attention between data points, relations between arbitrary samples can be modelled and leveraged for classifying test samples.}

%%%%%%%%%%%%%%%%%%%%%%%%%%%%%%%%%%%%%%%%%%%%%%%%
\subsection{Regularization Models}
\label{sec:reg_models}

The third group of approaches argues that extreme flexibility of deep learning models for tabular data is one of the main learning obstacles and strong regularization of learned parameters may improve the overall performance.

One of the first methods in this category was the Regularization Learning Network (RLN) proposed by Shavitt \& Segal \cite{shavitt2018regularization}, which uses a learned regularization scheme. \revised{The main idea is to apply trainable regularization coefficients to each single weight in a neural network, thereby lowering the sensitivity.
%\begin{equation}
%    \mathcal{L}_{C}\left(Z,W,\lambda\right)=\mathcal{L}\left(Z,W\right)+\sum_{i=1}^{n} \exp\left(\lambda_i\right) \cdot \left\Vert w_{i}\right\Vert,
%    \label{eq:rln_loss}
%\end{equation}
%where $Z=\left\{ \left(x_{m},y_{m}\right)\right\} _{m=1}^{M}$ are
%the training samples, $\mathcal{L}$ is the classification loss given the model with weights, $W=\left\{ w_{i}\right\} _{i=1}^{n}$. The norm, $\left\Vert \cdot\right\Vert$, can be chosen according to use-case-specific sparsity requirements, and the $\lambda_i$, i.e., the regularization coefficients, are learned for each weight $w_i$. 
To efficiently determine the corresponding coefficients, the authors propose a novel loss function termed ``Counterfactual Loss''.} The regularization coefficients lead to a very sparse network, which also provides the importance of the remaining input features.

In their experiments, RLNs outperform deep neural networks and obtain results comparable to those of the gradient boosting decision trees algorithm, but the evaluation relies on a data set with mainly numerical data to compare the models.
The RLN paper does not address the issues of categorical data. 
For the experiments and the example implementation data sets with exclusively numerical data (except for the gender attribute) were used.
A similar idea is proposed in \citep{borisov2019cancelout}, where regularization coefficients are learned only in the first layer with a goal to extract feature importance. 
%Therefore we tried different encoding methods and in the end decided to use leave-one-out encoding.

Kadra et al. \cite{Kadra2021reg} state that simple multilayer perceptrons can outperform state-of-the-art algorithms on tabular data if deep learning networks are properly regularized.
The authors propose a ``cocktail'' of regularization with thirteen different techniques that are applied jointly. 
From those, the optimal subset and their subsidiary hyperparameters are selected. 
They demonstrate in extensive experiments that the ``cocktails'' regularization can not only improve the performance of multilayer perceptrons, but these simple models also outperform tree-based architectures.
On the downside, the extensive per-data set regularization and hyperparameter optimization take much more computation time than the gradient boosting decision trees algorithm.

There are several other works \citep{valdes2021lockout, fiedler2021simple, lounici2021muddling} showing that strong regularization of deep neural networks can be beneficial for tabular data. 

\ifdefined\ieeeversion
\begin{figure}[t!]
     \centering
     \subfloat[TabNet attributions]{
         \centering
         \includegraphics[width=4cm]{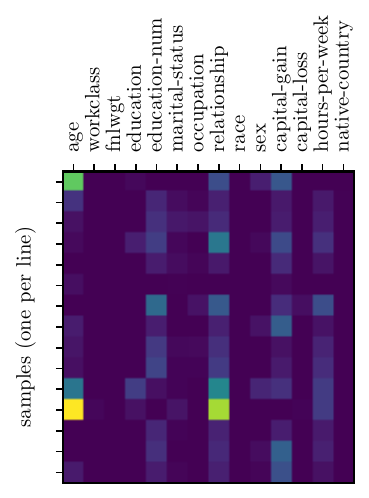}
     }
     \subfloat[KernelSHAP attributions]{
         \centering
         \includegraphics[width=4cm]{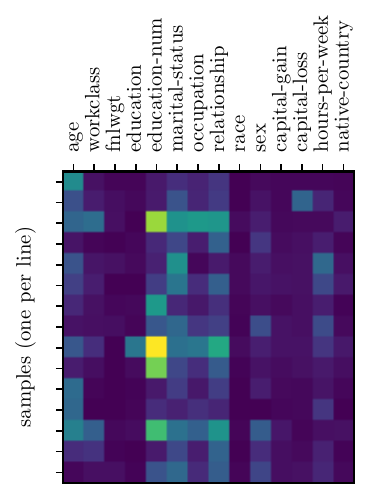}
     }
    \caption{Interpretable learning with the TabNet \cite{arik2019tabnet} architecture. We compare the attributions provided by the model for a sample from the UCI Adult data set with those provided by the game theoretic KernelSHAP framework \cite{lundberg2017unified}. }
    \label{fig:tabnet_kernelshap}
\end{figure}
\else
\begin{figure}[t!]
     \centering
     \begin{subfigure}[b]{0.45\columnwidth}
         \centering
         \includegraphics[width=\textwidth]{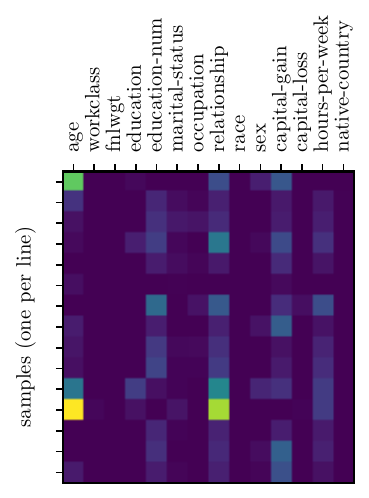}
         \caption{TabNet attributions}
     \end{subfigure}
     \begin{subfigure}[b]{0.45\columnwidth}
         \centering
         \includegraphics[width=\textwidth]{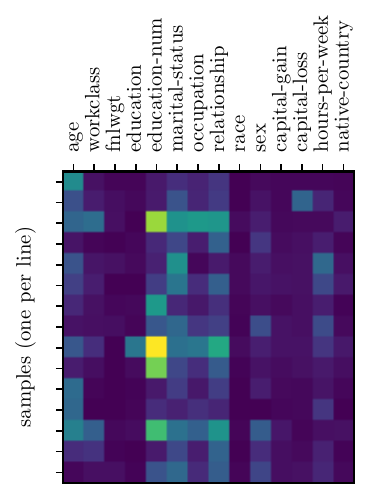}
         \caption{KernelSHAP attributions}
     \end{subfigure}
    
    \caption{Interpretable learning with the SAINT architecture from \cite{somepalli2021saint} by visual representations of various attention maps, (a) MNIST \citep{lecun-mnisthandwrittendigit-2010} and (b) a tabular data set \citep{somepalli2021saint}. }
    \label{fig:tabnet_kernelshap}
\end{figure}
\fi

\section{Tabular Data Generation}
\label{sec:tab_data_generation}

For many applications, the generation of realistic tabular data is fundamental. 
Two of the main purposes are data augmentation~\citep{chen2019faketables} and data imputation (i.e., the filling of missing values)~\citep{gondara2018mida, camino2020working} and rebalancing~\citep{engelmann2021conditional, Quintana2019towardsHumanComfort, koivu2020synthetic, darabi2021synthesising}.
Another highly relevant topic is privacy-aware machine learning~\citep{Choi2017medGAN, Fan2020designSpace, kamthe2021copula} where generated data can potentially be leveraged to overcome privacy concerns.

\subsection{Methods}
While the generation of images and text is highly explored~\citep{karras2020analyzing, lin2017adversarial, subramanian2017adversarial}, generating synthetic tabular data is still a challenge. 
The mixed structure of discrete and continuous features along with their different value distributions \added{still poses a significant challenge}.

Classical approaches for the data generation task include Copulas~\citep{patki2016synthetic, li2020sync} and Bayesian Networks \citep{zhang2017privbayes}. 
Among Bayesian Networks those based on the Chow-Liu approximation \cite{chow1968approximating} are especially popular.

In the deep-learning era, Generative Adversarial Networks (GANs)~\citep{goodfellow2014generative} have proven highly successful for the generation of images~\citep{radford2015unsupervised, karras2020analyzing}.
GANs were recently introduced as an original way to train a generative deep neural network model. 
They consist of two separate models: a generator ${G}$ that generates samples from the data distribution, and a discriminator ${D}$ that estimates the probability that a sample came from the ground truth distribution. 
Both ${G}$ and ${D}$ are usually chosen to be non-linear functions such as a multilayer perceptrons.
To learn a generator distribution ${p_g}$ over data ${\bm{x}}$, the generator ${G(\bm{z};\theta_g)}$ maps samples from a noise distribution ${p_z(\bm{z})}$ (e.g., the Gaussian distribution) to the input data space. 
The discriminator ${D(\bm{x}; \theta_d)}$ outputs the probability that a data point ${\bm{x}}$ comes from the training data's distribution $p_{data}$ rather than from the generator's output distribution ${p_g}$. % Both samples from the ground truth data and from the generator are fed to the discriminator. 
During joint training of ${G}$ and ${D}$, $G$ will start generating successively more realistic samples to fool the discriminator $D$.
%are both trained together: we adjust parameters for ${G}$ to minimize ${\log(1-D(G(\bm{z}))}$ and adjust parameters for $D$ to minimize ${\log D(x)}$, as if they are following the two-player min-max game with value function ${V(G, D)}$:
%\begin{align}
%\begin{split}
%\min_G \max_D V(G, D) = \min_G \max_D \mathbb{E}_{\bm{x} \sim p_{\text{data}}(\bm{x})}[\log D(\bm{x})] + \\
%\mathbb{E}_{\bm{z} \sim p_z(\bm{z})}[\log (1 - D(G(\bm{z})))].
%\label{eq:minimaxgame-definition}
%\end{split}
%\end{align}
For more details on GANs, we refer the interested reader to the original paper~\citep{goodfellow2014generative}.

Although it was found that GANs lag behind at the generation of discrete outputs such as natural language~\citep{subramanian2017adversarial}, they are still frequently chosen to generate tabular data.
Vanilla GANs or derivates such as the Wasserstein GAN (WGAN)~\citep{arjovsky2017wasserstein}, WGAN with gradient penalty (WGAN-GP)~\citep{gulrajani2017improved}, Cramér GAN~\citep{bellemare2017cramer}, or the Boundary seeking GAN~\citep{hjelm2017boundary}, which is designed to model discrete data, are commonly used in the literature to generate tabular data. 
Moreover, VeeGAN~\citep{srivastava2017veegan} is frequently used for tabular data.
Apart from GANs, autoencoder-based architectures -- in particular those relying on Variational Autoencoders (VAEs) \cite{Kingma2014VAE} -- have  been proposed~\citep{Ma2020VAEM, Xu2019CTGAN}. 

In Table~\ref{tab:generation_tabular_models}, we provide an overview of tabular generation approaches, that use deep learning techniques. 
Note that due to the enormous number of approaches, we list the most influential works that address the problem of data generation with a particular focus on tabular data. 
We exclude works that are targeted towards highly domain-specific tasks. % review of related work can also be found in~\citep{engelmann2021conditional}.

In the following section, we will briefly discuss the most relevant approaches that helped shape the domain.
For example, MedGAN by \cite{Choi2017medGAN} was one of the first works and provides a deep learning model to generate patient records. 
As all the features in their work are discrete, this model cannot be easily transferred to arbitrary tabular data sets. 
The table-GAN approach by~\citep{Park2018GAN} adapts the Deep Convolutional GAN for tabular data. 
Specifically, the features from one record are converted into a matrix, so that they can be processed by convolutional filters of a convolutional neural network.
However, it remains unclear to which extent the inductive bias used for images are suitable for tabular data.

% \begin{figure}[t]
%     \centering
%     \includegraphics[width=\columnwidth]{images/methods/CTGAN.pdf}
%     \caption{Overview of the CTGAN model. A condition is sampled first and passed to the conditional generator $G$ along with a random input $z$. The generated sample is opposed to a randomly picked example from the data set that also fulfills the condition and assessed by the conditional discriminator $D$. This approach allows to preserve dependency relations. Adapted from \citep{Xu2019CTGAN}.}
%     \label{fig:CTGAN}
% \end{figure}

The approach by Xu et al. \cite{Xu2019CTGAN} focuses on the correlation between the features of one data point. 
The authors first propose the \textit{mode-specific normalization} technique for data preprocessing that allows to transform non-Gaussian distributions in the continuous columns. 
They express numeric values in terms of a mixture component number and the deviation from that component's center. 
This allows to represent multi-modal and skewed distributions. 
Their generative solution, coined CTGAN, uses the conditional GAN architecture %depicted in Fig.~\ref{fig:CTGAN} 
to enforce learning proper conditional distributions for each column. 
To obtain categorical values and to allow for backpropagation in the presence of categorical values, the \textit{gumbel-softmax trick}~\citep{jang2016categorical} is utilized. 
The authors also propose a model based on Variational Autoencoders, named TVAE (Tabular Variational Autoencoder) which outperforms their suggested GAN approach. 
Both approaches can be considered state-of-the-art.

While GANs and VAEs are prevalent, other recently proposed architectures include machine-learned Causal Models~\citep{wen2021causal} and Invertible Flows~\citep{kamthe2021copula}. When privacy is the main factor of concern, models such as PATE-GAN~\citep{jordon2018pate} provide generative models with certain differential privacy guarantees. Although very relevant for practical applications, such privacy guarantees and related federated learning approaches with tabular data are outside the scope of this review.

Fan et~al.~\citep{Fan2020designSpace} compare a variety of different GAN architectures for tabular data synthesis and recommend using a simple, fully connected architecture with a vanilla GAN loss with minor changes to prevent mode-collapse. They also use the normalization proposed by~\citep{Xu2019CTGAN}. In their experiments, the Wasserstein GAN loss or the use of convolutional architectures on tabular data does boost the generative performance.

\begin{table*}[]
    \centering
    \begin{tabularx}{0.66
    \textwidth}{lcc}
    \toprule
         Method & Based upon & Application \\
         \midrule
         medGAN \cite{Choi2017medGAN} & Autoencoder+GAN & Medical Records\\
         
         TableGAN \cite{Park2018GAN} DCGAN & General \\
         
         Mottini et~al. \cite{Mottini2018GAN} & Cramér GAN  & Passenger Records \\
         Camino et~al. \cite{camino2018generating} &  medGAN, ARAE & General\\
         medBGAN, medWGAN \cite{Baowaly2019GAN} &  \makecell{WGAN-GP, \\Boundary seeking GAN}& Medical Records\\\addlinespace
         
         ITS-GAN \cite{chen2019faketables} & \makecell{GAN with AE \\ for constraints} & General \\\addlinespace
         CTGAN, TVAE \cite{Xu2019CTGAN} & Wasserstein GAN, VAE & General \\
         actGAN \cite{koivu2020synthetic}  &  WGAN-GP & Health Data \\
         
         VAEM \cite{Ma2020VAEM} & VAE (Hierarchical) & General \\
         
         OVAE \cite{vardhan2020generating} & Oblivious VAE & General \\
         
         TAEI \cite{darabi2021synthesising} & \makecell{ AE+SMOTE (in\\ multiple setups)}  & General \\\addlinespace
         
         Causal-TGAN \cite{zhao2021ctab} & Causal-Model, WGAN-GP & General \\
         
         Copula-Flow \cite{kamthe2021copula} & Invertible Flows & General \\
         \bottomrule
    \end{tabularx}
    \caption{Generation of tabular data using deep neural network models (in chronological order).}
    \label{tab:generation_tabular_models}
\end{table*}

\subsection{Assessing Generative Quality}
\label{sec:gan_quality}
To assess the quality of the generated data, several performance measures are used. The most common approach is to define a proxy classification task and train one model for it on the real training set and another on the artificially generated data set. With a highly capable generator, the predictive performance of the artificial-data model on the real-data test set should be almost on par with its real-data counterpart. This measure is often referred to as \textit{machine learning efficacy} and used in~\citep{Choi2017medGAN, Mottini2018GAN, Xu2019CTGAN}. In non-obvious classification tasks, an arbitrary feature can be used as a label and predicted~\citep{Choi2017medGAN, camino2018generating, Baowaly2019GAN}. Another approach is to visually inspect the modelled distributions per-feature, e.g., the cumulative distribution functions~\citep{chen2019faketables} or compare the expected values in scatter plots~\citep{Choi2017medGAN, camino2018generating}. A more quantitative approach is the use of statistical tests, such as the Kolmogorov-Smirnov test \citep{massey1951kolmogorov}, to assess the distributional difference~\citep{Baowaly2019GAN}. On synthetic data sets, the output distribution can be compared to the ground truth, e.g., in terms of log-likelihood~\citep{Xu2019CTGAN, wen2021causal}. Because overfitted models can also obtain good scores,~\citep{Xu2019CTGAN} propose evaluating the likelihood of a test set under an estimate of the GAN's output distribution. Especially in a privacy-preserving context, the distribution of the \textit{Distance to Closest Record} (DCR) can be calculated and compared to the respective distances on the test set~\citep{Park2018GAN}. \added{This measure is important to assess the extent of sample memorization. Overall, we conclude that a single measure is not sufficient to assess the generative quality. For instance, a generative model that memorizes the original samples will score well in the machine learning efficiency metric but fail the DCR check. Therefore, we highly recommend using several evaluation measures that focus on individual aspects of data quality. }

\section{Explanation Mechanisms for Deep Learning with Tabular Data}
\label{sec:explanation}
Explainable machine learning is concerned with the problem of providing explanations for complex machine learning models. \added{With stricter regulations for automated decision making \cite{regulation2016gdpr} and the adoption of machine learning models in high-stakes domains such as finance and healthcare \cite{bhatt2020explainable}, interpretability is becoming a key concern.}
Towards this goal, various streams of research follow different explainability paradigms. \revised{Among these, feature attribution methods and counterfactual explanations are two of the popular forms \citep{guidotti2018survey,gade2019explainable,pawelczyk2021carla}.} \added{Because these techniques are gaining importance for researchers and practitioners alike, we dedicate the following section to reviewing these methods.}

\subsection{Feature Highlighting Explanations}
\textit{Local} input attribution techniques seek to explain the behaviour of machine learning models instance by instance. \revised{Those methods aim to highlight the influence the inputs have on the prediction by assigning importance scores to the input features.} Some popular approaches for model explanations aim at constructing classification models that are explainable by design \citep{lou2012intelligible,alvarez2018towards,wang2019designing}. This is often achieved by enforcing the deep neural network model to be locally linear. Moreover, if the model's parameters are known and can be accessed, then the explanation technique can use these parameters to generate the model explanation. For such settings, relevance-propagation-based methods, e.g., \cite{bach2015pixel,montavon2019layer}, and gradient-based approaches, e.g.,~\citep{kasneci2016licon,sundararajan2017axiomatic,chattopadhay2018grad}, have been suggested. In cases where the parameters of the neural network cannot be accessed, model-agnostic approaches can prove useful. This group of approaches seeks to explain a model's behavior locally by applying surrogate models~\citep{ribeiro2016should, lundberg2017unified,ribeiro2018anchors,lundberg2020local,haug2021baselines}, which are interpretable by design and are used to explain
individual predictions of black-box machine learning models. In order to test the performance of these black-box explanations techniques, Liu et al. \cite{liu2021synthetic} suggest a python-based benchmarking library. 
%In contrast to model explanations, such local explanations do not have to be globally valid.

% \paragraph{\textbf{Explanations by example}} Example-based explanation methods select particular instances from the dataset or exploit particular instances provided by human experts to explain the behavior of an ML model or to explain the underlying data distribution~\cite{mittelstadt2019explaining, gade2019explainable}.

%\paragraph{\textbf{Libraries}}

\subsection{Counterfactual Explanations}

From the perspective of algorithmic recourse, the main purpose of counterfactual explanations is to suggest constructive interventions to the input of a deep neural network so that the output changes to the advantage of an end user. \added{In simple terms, a minimal change to the feature vector  that will flip the classification outcome is computed and provided as an explanation.}
By emphasizing both the feature importance and the recommendation aspect, counterfactual explanation methods can be further divided into three different groups: works that assume that all features can be independently manipulated \citep{wachter2017counterfactual} and works that focus on manifold constraints to capture feature dependencies.

In the class of independence-based methods, where the input features of the predictive model are assumed to be independent, some approaches use combinatorial solvers to generate recourse in the presence of feasibility constraints  \citep{ustun2019actionable,russell2019efficient,rawal2020individualized,karimi2019model}. Another line of research deploys gradient-based optimization to find low-cost counterfactual explanations in the presence of feasibility and diversity constraints \citep{Dhurandhar2018,mittelstadt2019explaining,mothilal2020fat}. 
The main problem with these approaches is that they abstract from input correlations. 
To alleviate this problem and to suggest realistic looking counterfactuals, researchers have suggested building recourse suggestions on generative models \citep{pawelczyk2019,downs2020interpretable,joshi2019towards,mahajan2019preserving,pmlr-v124-pawelczyk20a,antoran2020getting}. 
The main idea is to change the geometry of the intervention space to a lower dimensional latent space, which encodes different factors of variation while capturing input dependencies. 
To this end, these methods primarily use (tabular data) variational autoencoders \citep{Kingma2014VAE,nazabal2018handling}. 
In particular, Mahajan et al. \cite{mahajan2019preserving} demonstrate how to encode various feasibility constraints into such models.
\added{However, an extensive comparison across this class of methods is still missing since it is difficult to measure how realistic the generated data are in the context of algorithmic recourse.}

\added{More recently, a few works have suggested to develop counterfactual explanations that are robust to model shifts and noise in the recourse implementations \citep{upadhyay2021robust,dominguezolmedo2021adversarial,pawelczyk2022noisy}. 
A comprehensive treatment on how to extend these lines of work to arbitrary high cardinality categorical variables is still an open problem in the field.}

For a more fine-grained overview over the literature on counterfactual explanations we refer the interested reader to the most recent surveys \citep{karimi2020survey,verma2020counterfactual}. 
Finally, Pawelczyk et al. \cite{pawelczyk2021carla} have implemented an open-source python library which provides support for many of the aforementioned counterfactual explanation models.

%Work that deviates from this line of research was done by \cite{Poyiadzi2020,Kanamori2020}. The authors of \cite{Poyiadzi2020} provide \texttt{FACE}, which uses a shortest path algorithm on graphs to find counterfactual explanations. In contrast, \citet{Kanamori2020} use integer programming techniques to account for input dependencies

%%%%%%%%% Experiments %%%%%%%%%%
\section{Experiments}
\label{sec:experiments_main}
Although several experimental studies have been published in recent years \cite{Shwartz-Ziv2021, Kadra2021reg}, an exhaustive comparison between existing deep learning approaches for heterogeneous tabular data is still missing in the literature. For example, important aspects of deep learning models such as training and inference time, model size, and interpretability, are not discussed.   

To fill this gap, we present an extensive empirical comparison of machine and deep learning methods on real-world data sets with varying characteristics in this section. We discuss the data set choice (\ref{sec:dataset}), the results (\ref{sec:benchresults}), and present a comparison of the training and inference time for all the machine learning models considered in this survey (\ref{sec:benchtime}). We also discuss the size of deep learning models. Lastly, to the best of our knowledge, we present the first comparison of explainable deep learning methods for tabular data (\ref{sec:interpretbench}).
We release the full source code of our experiments for maximum transparency\footnote{Open benchmarking on tabular data for machine learning models: \href{https://github.com/kathrinse/TabSurvey}{https://github.com/kathrinse/TabSurvey}.}.

\subsection{Data Sets}
\label{sec:dataset}
In computer vision, there are many established data sets for the evaluation of new deep learning architectures such as MNIST \cite{lecun-mnisthandwrittendigit-2010}, CIFAR \cite{Krizhevsky09learningmultiple}, and ImageNet \cite{deng2009imagenet}.
On the contrary, there are no established standard heterogeneous data sets. \added{Carefully checking the works listed in Section~\ref{sec:neuralnetworks_for_tabdata}, we identified over 100 different data sets with different characteristics in their respective experimental evaluation sections. 
We note that the small overlap between the mentioned works makes it hard to compare the results across these works in general.}
Therefore, in this work, we deliberately select data sets covering the entire range of characteristics, such as data domain (e.g., finance, e-commerce, geography, \added{ physics}), \added{ different types of target variables (classification, regression)}, varying number of categorical variables \added{and continuous variables}, and differing sample sizes (small to large). Furthermore, most of the selected data sets were previously featured in multiple studies.

\added{The first data set of our study is the Home Equity Line of Credit (HELOC) data set provided by FICO \cite{HelocDataset}. 
This data set consists of anonymized information from real homeowners who applied for home equity lines of credit. 
A HELOC is a line of credit typically offered by a bank as a percentage of home equity.
The task consists of using the information about the applicant in their credit report to predict whether they will repay their HELOC account within a two-year period.}

We further use the Adult Income data set~\cite{Dua:2019}, \added{which is among the most popular tabular data sets used in the surveyed work (5 usages).} It includes basic information about individuals such as: age, gender, education, etc. The target variable is binary; it represents high and low income.

\added{The largest tabular data set in our study is HIGGS, which stems from particle physics. The task is to distinguish between signals with Higgs bosons (HIGGS) and a background process \cite{baldi2014searching}. Monte Carlo simulations \cite{mooney1997monte} were used to produce the data. In the first 21 columns (columns 2-22), the particle detectors in the accelerator measure kinematic properties. In the last seven columns, these properties are analyzed. In total, HIGGS includes eleven million rows. In contrast to other data sets of our study, the HIGGS data set contains only numerical or continuous variables. Since DeepFM, DeepGBM, and TabTransformer models require at least one categorical attribute, we binarize the twenty-first variable into a categorical variable with three groups. 
%Please confer our code for additional details.
}

%HIGGS contains mostly homogeneous and features only one artificial categorical feature among 26 continuous features.}

The Covertype data set~\cite{Dua:2019} is multi-classification data set which holds cartographic information about land cells (e.g., elevation, slope). The goal is to predict which one out of seven forest cover types is present in the cell.

Finally, we utilize the California Housing data set~\cite{pace1997sparse}, which contains information about a number of properties. The prediction task (regression) is to estimate price of the corresponding home.

The fundamental characteristics of the selected data sets are summarized in Table \ref{tab:tabular_datasets}.

%In all experiments, we use a categorical input encoding (instead of, e.g., one-hot encoding).

\begin{table}[]
\centering
{
% Setting the horizontal and vertical space
\renewcommand{\arraystretch}{0.8}
\setlength{\tabcolsep}{5pt} % 8pt
    
    \begin{tabular}{@{}lccccc@{}}  %{m{4cm}cl} %{@{}l@{}r@{}r@{}r@{}r@{}r@{}}}
        \toprule
        & \added{HELOC} & Adult & \added{HIGGS} & Covertype & California \\
        & & Income & &  & Housing \\
        \midrule
        
        Samples & 9.871 & 32.561 & 11 M. & 581.012 & 20.640  \\
        Num. features & 21 & 6 & 27 & 52 & 8\\
        Cat. features & 2 & 8 & 1 & 2 & 0\\
        Task & Binary & Binary & Binary & Multi-Class & Regression \\
        Classes & 2 & 2 & 2 & 7 & - \\
        \bottomrule
    \end{tabular}
}
    
    \caption{Main properties of the real-world heterogeneous tabular data sets used in this survey. \added{We also indicate the data set task, where ``Binary" stands for binary classification, and ``Multi-class" represents multi-class classification.}}
    \label{tab:tabular_datasets}
    
\end{table}

% \begin{figure}
%     \centering
%     \includegraphics[width=0.75\columnwidth]{images/cross_val.png}
%     \caption{The selected cross-validation scheme for the empirical evaluation of machine learning and deep learning methods. Note for the classification tasks we utilize the stratified splitting strategy.}
%     \label{fig:cv}
% \end{figure}

\subsection{Open Performance Benchmark on Tabular Data}
\label{sec:benchresults}
\newcommand{\wstd}[2]{#1\small{$\pm$#2}}
%\newcommand{\wstd}[2]{#1}

% Is this the correct way to get the values closer together?
%\setlength{\tabcolsep}{10pt}

% all experiments -> 
% https://docs.google.com/spreadsheets/d/1rVhCTE80vN0_b8yGDdZd5S9wqkx1JG1enEX0Y5Usaos/edit#gid=0

\begin{table*}[]
    \centering
\adjustbox{max width=\textwidth}{%
    \begin{tabular}{lccccccccc}  %{m{4cm}cl} {@{}l@{}r@{}r@{}r@{}r@{}r@{}}}
    \toprule
        % Data sets
        & \multicolumn{2}{c}{\added{HELOC}} 
        & \multicolumn{2}{c}{Adult} 
        & \multicolumn{2}{c}{\added{HIGGS}}
        & \multicolumn{2}{c}{Covertype}
        & \multicolumn{1}{c}{Cal.\ Housing} 
        \\
         
        % Metrics
        & Acc \textuparrow & AUC \textuparrow     % Heloc
        & Acc \textuparrow & AUC \textuparrow     % Adult
        & Acc \textuparrow & AUC \textuparrow     % HIGGS
        & Acc \textuparrow & AUC \textuparrow     % Covertype
        & MSE \textdownarrow      % California Housing
        \\
        \cmidrule(lr){2-3}\cmidrule(lr){4-5}\cmidrule(lr){6-7}\cmidrule(lr){8-9}\cmidrule(lr){10-10}
        
        % The values are automatically formatted by a python script that extracts the results from the google sheet
        \input{experiment_results/results5}\\
    \bottomrule
    \end{tabular}
}    
    \caption{Open performance benchmark results based on (stratified) 5-fold cross-validation. We use the same fold splitting strategy for every data set. \added{The top results for each dataset are in \textbf{bold}, we also \underline{underline} the second-best results.} The mean and standard deviation values are reported for each baseline model. Missing results indicate that the corresponding model could not be applied to the task type (regression or multi-class classification).}
    \label{tab:experiment_result}
\end{table*}

\subsubsection{Hyperparameter Selection} In order to do a fair evaluation, we use the Optuna library \cite{optuna_2019} with 100 iterations for each model to tune hyperparameters. Each hyperparameter configuration was cross-validated with five folds. The hyperparameter ranges used are publicly available online along with our code. We laid out the search space based on the information given in the corresponding papers and recommendations from the framework's authors. 

\subsubsection{Data Preprocessing}
We prepossessed the data in the same way for every machine learning model by applying zero-mean, unit-variance normalization to the numerical features and an ordinal encoding to the categorical ones. The missing values were substituted with zeros for the linear regression and models based on pure neural networks since these methods cannot accept them otherwise. \added{We apply the ordinal encoding to categorical values for all models. According to the work \cite{hancock2020survey}, the chosen encoding strategy shows comparable performance to more advanced methods. We explicitly specify which features are categorical for LightGBM, DeepFM, DeepGBM, TabNet, TabTransformer, and SAINT, since these approaches provide special functionality dedicated to categorical values, e.g., learning an embedding of the categories.}

\subsubsection{\added{Reproducibility and Extensibility}}
\label{sec:reproducibility}
\revised{For maximum reproducibility, we run all experiments in a docker container~\cite{merkel2014docker}. We underline again that our full code is publicly released so that the experiments can be replicated. The mentioned datasets are also publicly available and can be used as a benchmark for novel methods.
We would highly welcome contributed implementations of additional methods from the data science community.}

\subsubsection{Results}
The results of our experiments are shown in Table~\ref{tab:experiment_result}. They draw a different picture than many recent research papers may suggest: \revised{For all but the very large HIGGS data set, the best scores are still obtained by boosted decision tree ensembles. XGBoost and CatBoost outperform all deep learning-based approaches on the small and medium data sets, the regression data set, and the multi-class data set. For the large-scale HIGGS, SAINT outperforms the classical machine learning approaches. This suggests that for very large tabular data sets with predominantly continuous features, modern neural network architectures may have an advantage over classical approaches after all.
In general, however, our results are consistent with the inferior performance of deep learning techniques in comparison to approaches based on decision tree ensembles (such as gradient boosting decision trees) on tabular data that was observed in various Kaggle competitions~\cite{bojer2021kaggle}.}

\revised{Considering only deep learning approaches, we observe that SAINT provided competitive results across data sets. However, for the other models, the performance was highly dependent on the chosen data set. DeepFM performed best (among the deep learning models) on the Adult dataset and second-best on the California Housing data set, but returned only weak results on HELOC.}

\subsection{Run Time Comparison}
\label{sec:benchtime} 
\begin{figure*}[t!]
     \centering
     \subfloat{
         \centering
        \includegraphics[width=\columnwidth]{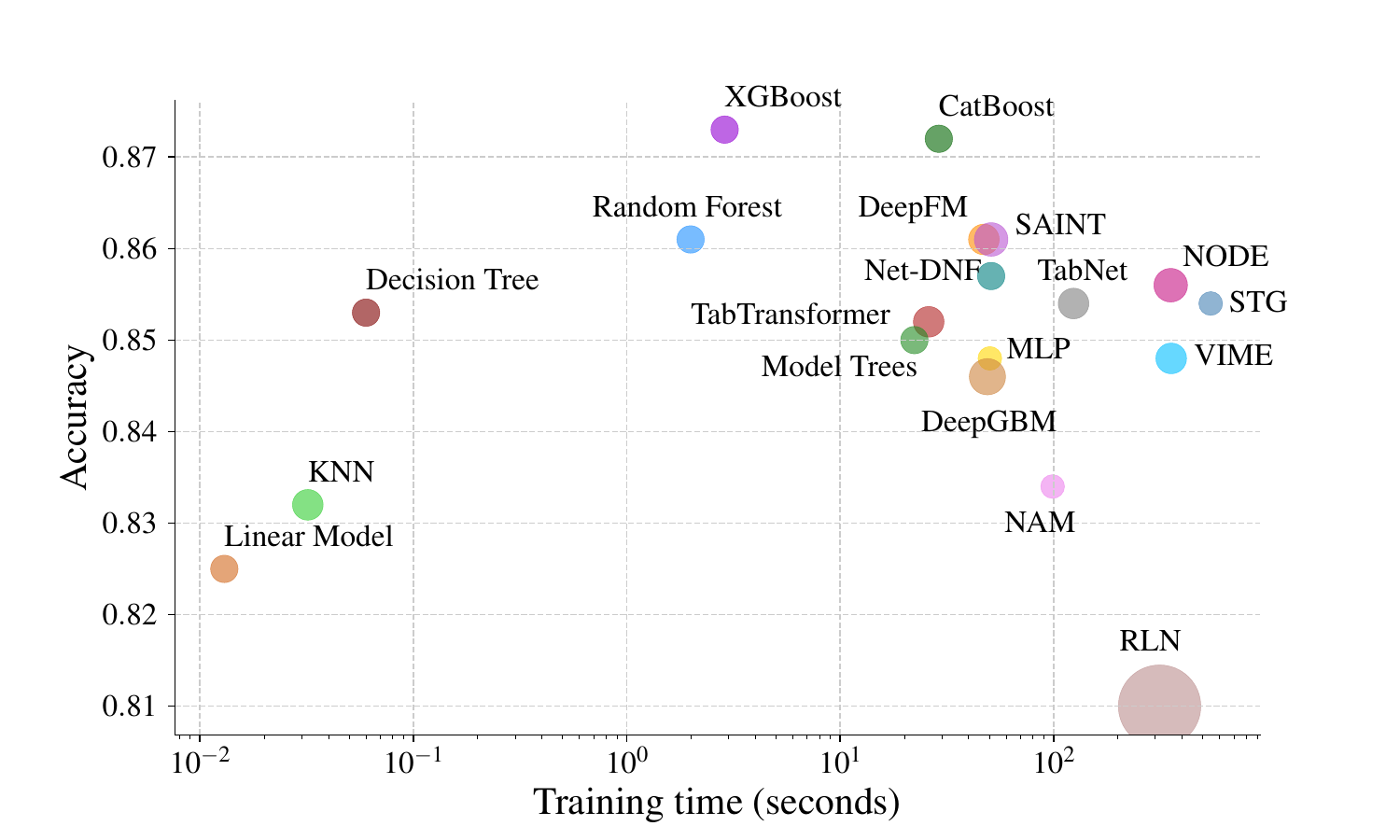}
     }
     \subfloat{
         \centering
        \includegraphics[width=\columnwidth]{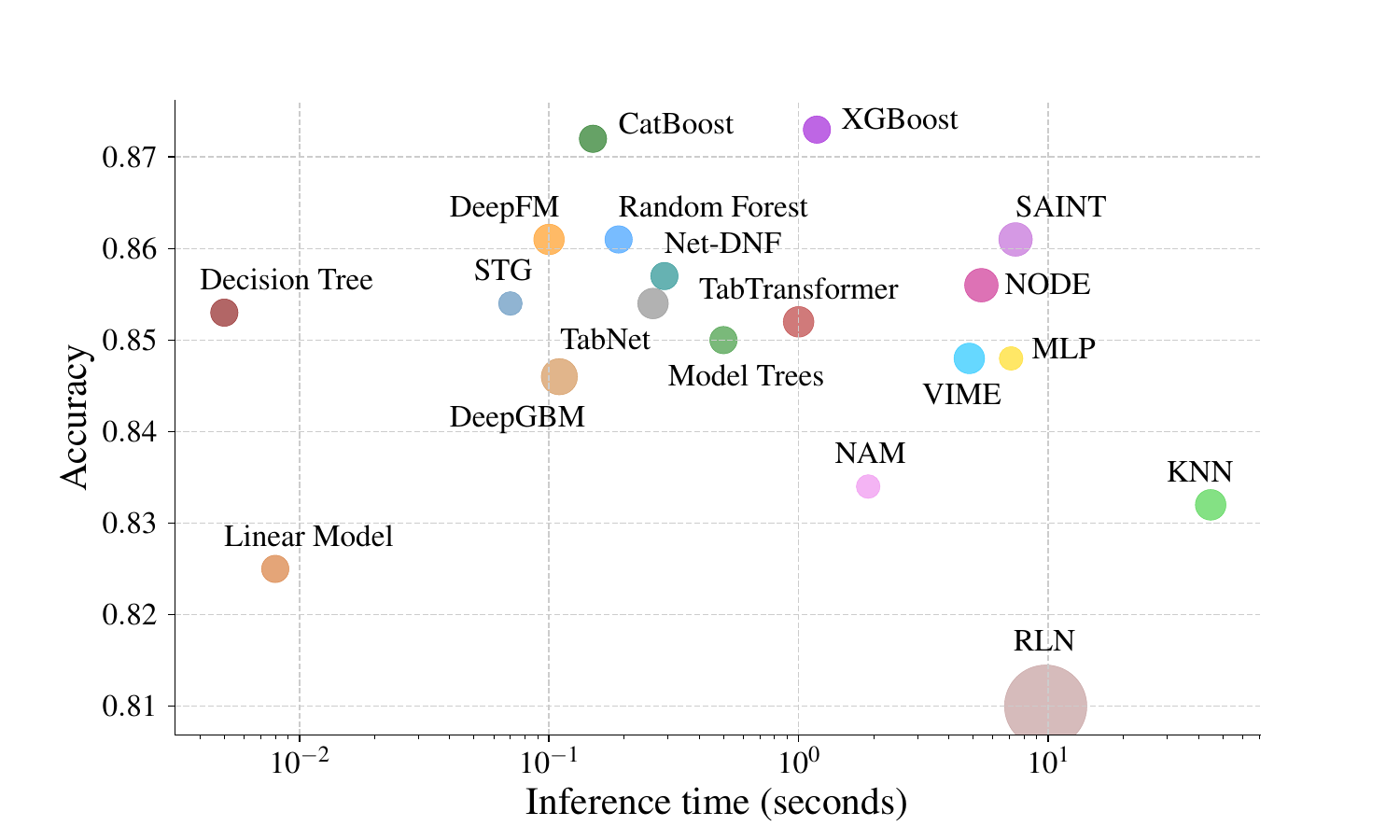}
     }
    \caption{\revised{Train (left) and inference (right) time benchmarks for selected methods on the Adult data set \added{with 32.561 samples}. The circle size reflects the accuracy standard deviation.}}
    \label{fig:adult_time_results}
\end{figure*}

\begin{figure*}[t!]
     \centering
     \subfloat{
         \centering
        \includegraphics[width=\columnwidth]{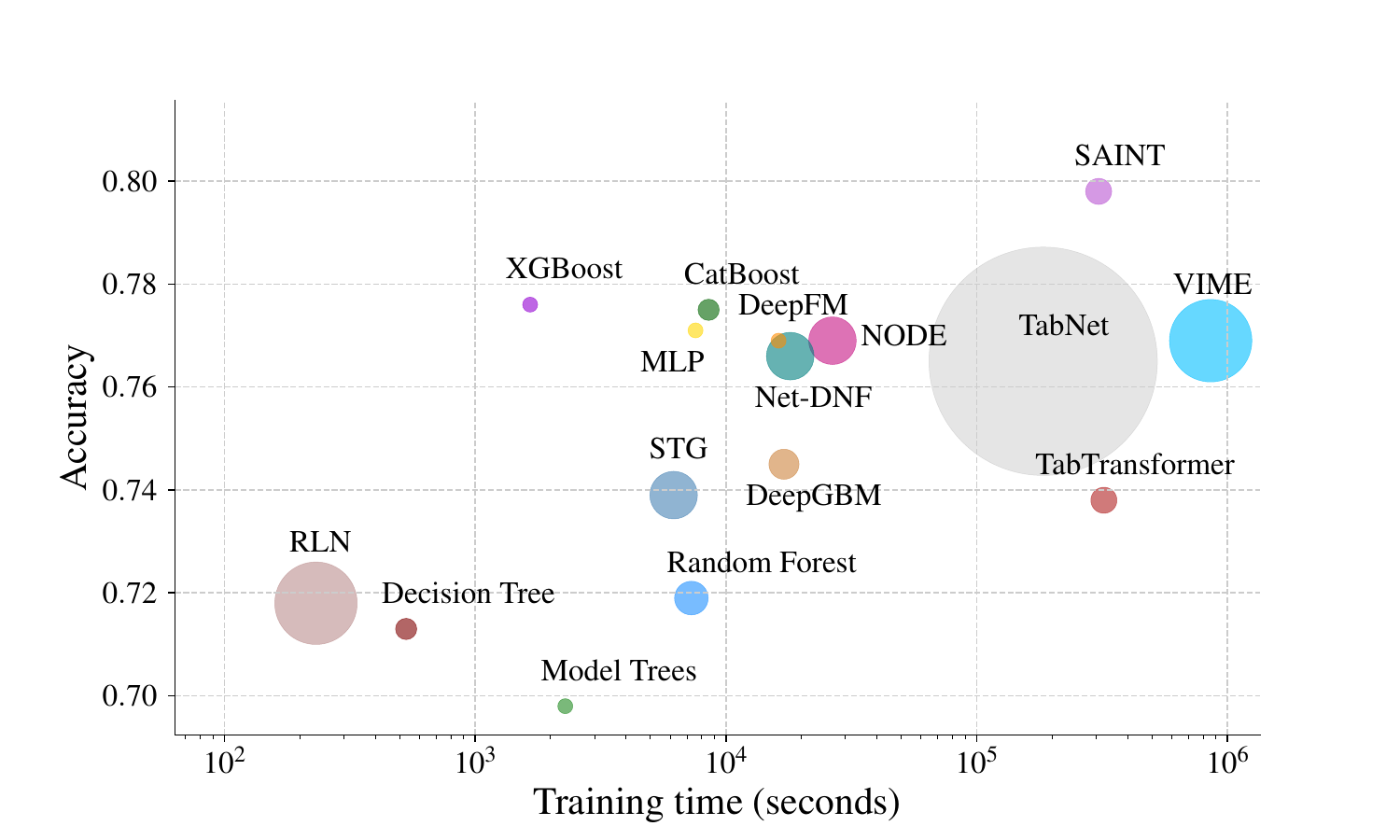}
     }
     \subfloat{
         \centering
        \includegraphics[width=\columnwidth]{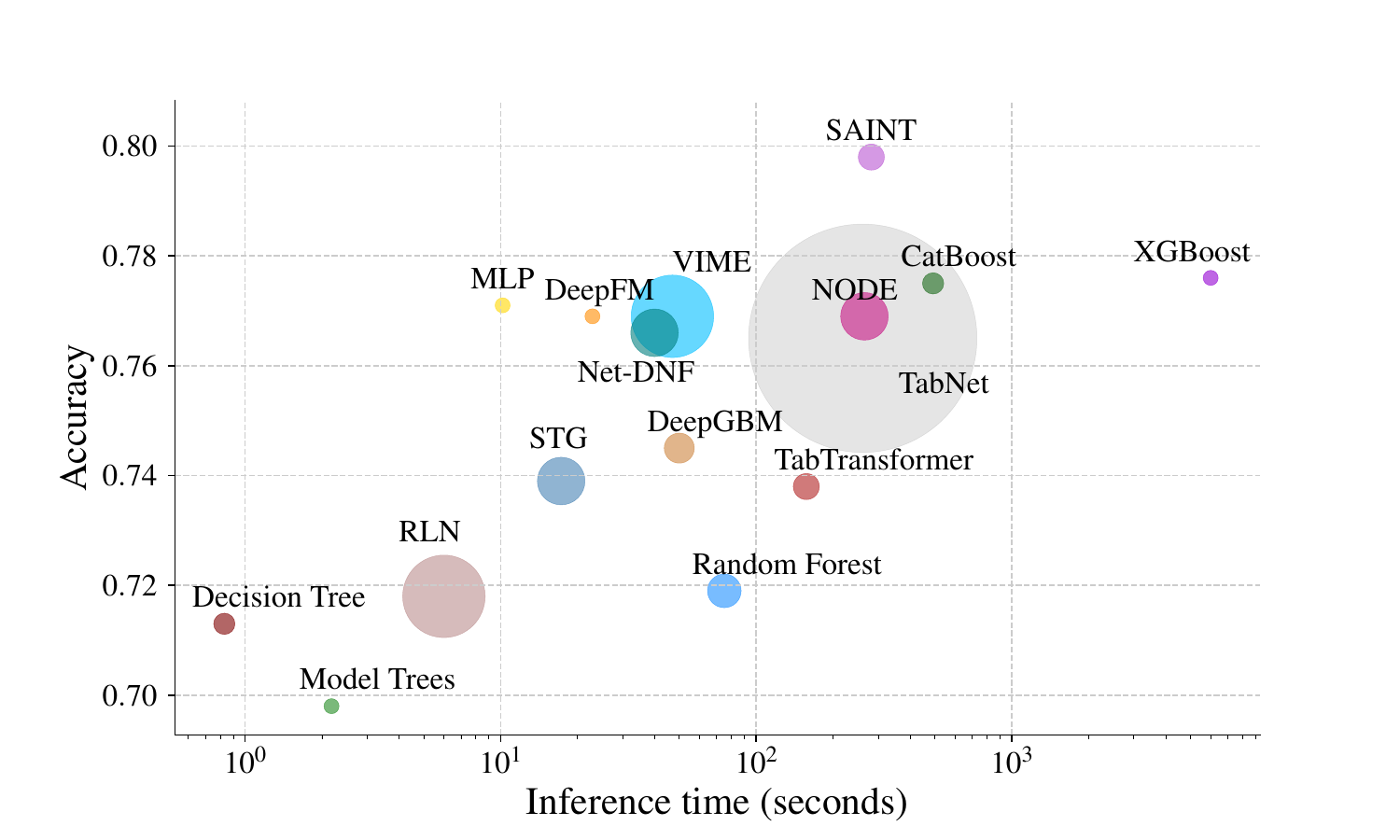}
     }
    \caption{\added{Train (left) and inference (right) time benchmarks for selected methods on the HIGGS data set with 11 million samples. The circle size reflects the accuracy standard deviation.}}
    \label{fig:higgs_time_results}
\end{figure*}

We also analyse the training and inference time of the models in comparison to their performance. \revised{We plot the time-performance characteristic for the models in Fig. \ref{fig:adult_time_results} and Fig. \ref{fig:higgs_time_results} for the Adult and the HIGGS data set respectively.} \added{While the training time of gradient-boosting-based models is lower than that of most deep neural network-based methods, their inference time on the HIGGS data set with 11 million samples is significantly higher: for XGBoost, the inference time amounts to 5995 seconds whereas inference times for MLP and SAINT are 10.18 and 282 seconds respectively. 
%Training times for XGBoost, MLP, and SAINTS models are 1658 seconds, 7568 seconds, and 306602 seconds. 
%These results indicate that deep neural networks are possibly more suitable for high-load machine learning systems. 
All gradient-boosting and deep learning models were trained on the same GPU.
}

%Models with optimal training-time-to-performance characteristics (there is no model with lower training time for a given performance) include the Linear Model, KNN, the Decision Tree, the Random Forest, and XGBoost.
%For the inference-time-to-performance ratio, Decision Trees, DeepFM, CatBoost and LightGBM have optimal characteristics. 

%% will regenerate the plots,
\begin{figure}[h]
    \centering
    \includegraphics[width=1\linewidth]{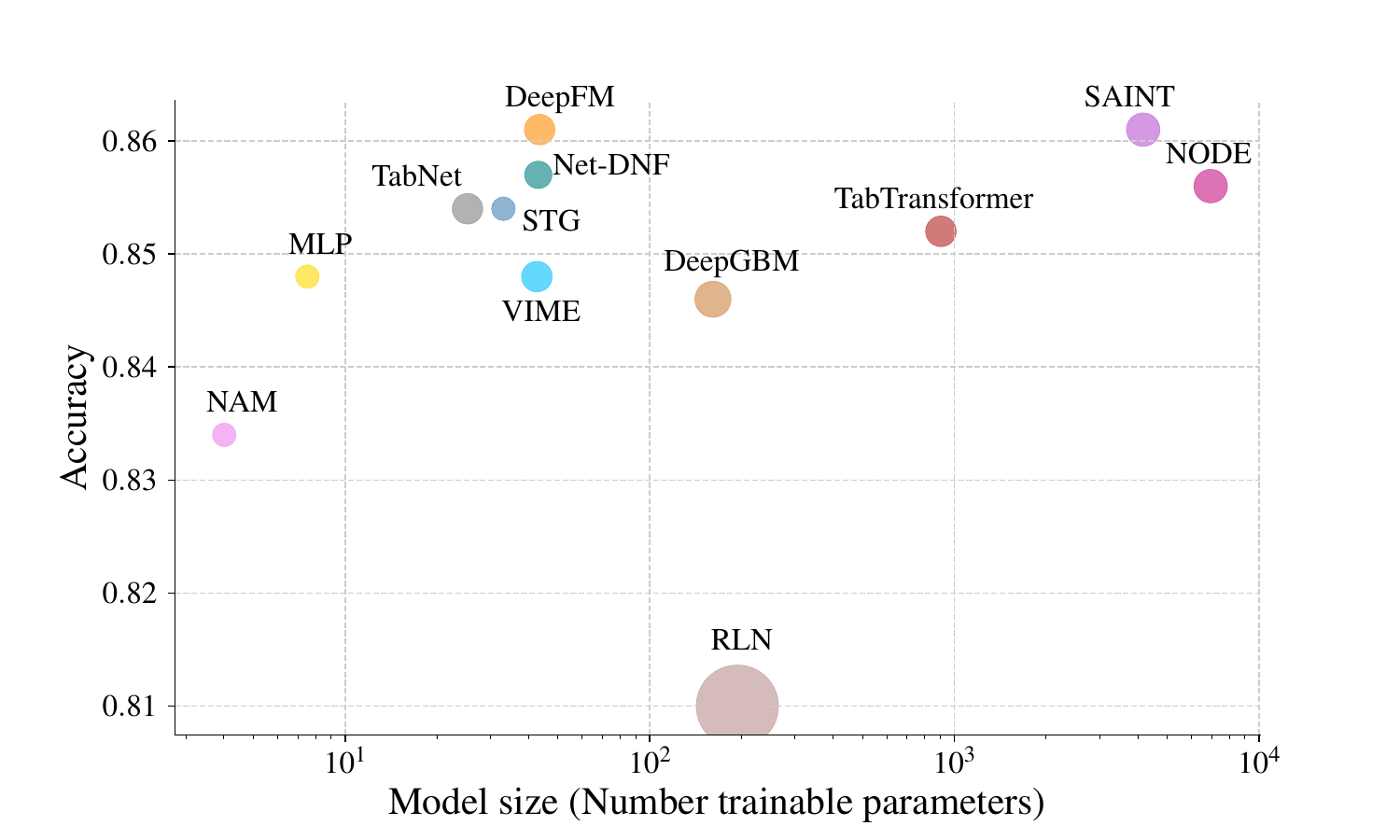}
    \caption{A size comparison of deep learning models on the Adult data set. The circle size reflects standard deviation.}
    \label{fig:models_size}
\end{figure}

%\subsection{Reproducibility Details}

% data preprocessing 
% framework 
%%%%%%%%%%%

\subsection{Interpretability Assessment}
\label{sec:interpretbench}
As opposed to the pure on-task performance, interpretability of the models is becoming an increasingly important characteristic. Therefore, we end this section with a distinct assessment of the interpretability properties claimed by some methods. \revised{The model size (number of parameters) can provide a first intuition of the interpretability of the models. Therefore, we provide a size comparison of deep learning models in Fig. \ref{fig:models_size}}.

Admittedly, explanations can be provided in very different forms, which may each have their own use-cases. Hence, we can only compare explanations that have a common form. In this work, we chose feature attributions as the explanation format because they are the prevalent form of post-hoc explainability for the models considered in this work. Remarkably, the models that build on the transformer architecture (Section \ref{sec:transformer_based}) often claim some extent of interpretability through the attention maps \citep{somepalli2021saint}. To verify this hypothesis and assess the attribution provided by some of the frameworks in practice, we run an ablation test with the features that were attributed the highest importance over all samples. Furthermore, due to the lack of ground truth attribution values, we compare individual attributions to the well-known KernelSHAP values \cite{lundberg2017unified}.

Evaluation of the quality of feature attribution is known to be a non-trivial problem \cite{rong2022evaluating}. \revised{We measure the fidelity
\cite{tomsett2020sanity} of the attributions by successively removing the features that have the highest mean importance assigned (Most Relevant First, MoRF \cite{tomsett2020sanity}). We then retrain the model on the reduced feature set. A sharp drop in predictive accuracy indicates that the discriminative features were successfully identified and removed.} We do the same for the inverse order, Least Relevant First (LeRF), which removes the features deemed unimportant. In this case, the accuracy should stay high as long as possible. For the attention maps of TabTransformer and SAINT, we either use the sum over the entire columns of the intra-feature attention maps as an importance estimate or only take the diagonal (feature self-attentions) as attributions. 

The obtained curves are visualized in Fig. \ref{fig:globalattrbench}. For the MoRF order, TabNet and TabTransformer with the diagonal of the attention head as attributions seem to perform best. For LeRF, TabNet is the only significantly better method than the others. For TabTransformer, taking the diagonal of the attention matrix seems to increase the performance, whereas for SAINT, there is almost no difference.
We additionally compare the attribution values obtained to values from the KernelSHAP attribution method. Unfortunately, there are no ground truth attributions to compare with. However, the SHAP framework has a solid grounding in game theory and is widely deployed \cite{sahakyan2021explainable}. We only compare the absolute values of the attributions, as the attention maps are constrained to be positive. As a measure of agreement, we compute the Spearman Rank Correlation between the attributions by the SHAP framework and the tabular data models. The correlation we observe is surprisingly low across all models, and sometimes it is even negative, which means that a higher SHAP attribution will probably result in a lower attribution by the model.

In these two simple benchmarks, the transformer models were not able to produce convincing feature attributions out-of-the-box. We come to the conclusion that more profound benchmarks of the claimed interpretability characteristics and their usefulness in practice are necessary.
\newcommand{\wstdsh}[2]{#1 $\pm$ #2}

\ifdefined\ieeeversion
\begin{figure}[t!]
     \centering
     \subfloat[Most Relevant First (MoRF)]{
         \centering
         \includegraphics[width=0.85\linewidth]{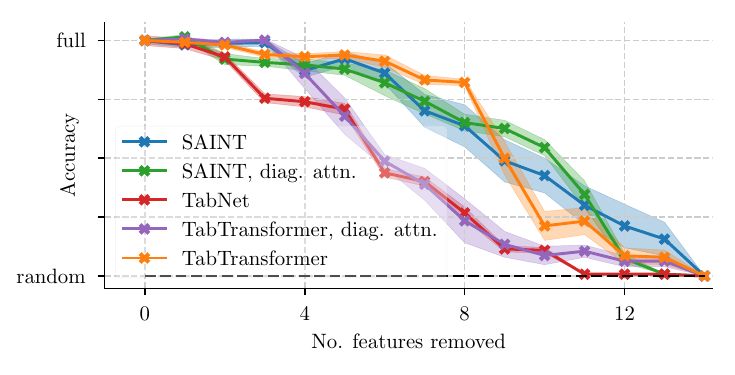}
     }\\
     \subfloat[Least Relevant First (LeRF)]{
         \centering
         \includegraphics[width=0.85\linewidth]{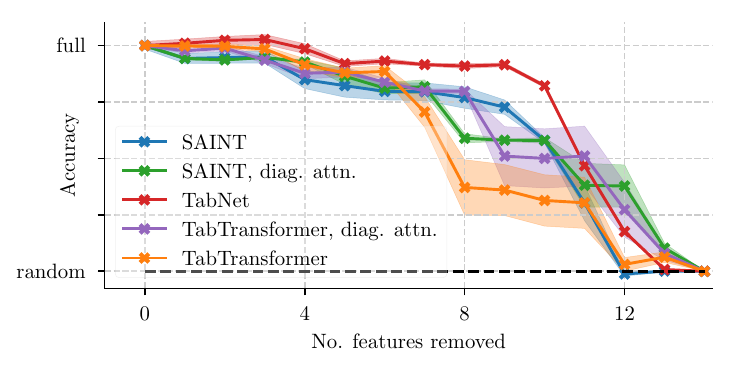}
     }
    \caption{Resulting curves of the global attribution benchmark for feature attributions (15 runs on Adult). Standard errors are indicated by the shaded area. For the MoRF order, an early drop in accuracy is desirable, while for LeRF, the accuracy should stay as high as possible.}
    \label{fig:globalattrbench}
\end{figure}
\fi

\begin{table}[]
\centering
{
% Setting the horizontal and vertical space
\renewcommand{\arraystretch}{0.8}
\setlength{\tabcolsep}{8pt}
    
    \begin{tabular}{@{}lr@{}}  %{m{4cm}cl} %{@{}l@{}r@{}r@{}r@{}r@{}r@{}}}
        \toprule
        Model, attention used & Spearman Corr.\\
        \midrule
        TabTransformer, columnw. attention & \wstdsh{-0.01}{0.008}\\
        TabTransformer, diag. attention & \wstdsh{0.00}{0.010}\\
        TabNet & \wstdsh{\textbf{0.07}}{0.009}\\
        SAINT, columnw. attention & \wstdsh{-0.04}{0.007}\\
        SAINT, diag. attention & \wstdsh{0.01}{0.007}\\
        \bottomrule
    \end{tabular}
}
    
    \caption{Spearman rank correlation of the provided attribution with KernelSHAP values as ground truth. Results were computed on 750 random samples from the Adult data set.}
    \label{tab:tabular_datasets}
    
\end{table}

%} % end of the new Section 

%%%%%%% Discussion %%%%%%%
\section{Discussion and Future Prospects}
\label{sec:discussion}

In this section, we summarize our findings and discuss current and future trends in deep learning approaches for tabular data (Section \ref{sec:discussionsummary}). Moreover, we identify several open research questions that could be tackled to advance the field of tabular deep neural networks (Section \ref{sec:open_questions}).

\subsection{Summary and Trends}
\label{sec:discussionsummary}

\textbf{Decision Tree Ensembles are still State-of-the-Art.}
In a fair comparison on multiple data sets, we demonstrated that models based on tree-ensembles, such as XGBoost, LightGBM, and CatBoost, still outperform the deep learning models on \added{most} data sets that we considered and come with the additional advantage of significantly less training time. Even though it has been six years since the XGBoost publication~\cite{XGBOOST} and over twenty years since the publishing of original gradient boosting paper~\cite{friedman2002stochastic}, we can state that despite much research effort in deep learning, the state of the art for tabular data remains largely unchanged. \added{However, we observed that for very large data sets, approaches based on deep learning may still be able to achieve competitive performance and even outperform classical models.}
In summary, we think that a fundamental reorientation of the domain may be necessary. \revised{For now, the question of whether the use of current deep learning techniques is beneficial for tabular data can generally be answered in the negative. This applies in particular to small heterogeneous data sets that are common in applications. Hence, instead of proposing more and more complex models, we argue that a more profound understanding of the reasons for this performance gap is needed.} 

\textbf{Unified Benchmarking}. Furthermore, our results highlight the need for unified benchmarks. There is no consensus in the machine learning community on how to make a fair and efficient comparison. 
%\cite{Kadra2021reg} uses about 40 different tabular data sets for assessing the predictive performance of machine learning algorithms, 
Shwartz-Ziv \& Armon \cite{Shwartz-Ziv2021} show that the choice of benchmarking data sets can have a non-negligible impact on the performance assessment. While we chose common data sets with varying characteristics for our experiments, a different choice of data sets or hyperparameter such as the encoding use (e.g., use one-hot encoding) may lead to a different outcome. Because of the excessive number of data sets \added{(in the eighteen works listed in Table~\ref{tab:overview_methods}, over 100 different data sets are used)}, there is a necessity for a standardized benchmarking procedure, which allows to identify significant progress with respect to the state of the art. With this work, we also propose an open-source benchmark for deep learning models on tabular data.
For \textit{tabular data generation} tasks, Xu et al. \cite{Xu2019CTGAN} proposes a sound evaluation framework with artificial and real-world data sets (Sec. \ref{sec:gan_quality}), but researchers need to agree on common benchmarks in this subdomain as well.

\textbf{Tabular Data Preprocessing.} Many of the challenges for deep neural networks on tabular data are related to the heterogeneity of the data (e.g., categorical and sparse values). Therefore, some deep learning solutions transform them into a homogeneous representation more suitable to neural networks. While the additional overhead is small, such transforms can boost performance considerably and should thus be among the first strategies applied in real-world scenarios.
%and data quality issues (missing data records, outliers, data set sizes), please refer to Section~\ref{sec:maintabulardata}. To overcome it, some deep learning solutions try to reduce the aforementioned effect by transforming heterogeneous data into homogeneous data, improving performance. 

\textbf{Architectures for Deep Learning on Tabular Data}.
Architecture-wise, there has been a clear trend towards transformer-based solutions (Section \ref{sec:transformer_based}) in recent years. These approaches offer multiple advantages over standard neural network architectures, for instance, learning with attention over both categorical and numerical features. Moreover, self-supervised or unsupervised pre-training that leverages unlabelled tabular data to train parts of the deep learning model is gaining popularity, not only among transformer-based approaches. 
Performance-wise, multiple independent evaluations demonstrate that deep neural network methods from the hybrid (Sec.~\ref{sec:hybrid_models}) and transformers-based (Sec. \ref{sec:transformer_based}) groups exhibit superior predictive performance compared to plain deep neural networks on various data sets~\citep{gorishniy2021revisiting, Ke2019tabnn, ke2019deepgbm, somepalli2021saint}. This underlines the importance of special-purpose architectures for tabular data.

\textbf{Regularization Models for Tabular Data}.
It has also been shown that regularization reduces the hypersensitivity of deep neural network models and improves the overall performance \citep{shavitt2018regularization, Kadra2021reg}. We believe that regularization is one of the crucial aspects for a more robust and accurate performance of deep neural networks on tabular data and is gaining momentum.

\textbf{Deep Generative Models for Tabular Data.} Powerful tabular data generation is essential for the development of high-quality models, particularly in a privacy context. With suitable data generators at hand, developers can use large, synthetic, and yet realistic data sets to develop better models, while not being subject to privacy concerns \citep{jordon2018pate}. Unfortunately, the generation task is as hard as inference in predictive models, so progress in both areas will likely go hand in hand. 

\textbf{Interpretable Deep Learning Models for Tabular Data.}
Interpretability is undoubtedly desirable, particularly for tabular data models frequently applied to personal data, e.g., in healthcare and finance. An increasing number of approaches offer it out-of-the-box but most current deep neural network models are still mainly concerned with the optimization of a chosen error metric. Therefore, extending existing open-source libraries \added{(see \citep{pawelczyk2021carla,liu2021synthetic})} aimed at interpreting black-box models helps advance the field. Moreover, interpretable deep tabular learning is essential for understanding model decisions and results, especially for life-critical applications. However, much of the state-of-the-art recourse literature does not offer easy support of heterogeneous tabular data and lacks metrics to evaluate the quality of heterogeneous data recourse. Finally, model explanations can also be used to identify and mitigate potential bias or eliminate unfair discrimination against certain groups \citep{ntoutsi2020bias}. 

\textbf{Learning From Evolving Data Streams.}  Many modern applications are subject to continuously evolving data streams, e.g., social media, online retail, or healthcare. Streaming data are usually heterogeneous and potentially unlimited. Therefore, observations must be processed in a single pass and cannot be stored. Indeed, online learning models can only access a fraction of the data at each time step. Furthermore, they have to deal with limited resources and shifting data distributions (i.e., concept drift). Hence, hyperparameter optimization and model selection, as typically involved in deep learning, are usually not feasible in a data stream. For this reason, despite the success of deep learning in other domains, less complex methods such as incremental decision trees \citep{domingos2000mining,manapragada2018extremely} are often preferred in online learning applications.

% \textbf{Is there one algorithm „to rule them all“?} 
\subsection{Open Research Questions}\label{sec:open_questions}
Several open problems need to be addressed in future research. In this section, we will list those we deem fundamental to the domain.

\textbf{Information-theoretic Analysis of Encodings.} Encoding methods are highly popular when dealing with tabular data. However, the majority of data preprocessing approaches for deep neural networks are lossy in terms of information content. Therefore, it is challenging to achieve an efficient, almost lossless transformation of heterogeneous tabular data into homogeneous data. Nevertheless, the information-theoretic view on these transformations remains to be investigated in detail and could shed light on the underlying mechanisms.

\textbf{Computational Efficiency in Hybrid Models.} The work by Shwartz-Ziv \& Armon \cite{Shwartz-Ziv2021} suggests that the combination of a gradient boosting decision tree and deep neural networks may improve the predictive performance of a machine learning system. However, it also leads to growing complexity. Training or inference times, which far exceed those of classical machine learning approaches, are a recurring problem when developing hybrid models. We conclude that the integration of state-of-the-art approaches from classical machine learning and deep learning has not been conclusively resolved yet and future work should be conducted on how to mitigate the trade-off between predictive performance and computational complexity.

\textbf{Specialized Regularizations.} We applaud recent research on regularization methods, in which we see a promising direction that necessitates further exploration. Whether context- and architecture-specific regularizations for tabular data can be found remains an open question. \added{However, a recent work \cite{balestriero2022effects} indicates that regularization techniques for deep neural networks such as weight decay and data augmentation produce an unfair model across classes.} Additionally, it is relevant to explore the theoretical constraints that govern the success of regularization on tabular data more profoundly. 

\textbf{Novel Processes for Tabular Data Generation.} For tabular data generation, modified Generative Adversarial Networks and Variational Autoencoders are prevalent. However, the modelling of dependencies and categorical distributions remains the key challenge. Novel architectures in this area, such as diffusion models, have not been adapted to the domain of tabular data. Furthermore, the definition of an entirely new generative process particularly focused on tabular data might be worth investigating.

\textbf{Interpretability.} Going forward, counterfactual explanations for deep tabular learning can be used to improve the perceived fairness in human-AI interaction scenarios and to enable personalized decision-making \citep{verma2020counterfactual}. 
However, the heterogeneity of tabular data poses problems for counterfactual explanation methods to be reliably deployed in practice. Devising techniques aimed at effectively handling heterogeneous tabular data in the presence of feasibility constraints is still an unsolved task \citep{pawelczyk2021carla}.
%Many modern and promising approaches follow the interpretability-by-design principle (Table \ref{tab:overview_methods}).

\textbf{Transfer of Deep Learning Methods to Data Streams.} Recent work shows that some of the limitations of neural networks in an evolving data stream can be overcome \citep{sahoo2017online,duda2020training}. Conversely, changes in the parameters of a neural network may be effectively used to weigh the importance of input features over time \citep{haug2020leveraging} or to detect concept drift \citep{haug2021learning}. Accordingly, we argue that deep learning for streaming data -- in particular strategies for dealing with evolving and heterogeneous tabular data -- should receive more attention in the future.

\textbf{Transfer Learning for Tabular Data.} Reusing knowledge gained solving one problem and applying it to a different task is the research problem addressed by transfer learning. While transfer learning is successfully used in computer vision and natural language processing applications \citep{tan2018survey}, there are no efficient and generally accepted ways to do transfer learning for tabular data. Hence, a general research question can be how to share knowledge between multiple (related) tabular data sets efficiently.

\textbf{Data Augmentation for Tabular Data.} Data augmentation has proven highly effective to prevent overfitting, especially in computer vision \citep{shorten2019surveyaugmentation}. While some data augmentation techniques for tabular data exist, e.g., SMOTE-NC \citep{chawla2002smote}, simple models fail to capture the dependency structure of the data. Therefore, generating additional samples in a continuous latent space is a promising direction. This was investigated by Darabi \& Elor \cite{darabi2021synthesising} for minority oversampling. Nevertheless, the reported improvements are only marginal. Thus, future work is required to find simple, yet effective random transformations to enhance tabular training sets.

\textbf{Self-supervised Learning.} Large-scale labelled data are usually required to train deep neural networks; however, the data labelling is an expensive task. To avoid this expensive step, self-supervised methods propose to learn general feature representations from available unlabelled data. These methods have also shown astonishing results in computer vision and natural language processing  \citep{jing2020self, liu2021self}. Only a few recent works in this direction \citep{yoon2020vime, bahri2021scarf, ucar2021subtab} deal with heterogeneous data. Hence, novel self-supervised learning approaches dedicated to tabular data might be worth investigating.

%%%%%%%%%%%%%%%%%%%%%%%%%%%%%%%%%%
%%%%%%%%%%%%%%%%%%%%%%%%%%%%%%%%%%
%%%%%%%%%%%%%%%%%%%%%%%%%%%%%%%%%%
\section{Conclusion}
\label{sec:conclusion}
This survey is the first work to systematically explore deep neural network approaches for heterogeneous tabular data. In this context, we highlighted the main challenges and research advances in modelling, generating, and explaining tabular data. 
%We provided empirical evidence that decision tree ensembles still represent the state of the art for learning with \added{small (less then 1M samples)}tabular data sets.
We introduced a unified taxonomy that categorizes deep learning approaches for tabular data into three branches: data transformation methods, specialized architectures, and regularization models. 
We believe our taxonomy will help catalogue future research and better understand and address the remaining challenges in applying deep learning to tabular data. We hope it will help researchers and practitioners to find the most appropriate strategies and methods for their applications.

Additionally, we also conducted an unbiased evaluation of the state-of-the-art deep learning approaches on multiple real-world data sets. Deep neural network-based methods for heterogeneous tabular data are still inferior to machine learning methods based on decision tree ensembles for small and medium-sized data sets \added{(less than $\sim$1M samples).} \added{Only for a very large data set mainly consisting of continuous and numerical variables, the deep learning model SAINT outperformed these classical approaches.}
Furthermore, we assessed explanation properties of deep learning models with the self-attention mechanism. Although the TabNet model shows promising explanation explanatory capabilities, inconsistencies between the explanations remain an open issue.

%Additionally, we also discussed the present solutions for tabular data generation using neural networks and named open problems and important future research directions. 

Due to the importance of tabular data to industry and academia, new ideas in this area are in high demand and can have a significant impact. With this review, we hope to provide interested readers with the references and insights they need to address open challenges and effectively advance the field.

%% file: images/taxonomy.tex
\tikzset{every picture/.style={line width=0.75pt}} %set default line width to 0.75pt        
\scalebox{0.8}{
\begin{tikzpicture}[x=0.75pt,y=0.75pt,yscale=-1,xscale=1]
%uncomment if require: \path (0,549); %set diagram left start at 0, and has height of 549

%Shape: Rectangle [id:dp23376225133769513] 
\draw  [fill={rgb, 255:red, 184; green, 233; blue, 134 }  ,fill opacity=0.59 ] (10,28) -- (286.5,28) -- (286.5,61.27) -- (10,61.27) -- cycle ;
%Straight Lines [id:da11524719931091665] 
\draw    (26,62) -- (26,486) -- (26,514.86) ;
%Straight Lines [id:da668437037901723] 
\draw    (26.11,94.63) -- (75.37,94.63) ;
%Shape: Rectangle [id:dp60247793872065] 
\draw  [fill={rgb, 255:red, 74; green, 144; blue, 226 }  ,fill opacity=0.42 ] (75.37,77.72) -- (313.5,77.72) -- (313.5,110.78) -- (75.37,110.78) -- cycle ;
%Straight Lines [id:da9744426476940199] 
\draw    (109.71,111.55) -- (109.46,185.34) ;
%Straight Lines [id:da7053973369685859] 
\draw    (109.45,140.33) -- (143.41,140.72) ;
%Shape: Rectangle [id:dp9266613360635536] 
\draw  [fill={rgb, 255:red, 74; green, 144; blue, 226 }  ,fill opacity=0.42 ] (142.85,124.77) -- (373.5,124.77) -- (373.5,155.52) -- (142.85,155.52) -- cycle ;
%Straight Lines [id:da5775820527957471] 
\draw    (109.46,185.34) -- (126.93,185.34) -- (143.41,185.3) ;
%Straight Lines [id:da19923269144773303] 
\draw    (26.11,252.22) -- (75.37,252.22) ;
%Shape: Rectangle [id:dp9549657985533372] 
\draw  [fill={rgb, 255:red, 248; green, 231; blue, 28 }  ,fill opacity=0.44 ] (74.92,235.31) -- (278.5,235.31) -- (278.5,268.36) -- (74.92,268.36) -- cycle ;
%Straight Lines [id:da5246493939630164] 
\draw    (202.31,316.17) -- (202.76,397.76) ;
%Straight Lines [id:da09670297892514668] 
\draw    (202.76,353.17) -- (241.13,353.07) ;
%Shape: Rectangle [id:dp4658657638396484] 
\draw  [fill={rgb, 255:red, 248; green, 231; blue, 28 }  ,fill opacity=0.44 ] (241.6,337.19) -- (392.5,337.19) -- (392.5,367.93) -- (241.6,367.93) -- cycle ;
%Straight Lines [id:da5005067168814944] 
\draw    (202.76,397.76) -- (241.13,397.66) ;
%Shape: Boxed Line [id:dp44063969955833215] 
\draw    (110.75,300.29) -- (143.41,299.81) ;
%Shape: Rectangle [id:dp8501234969676266] 
\draw  [fill={rgb, 255:red, 248; green, 231; blue, 28 }  ,fill opacity=0.44 ] (143.35,436.19) -- (359.5,436.19) -- (359.5,469.25) -- (143.35,469.25) -- cycle ;
%Shape: Rectangle [id:dp08653706414728246] 
\draw  [fill={rgb, 255:red, 208; green, 2; blue, 27 }  ,fill opacity=0.2 ] (77.66,497.95) -- (264.52,497.95) -- (264.52,531) -- (77.66,531) -- cycle ;
%Straight Lines [id:da7181422912591837] 
\draw    (26.11,514.86) -- (51.54,515.06) -- (76.42,515.27) ;
%Shape: Rectangle [id:dp456850885777484] 
\draw  [fill={rgb, 255:red, 248; green, 231; blue, 28 }  ,fill opacity=0.44 ] (142.85,286.19) -- (263.77,286.19) -- (263.77,316.94) -- (142.85,316.94) -- cycle ;
%Straight Lines [id:da4640696217302922] 
\draw    (110.01,266.98) -- (111.75,451.98) ;
%Shape: Rectangle [id:dp9373772761836863] 
\draw  [fill={rgb, 255:red, 248; green, 231; blue, 28 }  ,fill opacity=0.44 ] (240.88,382.28) -- (393.5,382.28) -- (393.5,413.03) -- (240.88,413.03) -- cycle ;
%Straight Lines [id:da28724453328052935] 
\draw    (111.75,451.98) -- (143.1,451.98) ;
%Shape: Rectangle [id:dp7233629356468493] 
\draw  [fill={rgb, 255:red, 74; green, 144; blue, 226 }  ,fill opacity=0.42 ] (142.85,168.77) -- (373.5,168.77) -- (373.5,199.52) -- (142.85,199.52) -- cycle ;

% Text Node
\draw (160,445) node [anchor=north west][inner sep=0.75pt]  [font=\large] [align=left] {{Transformer-based Models}};
% Text Node
\draw (135,44.64) node  [font=\large] [align=left] {\begin{minipage}[lt]{156.51pt}\setlength\topsep{0pt}
\begin{center}
{Deep Learning for Tabular Data}
\end{center}

\end{minipage}};
% Text Node
\draw (151.17,132) node [anchor=north west][inner sep=0.75pt]  [font=\large] [align=left] {{Single-Dimensional Encoding}};
% Text Node
\draw (150.31,291.07) node [anchor=north west][inner sep=0.75pt]  [font=\large] [align=left] {{Hybrid Models}};
% Text Node
\draw (194.44,94.25) node  [font=\large] [align=left] {{Data Transformation Methods}};
% Text Node
\draw (247.19,387.16) node [anchor=north west][inner sep=0.75pt]  [font=\large] [align=left] {{Partly differentiable}};
% Text Node
\draw (156.67,175) node [anchor=north west][inner sep=0.75pt]  [font=\large] [align=left] {{Multi-Dimensional Encoding}};
% Text Node
\draw (87.59,503.97) node [anchor=north west][inner sep=0.75pt]  [font=\large] [align=left] {{Regularization Models }};
% Text Node
\draw (250.55,342.06) node [anchor=north west][inner sep=0.75pt]  [font=\large] [align=left] {{Fully differentiable}};
% Text Node
\draw (176.71,251.83) node  [font=\large] [align=left] {{Specialized Architectures}};

\end{tikzpicture}}

%% file: experiment_results/results5.tex
Linear Model & \wstd{73.0}{0.0} & \wstd{80.1}{0.1} & \wstd{82.5}{0.2} & \wstd{85.4}{0.2} & \wstd{64.1}{0.0} & \wstd{68.4}{0.0} & \wstd{72.4}{0.0} & \wstd{92.8}{0.0} & \wstd{0.528}{0.008} \\

KNN~\cite{fix1951discriminatory} & \wstd{72.2}{0.0} & \wstd{79.0}{0.1} & \wstd{83.2}{0.2} & \wstd{87.5}{0.2} & \wstd{62.3}{0.1} & \wstd{67.1}{0.0} & \wstd{70.2}{0.1} & \wstd{90.1}{0.2} & \wstd{0.421}{0.009} \\

% SVM & - & - & - & - & - & - & - & - & - \\

Decision Tree~\cite{breiman2017classification} & \wstd{80.3}{0.0} & \wstd{89.3}{0.1} & \wstd{85.3}{0.2} & \wstd{89.8}{0.1} & \wstd{71.3}{0.0} & \wstd{78.7}{0.0} & \wstd{79.1}{0.0} & \wstd{95.0}{0.0} & \wstd{0.404}{0.007} \\

Random Forest~\cite{breiman2001random} & \wstd{82.1}{0.2} & \wstd{90.0}{0.2} & \wstd{86.1}{0.2} & \wstd{91.7}{0.2} & \wstd{71.9}{0.0} & \wstd{79.7}{0.0} & \wstd{78.1}{0.1} & \wstd{96.1}{0.0} & \wstd{0.272}{0.006} \\

XGBoost~\cite{XGBOOST} & \underline{\wstd{83.5}{0.2}} & \wstd{92.2}{0.0} & \underline{\wstd{87.3}{0.2}} & \underline{\wstd{92.8}{0.1}} & \underline{\wstd{77.6}{0.0}} & \underline{\wstd{85.9}{0.0}} & \textbf{\wstd{97.3}{0.0}} & \textbf{\wstd{99.9}{0.0}} & \wstd{0.206}{0.005} \\

LightGBM~\cite{ke2017lightgbm} & \underline{\wstd{83.5}{0.1}} & \underline{\wstd{92.3}{0.0}} & \textbf{\wstd{87.4}{0.2}} & \textbf{\wstd{92.9}{0.1}} & \wstd{77.1}{0.0} & \wstd{85.5}{0.0} & \wstd{93.5}{0.0} & \wstd{99.7}{0.0} & \textbf{\wstd{0.195}{0.005}} \\

CatBoost~\cite{prokhorenkova2018catboost} & \textbf{\wstd{83.6}{0.3}} & \textbf{\wstd{92.4}{0.1}} & \wstd{87.2}{0.2} & \underline{\wstd{92.8}{0.1}} & \wstd{77.5}{0.0} & \wstd{85.8}{0.0} & \underline{\wstd{96.4}{0.0}} & \underline{\wstd{99.8}{0.0}} & \underline{\wstd{0.196}{0.004}} \\

Model Trees~\cite{broelemann2018gradient} & \wstd{82.6}{0.2} & \wstd{91.5}{0.0} & \wstd{85.0}{0.2} & \wstd{90.4}{0.1} & \wstd{69.8}{0.0} & \wstd{76.7}{0.0} & - & - & \wstd{0.385}{0.019} \\

\midrule

MLP~\cite{mcculloch1943logical} & \wstd{73.2}{0.3} & \wstd{80.3}{0.1} & \wstd{84.8}{0.1} & \wstd{90.3}{0.2} & \wstd{77.1}{0.0} & \wstd{85.6}{0.0} & \wstd{91.0}{0.4} & \wstd{76.1}{3.0} & \wstd{0.263}{0.008} \\

DeepFM~\cite{guo2017deepfm}  & \wstd{73.6}{0.2} & \wstd{80.4}{0.1} & \wstd{86.1}{0.2} & \wstd{91.7}{0.1} & \wstd{76.9}{0.0} & \wstd{83.4}{0.0} & - & - & \wstd{0.260}{0.006} \\

DeepGBM~\cite{ke2019deepgbm} & \wstd{78.0}{0.4} & \wstd{84.1}{0.1} & \wstd{84.6}{0.3} & \wstd{90.8}{0.1} & \wstd{74.5}{0.0} & \wstd{83.0}{0.0} & - & - & \wstd{0.856}{0.065} \\

RLN~\cite{shavitt2018regularization} & \wstd{73.2}{0.4} & \wstd{80.1}{0.4} & \wstd{81.0}{1.6} & \wstd{75.9}{8.2} & \wstd{71.8}{0.2} & \wstd{79.4}{0.2} & \wstd{77.2}{1.5} & \wstd{92.0}{0.9} & \wstd{0.348}{0.013} \\

TabNet~\cite{arik2019tabnet} & \wstd{81.0}{0.1} & \wstd{90.0}{0.1} & \wstd{85.4}{0.2} & \wstd{91.1}{0.1} & \wstd{76.5}{1.3} & \wstd{84.9}{1.4} & \wstd{93.1}{0.2} & \wstd{99.4}{0.0} & \wstd{0.346}{0.007} \\

VIME~\cite{yoon2020vime} & \wstd{72.7}{0.0} & \wstd{79.2}{0.0} & \wstd{84.8}{0.2} & \wstd{90.5}{0.2} & \wstd{76.9}{0.2} & \wstd{85.5}{0.1} & \wstd{90.9}{0.1} & \wstd{82.9}{0.7} & \wstd{0.275}{0.007} \\

TabTransformer~\cite{Huang2020tabtrans} & \wstd{73.3}{0.1} & \wstd{80.1}{0.2} & \wstd{85.2}{0.2} & \wstd{90.6}{0.2} & \wstd{73.8}{0.0} & \wstd{81.9}{0.0} & \wstd{76.5}{0.3} & \wstd{72.9}{2.3} & \wstd{0.451}{0.014} \\

NODE~\cite{popov2019neural} & \wstd{79.8}{0.2} & \wstd{87.5}{0.2} & \wstd{85.6}{0.3} & \wstd{91.1}{0.2} & \wstd{76.9}{0.1} & \wstd{85.4}{0.1} & \wstd{89.9}{0.1} & \wstd{98.7}{0.0} & \wstd{0.276}{0.005} \\

Net-DNF~\cite{katzir2021netdnf} & \wstd{82.6}{0.4} & \wstd{91.5}{0.2} & \wstd{85.7}{0.2} & \wstd{91.3}{0.1} & \wstd{76.6}{0.1} & \wstd{85.1}{0.1} & \wstd{94.2}{0.1} & \wstd{99.1}{0.0} & - \\

STG~\cite{icml2020_5085} & \wstd{73.1}{0.1} & \wstd{80.0}{0.1} & \wstd{85.4}{0.1} & \wstd{90.9}{0.1} & \wstd{73.9}{0.1} & \wstd{81.9}{0.1} & \wstd{81.8}{0.3} & \wstd{96.2}{0.0} & \wstd{0.285}{0.006} \\

NAM~\cite{agarwal2020neural}  & \wstd{73.3}{0.1} & \wstd{80.7}{0.3} & \wstd{83.4}{0.1} & \wstd{86.6}{0.1} & \wstd{53.9}{0.6} & \wstd{55.0}{1.2} & - & - & \wstd{0.725}{0.022} \\

SAINT~\cite{somepalli2021saint}  & \wstd{82.1}{0.3} & \wstd{90.7}{0.2} & \wstd{86.1}{0.3} & \wstd{91.6}{0.2} & \textbf{\wstd{79.8}{0.0}} & \textbf{\wstd{88.3}{0.0}} & \wstd{96.3}{0.1} & \underline{\wstd{99.8}{0.0}} & \wstd{0.226}{0.004}

%% file: tabdata_ieee.bbl
\begin{thebibliography}{100}
\providecommand{\url}[1]{#1}
\csname url@samestyle\endcsname
\providecommand{\newblock}{\relax}
\providecommand{\bibinfo}[2]{#2}
\providecommand{\BIBentrySTDinterwordspacing}{\spaceskip=0pt\relax}
\providecommand{\BIBentryALTinterwordstretchfactor}{4}
\providecommand{\BIBentryALTinterwordspacing}{\spaceskip=\fontdimen2\font plus
\BIBentryALTinterwordstretchfactor\fontdimen3\font minus
  \fontdimen4\font\relax}
\providecommand{\BIBforeignlanguage}[2]{{%
\expandafter\ifx\csname l@#1\endcsname\relax
\typeout{** WARNING: IEEEtran.bst: No hyphenation pattern has been}%
\typeout{** loaded for the language `#1'. Using the pattern for}%
\typeout{** the default language instead.}%
\else
\language=\csname l@#1\endcsname
\fi
#2}}
\providecommand{\BIBdecl}{\relax}
\BIBdecl

\bibitem{schmidhuber2015deep}
J.~Schmidhuber, ``Deep learning in neural networks: An overview,'' \emph{Neural
  networks}, vol.~61, pp. 85--117, 2015.

\bibitem{deeplearningbook}
I.~Goodfellow, Y.~Bengio, and A.~Courville, \emph{Deep learning}.\hskip 1em
  plus 0.5em minus 0.4em\relax MIT press, 2016.

\bibitem{hochreiter1997long}
S.~Hochreiter and J.~Schmidhuber, ``Long short-term memory,'' \emph{Neural
  computation}, vol.~9, no.~8, pp. 1735--1780, 1997.

\bibitem{vaswani2017attention}
A.~Vaswani, N.~Shazeer, N.~Parmar, J.~Uszkoreit, L.~Jones, A.~N. Gomez,
  {\L}.~Kaiser, and I.~Polosukhin, ``Attention is all you need,'' in
  \emph{Advances in neural information processing systems}, 2017, pp.
  5998--6008.

\bibitem{arik2019tabnet}
S.~O. Arik and T.~Pfister, ``\text{TabNet}: Attentive interpretable tabular
  learning,'' \emph{arxiv:1908.07442}, 2019.

\bibitem{popov2019neural}
S.~Popov, S.~Morozov, and A.~Babenko, ``Neural oblivious decision ensembles for
  deep learning on tabular data,'' \emph{arxiv:1909.06312}, 2019.

\bibitem{Shwartz-Ziv2021}
R.~Shwartz-Ziv and A.~Armon, ``{Tabular Data: Deep Learning is Not All You
  Need},'' \emph{arXiv preprint arXiv:2106.03253}, 2021.

\bibitem{elsayed2021we}
S.~Elsayed, D.~Thyssens, A.~Rashed, H.~S. Jomaa, and L.~Schmidt-Thieme, ``Do we
  really need deep learning models for time series forecasting?'' \emph{arXiv
  preprint arXiv:2101.02118}, 2021.

\bibitem{somepalli2021saint}
G.~Somepalli, M.~Goldblum, A.~Schwarzschild, C.~B. Bruss, and T.~Goldstein,
  ``\text{SAINT}: Improved neural networks for tabular data via row attention
  and contrastive pre-training,'' \emph{arXiv preprint arXiv:2106.01342}, 2021.

\bibitem{Kadra2021reg}
A.~Kadra, M.~Lindauer, F.~Hutter, and J.~Grabocka, ``{Well-tuned Simple Nets
  Excel on Tabular Datasets},'' in \emph{Advances in Neural Information
  Processing Systems}, 2021.

\bibitem{ulmer2020trust}
D.~Ulmer, L.~Meijerink, and G.~Cin{\`a}, ``Trust issues: Uncertainty estimation
  does not enable reliable ood detection on medical tabular data,'' in
  \emph{Machine Learning for Health}.\hskip 1em plus 0.5em minus 0.4em\relax
  PMLR, 2020, pp. 341--354.

\bibitem{somani2021deep}
S.~Somani, A.~J. Russak, F.~Richter, S.~Zhao, A.~Vaid, F.~Chaudhry, J.~K.
  De~Freitas, N.~Naik, R.~Miotto, G.~N. Nadkarni \emph{et~al.}, ``Deep learning
  and the electrocardiogram: review of the current state-of-the-art,'' \emph{EP
  Europace}, 2021.

\bibitem{borisov2021robust}
V.~Borisov, E.~Kasneci, and G.~Kasneci, ``Robust cognitive load detection from
  wrist-band sensors,'' \emph{Computers in Human Behavior Reports}, vol.~4, p.
  100116, 2021.

\bibitem{clements2020sequential}
J.~M. Clements, D.~Xu, N.~Yousefi, and D.~Efimov, ``Sequential deep learning
  for credit risk monitoring with tabular financial data,'' \emph{arXiv
  preprint arXiv:2012.15330}, 2020.

\bibitem{guo2017deepfm}
H.~Guo, R.~Tang, Y.~Ye, Z.~Li, and X.~He, ``Deepfm: a factorization-machine
  based neural network for ctr prediction,'' \emph{arXiv preprint
  arXiv:1703.04247}, 2017.

\bibitem{zhang2019deep}
S.~Zhang, L.~Yao, A.~Sun, and Y.~Tay, ``Deep learning based recommender system:
  A survey and new perspectives,'' \emph{ACM Computing Surveys (CSUR)},
  vol.~52, no.~1, pp. 1--38, 2019.

\bibitem{ahmed2017survey}
M.~Ahmed, H.~Afzal, A.~Majeed, and B.~Khan, ``A survey of evolution in
  predictive models and impacting factors in customer churn,'' \emph{Advances
  in Data Science and Adaptive Analysis}, vol.~9, no.~03, p. 1750007, 2017.

\bibitem{tang2020customer}
Q.~Tang, G.~Xia, X.~Zhang, and F.~Long, ``A customer churn prediction model
  based on xgboost and mlp,'' in \emph{2020 International Conference on
  Computer Engineering and Application (ICCEA)}.\hskip 1em plus 0.5em minus
  0.4em\relax IEEE, 2020, pp. 608--612.

\bibitem{buczak2015survey}
A.~L. Buczak and E.~Guven, ``A survey of data mining and machine learning
  methods for cyber security intrusion detection,'' \emph{IEEE Communications
  surveys \& tutorials}, vol.~18, no.~2, pp. 1153--1176, 2015.

\bibitem{Cartella2021}
F.~Cartella, O.~Anuncia{\c{c}}{\~{a}}o, Y.~Funabiki, D.~Yamaguchi, T.~Akishita,
  and O.~Elshocht, ``{Adversarial attacks for tabular data: application to
  fraud detection and imbalanced data},'' \emph{CEUR Workshop Proceedings},
  vol. 2808, 2021.

\bibitem{liu2021machine}
B.~Liu, M.~Ding, S.~Shaham, W.~Rahayu, F.~Farokhi, and Z.~Lin, ``When machine
  learning meets privacy: A survey and outlook,'' \emph{ACM Computing Surveys
  (CSUR)}, vol.~54, no.~2, pp. 1--36, 2021.

\bibitem{urban2021deep}
C.~J. Urban and K.~M. Gates, ``Deep learning: A primer for psychologists.''
  \emph{Psychological Methods}, 2021.

\bibitem{shoman2020deep}
M.~Shoman, A.~Aboah, and Y.~Adu-Gyamfi, ``Deep learning framework for
  predicting bus delays on multiple routes using heterogenous datasets,''
  \emph{Journal of Big Data Analytics in Transportation}, vol.~2, no.~3, pp.
  275--290, 2020.

\bibitem{pang2021deep}
G.~Pang, C.~Shen, L.~Cao, and A.~V.~D. Hengel, ``Deep learning for anomaly
  detection: A review,'' \emph{ACM Computing Surveys (CSUR)}, vol.~54, no.~2,
  pp. 1--38, 2021.

\bibitem{sahoo2017online}
D.~Sahoo, Q.~Pham, J.~Lu, and S.~C. Hoi, ``Online deep learning: Learning deep
  neural networks on the fly,'' \emph{arXiv preprint arXiv:1711.03705}, 2017.

\bibitem{he2021automl}
X.~He, K.~Zhao, and X.~Chu, ``\text{AutoML}: A survey of the
  state-of-the-art,'' \emph{Knowledge-Based Systems}, vol. 212, p. 106622,
  2021.

\bibitem{artzi2021classification}
M.~Artzi, E.~Redmard, O.~Tzemach, J.~Zeltser, O.~Gropper, J.~Roth, B.~Shofty,
  D.~A. Kozyrev, S.~Constantini, and L.~Ben-Sira, ``Classification of pediatric
  posterior fossa tumors using convolutional neural network and tabular data,''
  \emph{IEEE Access}, vol.~9, pp. 91\,966--91\,973, 2021.

\bibitem{shi2021multimodal}
X.~Shi, J.~Mueller, N.~Erickson, M.~Li, and A.~Smola, ``Multimodal
  \text{AutoML} on structured tables with text fields,'' in \emph{8th ICML
  Workshop on Automated Machine Learning (AutoML)}, 2021.

\bibitem{fakoor2020fastdad}
R.~Fakoor, J.~W. Mueller, N.~Erickson, P.~Chaudhari, and A.~J. Smola, ``Fast,
  accurate, and simple models for tabular data via augmented distillation,''
  \emph{Advances in Neural Information Processing Systems}, vol.~34, 2020.

\bibitem{gijsbers2019open}
P.~Gijsbers, E.~LeDell, J.~Thomas, S.~Poirier, B.~Bischl, and J.~Vanschoren,
  ``An open source \text{AutoML} benchmark,'' \emph{arXiv preprint
  arXiv:1907.00909}, 2019.

\bibitem{yin2020tabert}
P.~Yin, G.~Neubig, W.-t. Yih, and S.~Riedel, ``Ta{BERT}: Pretraining for joint
  understanding of textual and tabular data,'' \emph{arxiv:2005.08314}, 2020.

\bibitem{wang2019generative}
Z.~Wang, Q.~She, and T.~E. Ward, ``Generative adversarial networks in computer
  vision: A survey and taxonomy,'' \emph{arXiv preprint arXiv:1906.01529},
  2019.

\bibitem{baltruvsaitis2018multimodal}
T.~Baltru{\v{s}}aitis, C.~Ahuja, and L.-P. Morency, ``Multimodal machine
  learning: A survey and taxonomy,'' \emph{IEEE transactions on pattern
  analysis and machine intelligence}, vol.~41, no.~2, pp. 423--443, 2018.

\bibitem{lichtenwalter2021deep}
D.~Lichtenwalter, P.~Burggr{\"a}f, J.~Wagner, and T.~Wei{\ss}er, ``Deep
  multimodal learning for manufacturing problem solving,'' \emph{Procedia
  CIRP}, vol.~99, pp. 615--620, 2021.

\bibitem{polsterl2021combining}
S.~P{\"o}lsterl, T.~N. Wolf, and C.~Wachinger, ``Combining 3d image and tabular
  data via the dynamic affine feature map transform,'' \emph{arXiv preprint
  arXiv:2107.05990}, 2021.

\bibitem{soares2021predicting}
D.~d.~B. Soares, F.~Andrieux, B.~Hell, J.~Lenhardt, J.~Badosa, S.~Gavoille,
  S.~Gaiffas, and E.~Bacry, ``Predicting the solar potential of rooftops using
  image segmentation and structured data,'' \emph{arXiv preprint
  arXiv:2106.15268}, 2021.

\bibitem{medvedev2020new}
D.~Medvedev and A.~D'yakonov, ``New properties of the data distillation method
  when working with tabular data,'' \emph{arXiv preprint arXiv:2010.09839},
  2020.

\bibitem{li2020tnt}
J.~Li, Y.~Li, X.~Xiang, S.-T. Xia, S.~Dong, and Y.~Cai, ``Tnt: An interpretable
  tree-network-tree learning framework using knowledge distillation,''
  \emph{Entropy}, vol.~22, no.~11, p. 1203, 2020.

\bibitem{roschewitz2021ifedavg}
D.~Roschewitz, M.-A. Hartley, L.~Corinzia, and M.~Jaggi, ``Ifedavg:
  Interpretable data-interoperability for federated learning,'' \emph{arXiv
  preprint arXiv:2107.06580}, 2021.

\bibitem{sanchez2020improving}
A.~S{\'a}nchez-Morales, J.-L. Sancho-G{\'o}mez, J.-A.
  Mart{\'\i}nez-Garc{\'\i}a, and A.~R. Figueiras-Vidal, ``Improving deep
  learning performance with missing values via deletion and compensation,''
  \emph{Neural Computing and Applications}, vol.~32, no.~17, pp.
  13\,233--13\,244, 2020.

\bibitem{gondara2018mida}
L.~Gondara and K.~Wang, ``Mida: Multiple imputation using denoising
  autoencoders,'' in \emph{Pacific-Asia conference on knowledge discovery and
  data mining}.\hskip 1em plus 0.5em minus 0.4em\relax Springer, 2018, pp.
  260--272.

\bibitem{camino2020working}
R.~D. Camino, C.~Hammerschmidt \emph{et~al.}, ``Working with deep generative
  models and tabular data imputation,'' \emph{ICML 2020 Artemiss Workshop},
  2020.

\bibitem{engelmann2021conditional}
J.~Engelmann and S.~Lessmann, ``Conditional wasserstein gan-based oversampling
  of tabular data for imbalanced learning,'' \emph{Expert Systems with
  Applications}, vol. 174, p. 114582, 2021.

\bibitem{darabi2021synthesising}
S.~Darabi and Y.~Elor, ``Synthesising multi-modal minority samples for tabular
  data,'' \emph{arXiv preprint arXiv:2105.08204}, 2021.

\bibitem{kamthe2021copula}
S.~Kamthe, S.~Assefa, and M.~Deisenroth, ``Copula flows for synthetic data
  generation,'' \emph{arXiv preprint arXiv:2101.00598}, 2021.

\bibitem{Choi2017medGAN}
E.~Choi, S.~Biswal, B.~Malin, J.~Duke, W.~F. Stewart, and J.~Sun, ``{Generating
  Multi-label Discrete Patient Records using Generative Adversarial
  Networks},'' \emph{Machine learning for healthcare conference}, pp. 286--305,
  2017.

\bibitem{ccpa2021}
C.~OAG, ``Ccpa regulations: Final regulation text.'' \emph{Office of the
  Attorney General, California Department of Justice}, 2021.

\bibitem{regulation2016gdpr}
\BIBentryALTinterwordspacing
GDPR, ``Regulation (eu) 2016/679 of the european parliament and of the
  council,'' \emph{Official Journal of the European Union}, 2016. [Online].
  Available: \url{http://www.privacyregulation. eu/en/13.htm}
\BIBentrySTDinterwordspacing

\bibitem{voigt2017eu}
P.~Voigt and A.~Von~dem Bussche, ``The eu general data protection regulation
  (gdpr),'' \emph{A Practical Guide, 1st Ed., Cham: Springer International
  Publishing}, vol.~10, p. 3152676, 2017.

\bibitem{sahakyan2021explainable}
M.~Sahakyan, Z.~Aung, and T.~Rahwan, ``Explainable artificial intelligence for
  tabular data: A survey,'' \emph{IEEE Access}, 2021.

\bibitem{GRISCI2021111}
B.~I. Grisci, M.~J. Krause, and M.~Dorn, ``Relevance aggregation for neural
  networks interpretability and knowledge discovery on tabular data,''
  \emph{Information Sciences}, vol. 559, pp. 111--129, 2021.

\bibitem{bhatt2020explainable}
U.~Bhatt, A.~Xiang, S.~Sharma, A.~Weller, A.~Taly, Y.~Jia, J.~Ghosh, R.~Puri,
  J.~M. Moura, and P.~Eckersley, ``Explainable machine learning in
  deployment,'' in \emph{Proceedings of the 2020 conference on fairness,
  accountability, and transparency}, 2020, pp. 648--657.

\bibitem{XGBOOST}
T.~Chen and C.~Guestrin, ``Xgboost: A scalable tree boosting system,'' in
  \emph{Proceedings of the 22nd ACM SIGKDD International Conference on
  Knowledge Discovery and Data Mining (KDD)}, 2016, pp. 785--794.

\bibitem{hancock2020survey}
J.~T. Hancock and T.~M. Khoshgoftaar, ``Survey on categorical data for neural
  networks,'' \emph{Journal of Big Data}, vol.~7, pp. 1--41, 2020.

\bibitem{gorishniy2021revisiting}
Y.~Gorishniy, I.~Rubachev, V.~Khrulkov, and A.~Babenko, ``Revisiting deep
  learning models for tabular data,'' \emph{arXiv preprint arXiv:2106.11959},
  2021.

\bibitem{he2016deep}
K.~He, X.~Zhang, S.~Ren, and J.~Sun, ``Deep residual learning for image
  recognition,'' in \emph{Proceedings of the IEEE conference on computer vision
  and pattern recognition}, 2016, pp. 770--778.

\bibitem{katzir2021netdnf}
L.~Katzir, G.~Elidan, and R.~El-Yaniv, ``\text{Net-DNF}: Effective deep
  modeling of tabular data,'' in \emph{International Conference on Learning
  Representations}, 2021.

\bibitem{statistics2003Lane}
R.~U. David M.~Lane, \emph{Introduction to Statistics}.\hskip 1em plus 0.5em
  minus 0.4em\relax David Lane, 2003.

\bibitem{ryan2020deep}
M.~Ryan, \emph{Deep learning with structured data}.\hskip 1em plus 0.5em minus
  0.4em\relax Simon and Schuster, 2020.

\bibitem{cvitkovic2020deep}
M.~W. Cvitkovic \emph{et~al.}, ``Deep learning in unconventional domains,''
  Ph.D. dissertation, California Institute of Technology, 2020.

\bibitem{Dua:2019}
\BIBentryALTinterwordspacing
D.~Dua and C.~Graff, ``{UCI} machine learning repository,'' 2017. [Online].
  Available: \url{http://archive.ics.uci.edu/ml}
\BIBentrySTDinterwordspacing

\bibitem{miles1881sunstroke}
A.~J. Miles, ``The sunstroke epidemic of cincinnati, ohio, during the summer of
  1881,'' \emph{Public health papers and reports}, vol.~7, p. 293, 1881.

\bibitem{fisher1936use}
R.~A. Fisher, ``The use of multiple measurements in taxonomic problems,''
  \emph{Annals of eugenics}, vol.~7, no.~2, pp. 179--188, 1936.

\bibitem{jdanov2019human}
D.~A. Jdanov, D.~Jasilionis, V.~M. Shkolnikov, and M.~Barbieri, ``Human
  mortality database,'' \emph{Encyclopedia of gerontology and population
  aging/editors Danan Gu, Matthew E. Dupre. Cham: Springer International
  Publishing, 2020}, 2019.

\bibitem{fix1951discriminatory}
E.~Fix, \emph{Discriminatory analysis: nonparametric discrimination,
  consistency properties}.\hskip 1em plus 0.5em minus 0.4em\relax USAF school
  of Aviation Medicine, 1951.

\bibitem{giles1992learning}
C.~L. Giles, C.~B. Miller, D.~Chen, H.-H. Chen, G.-Z. Sun, and Y.-C. Lee,
  ``Learning and extracting finite state automata with second-order recurrent
  neural networks,'' \emph{Neural Computation}, vol.~4, no.~3, pp. 393--405,
  1992.

\bibitem{horne1995experimental}
B.~G. Horne and C.~L. Giles, ``An experimental comparison of recurrent neural
  networks,'' \emph{Advances in neural information processing systems}, pp.
  697--704, 1995.

\bibitem{willenborg1996statistical}
L.~Willenborg and T.~De~Waal, \emph{Statistical disclosure control in
  practice}.\hskip 1em plus 0.5em minus 0.4em\relax Springer Science \&
  Business Media, 1996, vol. 111.

\bibitem{richardson2007predicting}
M.~Richardson, E.~Dominowska, and R.~Ragno, ``Predicting clicks: estimating the
  click-through rate for new ads,'' in \emph{Proceedings of the 16th
  international conference on World Wide Web}, 2007, pp. 521--530.

\bibitem{ke2019deepgbm}
G.~Ke, Z.~Xu, J.~Zhang, J.~Bian, and T.-Y. Liu, ``Deepgbm: A deep learning
  framework distilled by gbdt for online prediction tasks,'' in
  \emph{Proceedings of the 25th ACM SIGKDD International Conference on
  Knowledge Discovery \& Data Mining}, 2019, pp. 384--394.

\bibitem{wang2021masknet}
Z.~Wang, Q.~She, and J.~Zhang, ``Masknet: Introducing feature-wise
  multiplication to ctr ranking models by instance-guided mask,''
  \emph{arXiv:2102.07619}, 2021.

\bibitem{shavitt2018regularization}
I.~Shavitt and E.~Segal, ``Regularization learning networks: deep learning for
  tabular datasets,'' in \emph{Advances in Neural Information Processing
  Systems}, 2018, pp. 1379--1389.

\bibitem{brown2020language}
T.~B. Brown, B.~Mann, N.~Ryder, M.~Subbiah, J.~Kaplan, P.~Dhariwal,
  A.~Neelakantan, P.~Shyam, G.~Sastry, A.~Askell \emph{et~al.}, ``Language
  models are few-shot learners,'' \emph{arXiv: 2005.14165}, 2020.

\bibitem{dosovitskiy2021visiontransformer}
A.~Dosovitskiy, L.~Beyer, A.~Kolesnikov, D.~Weissenborn, X.~Zhai,
  T.~Unterthiner, M.~Dehghani, M.~Minderer, G.~Heigold, S.~Gelly, J.~Uszkoreit,
  and N.~Houlsby, ``An image is worth 16x16 words: Transformers for image
  recognition at scale,'' in \emph{International Conference on Learning
  Representations}, 2021.

\bibitem{khan2021transformers}
S.~Khan, M.~Naseer, M.~Hayat, S.~W. Zamir, F.~S. Khan, and M.~Shah,
  ``Transformers in vision: A survey,'' \emph{arXiv preprint arXiv:2101.01169},
  2021.

\bibitem{karr2006data}
A.~F. Karr, A.~P. Sanil, and D.~L. Banks, ``Data quality: A statistical
  perspective,'' \emph{Statistical Methodology}, vol.~3, no.~2, pp. 137--173,
  2006.

\bibitem{Xu2018TGAN}
L.~Xu and K.~Veeramachaneni, ``{Synthesizing Tabular Data using Generative
  Adversarial Networks},'' \emph{arXiv preprint arXiv:1811.11264}, 2018.

\bibitem{ke2017lightgbm}
G.~Ke, Q.~Meng, T.~Finley, T.~Wang, W.~Chen, W.~Ma, Q.~Ye, and T.-Y. Liu,
  ``Lightgbm: A highly efficient gradient boosting decision tree,'' in
  \emph{Advances in neural information processing systems}, 2017, pp.
  3146--3154.

\bibitem{prokhorenkova2018catboost}
L.~Prokhorenkova, G.~Gusev, A.~Vorobev, A.~V. Dorogush, and A.~Gulin,
  ``Catboost: unbiased boosting with categorical features,'' in \emph{Advances
  in neural information processing systems}, 2018, pp. 6638--6648.

\bibitem{Zhu2021igtd}
Y.~Zhu, T.~Brettin, F.~Xia, A.~Partin, M.~Shukla, H.~Yoo, Y.~A. Evrard, J.~H.
  Doroshow, and R.~L. Stevens, ``Converting tabular data into images for deep
  learning with convolutional neural networks,'' \emph{Scientific Reports},
  vol.~11, no.~1, pp. 1--11, 2021.

\bibitem{rahaman2019spectral}
N.~Rahaman, A.~Baratin, D.~Arpit, F.~Draxler, M.~Lin, F.~Hamprecht, Y.~Bengio,
  and A.~Courville, ``On the spectral bias of neural networks,'' in
  \emph{International Conference on Machine Learning}.\hskip 1em plus 0.5em
  minus 0.4em\relax PMLR, 2019, pp. 5301--5310.

\bibitem{mitchell2017spatial}
B.~R. Mitchell \emph{et~al.}, ``The spatial inductive bias of deep learning,''
  Ph.D. dissertation, Johns Hopkins University, 2017.

\bibitem{GoodfellowDLBook2016}
I.~Goodfellow, Y.~Bengio, and A.~Courville, \emph{Deep Learning}.\hskip 1em
  plus 0.5em minus 0.4em\relax MIT Press, 2016.

\bibitem{gorishniy2022embeddings}
Y.~Gorishniy, I.~Rubachev, and A.~Babenko, ``On embeddings for numerical
  features in tabular deep learning,'' \emph{arXiv preprint arXiv:2203.05556},
  2022.

\bibitem{fitkov2012evaluating}
E.~Fitkov-Norris, S.~Vahid, and C.~Hand, ``Evaluating the impact of categorical
  data encoding and scaling on neural network classification performance: the
  case of repeat consumption of identical cultural goods,'' in
  \emph{International Conference on Engineering Applications of Neural
  Networks}.\hskip 1em plus 0.5em minus 0.4em\relax Springer, 2012, pp.
  343--352.

\bibitem{baylor2017tfx}
D.~Baylor, E.~Breck, H.-T. Cheng, N.~Fiedel, C.~Y. Foo, Z.~Haque, S.~Haykal,
  M.~Ispir, V.~Jain, L.~Koc \emph{et~al.}, ``\text{TFX}: A tensorflow-based
  production-scale machine learning platform,'' in \emph{Proceedings of the
  23rd ACM SIGKDD International Conference on Knowledge Discovery and Data
  Mining}, 2017, pp. 1387--1395.

\bibitem{sun2019supertml}
B.~Sun, L.~Yang, W.~Zhang, M.~Lin, P.~Dong, C.~Young, and J.~Dong, ``Supertml:
  Two-dimensional word embedding for the precognition on structured tabular
  data,'' in \emph{Proceedings of the IEEE/CVF Conference on Computer Vision
  and Pattern Recognition Workshops}, 2019, pp. 0--0.

\bibitem{yoon2020vime}
J.~Yoon, Y.~Zhang, J.~Jordon, and M.~van~der Schaar, ``Vime: Extending the
  success of self-and semi-supervised learning to tabular domaindim,''
  \emph{Advances in Neural Information Processing Systems}, vol.~33, 2020.

\bibitem{bahri2021scarf}
D.~Bahri, H.~Jiang, Y.~Tay, and D.~Metzler, ``\text{SCARF}: Self-supervised
  contrastive learning using random feature corruption,'' \emph{arXiv preprint
  arXiv:2106.15147}, 2021.

\bibitem{cheng2016wide}
H.-T. Cheng, L.~Koc, J.~Harmsen, T.~Shaked, T.~Chandra, H.~Aradhye,
  G.~Anderson, G.~Corrado, W.~Chai, M.~Ispir \emph{et~al.}, ``Wide \& deep
  learning for recommender systems,'' in \emph{Proceedings of the 1st workshop
  on deep learning for recommender systems}, 2016, pp. 7--10.

\bibitem{frosst2017distilling}
N.~Frosst and G.~Hinton, ``Distilling a neural network into a soft decision
  tree,'' \emph{arXiv preprint arXiv:1711.09784}, 2017.

\bibitem{lian2018xdeepfm}
J.~Lian, X.~Zhou, F.~Zhang, Z.~Chen, X.~Xie, and G.~Sun, ``xdeepfm: Combining
  explicit and implicit feature interactions for recommender systems,'' in
  \emph{Proceedings of the 24th ACM SIGKDD International Conference on
  Knowledge Discovery \& Data Mining}, 2018, pp. 1754--1763.

\bibitem{Ke2019tabnn}
G.~Ke, J.~Zhang, Z.~Xu, J.~Bian, and T.-Y. Liu, ``\text{TabNN}: A universal
  neural network solution for tabular data,'' 2018.

\bibitem{luo2020NON}
Y.~Luo, H.~Zhou, W.-W. Tu, Y.~Chen, W.~Dai, and Q.~Yang, ``Network on network
  for tabular data classification in real-world applications,'' in
  \emph{Proceedings of the 43rd International ACM SIGIR Conference on Research
  and Development in Information Retrieval}, 2020, pp. 2317--2326.

\bibitem{liu2020dnn2lr}
Z.~Liu, Q.~Liu, H.~Zhang, and Y.~Chen, ``Dnn2lr: Interpretation-inspired
  feature crossing for real-world tabular data,'' \emph{arXiv preprint
  arXiv:2008.09775}, 2020.

\bibitem{Ivanov2021bgnn}
S.~Ivanov and L.~Prokhorenkova, ``{Boost then Convolve: Gradient Boosting Meets
  Graph Neural Networks},'' in \emph{Interational Conference on Learning
  Representations}, 2021.

\bibitem{Luo2021sdtr}
H.~Luo, F.~Cheng, H.~Yu, and Y.~Yi, ``{SDTR: Soft Decision Tree Regressor for
  Tabular Data},'' \emph{IEEE Access}, vol.~9, pp. 55\,999--56\,011, 2021.

\bibitem{Huang2020tabtrans}
X.~Huang, A.~Khetan, M.~Cvitkovic, and Z.~Karnin, ``{TabTransformer: Tabular
  Data Modeling Using Contextual Embeddings},'' \emph{arxiv:2012.06678}, 2020.

\bibitem{cai2021arm}
S.~Cai, K.~Zheng, G.~Chen, H.~Jagadish, B.~C. Ooi, and M.~Zhang, ``Arm-net:
  Adaptive relation modeling network for structured data,'' in
  \emph{Proceedings of the 2021 International Conference on Management of
  Data}, 2021, pp. 207--220.

\bibitem{kossen2021selfattention}
J.~Kossen, N.~Band, C.~Lyle, A.~Gomez, T.~Rainforth, and Y.~Gal,
  ``Self-attention between datapoints: Going beyond individual input-output
  pairs in deep learning,'' in \emph{Advances in Neural Information Processing
  Systems}, A.~Beygelzimer, Y.~Dauphin, P.~Liang, and J.~W. Vaughan, Eds.,
  2021.

\bibitem{MicciBarreca2001}
D.~Micci-Barreca, ``A preprocessing scheme for high-cardinality categorical
  attributes in classification and prediction problems,'' \emph{SIGKDD
  Explor.}, vol.~3, pp. 27--32, 2001.

\bibitem{friedman2002stochastic}
J.~H. Friedman, ``Stochastic gradient boosting,'' \emph{Computational
  statistics \& data analysis}, vol.~38, no.~4, pp. 367--378, 2002.

\bibitem{langley1994oblivious}
P.~Langley and S.~Sage, ``Oblivious decision trees and abstract cases,'' in
  \emph{Working notes of the AAAI-94 workshop on case-based reasoning}.\hskip
  1em plus 0.5em minus 0.4em\relax Seattle, WA, 1994, pp. 113--117.

\bibitem{peters2019sparse}
B.~Peters, V.~Niculae, and A.~F. Martins, ``Sparse sequence-to-sequence
  models,'' \emph{arxiv:1905.05702}, 2019.

\bibitem{lecun-mnisthandwrittendigit-2010}
\BIBentryALTinterwordspacing
Y.~LeCun and C.~Cortes, ``{MNIST} handwritten digit database,'' 2010. [Online].
  Available: \url{http://yann.lecun.com/exdb/mnist/}
\BIBentrySTDinterwordspacing

\bibitem{guo2016entity}
C.~Guo and F.~Berkhahn, ``Entity embeddings of categorical variables,''
  \emph{arXiv preprint arXiv:1604.06737}, 2016.

\bibitem{rendle2010factorization}
S.~Rendle, ``Factorization machines,'' in \emph{2010 IEEE International
  conference on data mining}.\hskip 1em plus 0.5em minus 0.4em\relax IEEE,
  2010, pp. 995--1000.

\bibitem{moosmann2006fast}
F.~Moosmann, B.~Triggs, and F.~Jurie, ``Fast discriminative visual codebooks
  using randomized clustering forests,'' in \emph{Twentieth Annual Conference
  on Neural Information Processing Systems (NIPS'06)}.\hskip 1em plus 0.5em
  minus 0.4em\relax MIT Press, 2006, pp. 985--992.

\bibitem{FACEBOOK}
X.~He, J.~Pan, O.~Jin, T.~Xu, B.~Liu, T.~Xu, Y.~Shi, A.~Atallah, R.~Herbrich,
  S.~Bowers \emph{et~al.}, ``Practical lessons from predicting clicks on ads at
  facebook,'' in \emph{Proceedings of the Eighth International Workshop on Data
  Mining for Online Advertising}, 2014, pp. 1--9.

\bibitem{scarselli2008graph}
F.~Scarselli, M.~Gori, A.~C. Tsoi, M.~Hagenbuchner, and G.~Monfardini, ``The
  graph neural network model,'' \emph{IEEE transactions on neural networks},
  vol.~20, no.~1, pp. 61--80, 2008.

\bibitem{wang2019language}
C.~Wang, M.~Li, and A.~J. Smola, ``Language models with transformers,''
  \emph{arXiv preprint arXiv:1904.09408}, 2019.

\bibitem{Martins2016sparsemax}
A.~F.~T. Martins and R.~F. Astudillo, ``From softmax to sparsemax: {A} sparse
  model of attention and multi-label classification,'' \emph{arxiv:1602.02068},
  2016.

\bibitem{van1995python}
G.~Van~Rossum and F.~L. Drake~Jr, \emph{Python reference manual}.\hskip 1em
  plus 0.5em minus 0.4em\relax Centrum voor Wiskunde en Informatica Amsterdam,
  1995.

\bibitem{joseph2021pytorch}
M.~Joseph, ``Pytorch tabular: A framework for deep learning with tabular
  data,'' \emph{arXiv preprint arXiv:2104.13638}, 2021.

\bibitem{boughorbel2021fairness}
S.~Boughorbel, F.~Jarray, and A.~Kadri, ``Fairness in tabnet model by
  disentangled representation for the prediction of hospital no-show,''
  \emph{arXiv preprint arXiv:2103.04048}, 2021.

\bibitem{mehrabi2021survey}
N.~Mehrabi, F.~Morstatter, N.~Saxena, K.~Lerman, and A.~Galstyan, ``A survey on
  bias and fairness in machine learning,'' \emph{ACM Computing Surveys (CSUR)},
  vol.~54, no.~6, pp. 1--35, 2021.

\bibitem{yun2019cutmix}
S.~Yun, D.~Han, S.~J. Oh, S.~Chun, J.~Choe, and Y.~Yoo, ``Cutmix:
  Regularization strategy to train strong classifiers with localizable
  features,'' in \emph{Proceedings of the IEEE/CVF International Conference on
  Computer Vision}, 2019, pp. 6023--6032.

\bibitem{zhang2017mixup}
H.~Zhang, M.~Cisse, Y.~N. Dauphin, and D.~Lopez-Paz, ``mixup: Beyond empirical
  risk minimization,'' \emph{arXiv preprint arXiv:1710.09412}, 2017.

\bibitem{borisov2019cancelout}
V.~Borisov, J.~Haug, and G.~Kasneci, ``\text{CancelOut}: A layer for feature
  selection in deep neural networks,'' in \emph{International Conference on
  Artificial Neural Networks}.\hskip 1em plus 0.5em minus 0.4em\relax Springer,
  2019, pp. 72--83.

\bibitem{valdes2021lockout}
G.~Valdes, W.~Arbelo, Y.~Interian, and J.~H. Friedman, ``Lockout: Sparse
  regularization of neural networks,'' \emph{arXiv preprint arXiv:2107.07160},
  2021.

\bibitem{fiedler2021simple}
J.~Fiedler, ``Simple modifications to improve tabular neural networks,''
  \emph{arXiv preprint arXiv:2108.03214}, 2021.

\bibitem{lounici2021muddling}
K.~Lounici, K.~Meziani, and B.~Riu, ``Muddling label regularization: Deep
  learning for tabular datasets,'' \emph{arXiv preprint arXiv:2106.04462},
  2021.

\bibitem{lundberg2017unified}
S.~Lundberg and S.-I. Lee, ``A unified approach to interpreting model
  predictions,'' \emph{NeurIPS}, 2017.

\bibitem{chen2019faketables}
H.~Chen, S.~Jajodia, J.~Liu, N.~Park, V.~Sokolov, and V.~Subrahmanian,
  ``Faketables: Using gans to generate functional dependency preserving tables
  with bounded real data.'' in \emph{IJCAI}, 2019, pp. 2074--2080.

\bibitem{Quintana2019towardsHumanComfort}
M.~Quintana and C.~Miller, ``Towards class-balancing human comfort datasets
  with gans,'' in \emph{Proceedings of the 6th ACM International Conference on
  Systems for Energy-Efficient Buildings, Cities, and Transportation}, 2019,
  pp. 391--392.

\bibitem{koivu2020synthetic}
A.~Koivu, M.~Sairanen, A.~Airola, and T.~Pahikkala, ``Synthetic minority
  oversampling of vital statistics data with generative adversarial networks,''
  \emph{Journal of the American Medical Informatics Association}, vol.~27,
  no.~11, pp. 1667--1674, 2020.

\bibitem{Fan2020designSpace}
J.~Fan, J.~Chen, T.~Liu, Y.~Shen, G.~Li, and X.~Du, ``Relational data synthesis
  using generative adversarial networks: A design space exploration,''
  \emph{Proc. VLDB Endow.}, vol.~13, no.~12, p. 1962–1975, Jul. 2020.

\bibitem{karras2020analyzing}
T.~Karras, S.~Laine, M.~Aittala, J.~Hellsten, J.~Lehtinen, and T.~Aila,
  ``Analyzing and improving the image quality of stylegan,'' in
  \emph{Proceedings of the IEEE/CVF Conference on Computer Vision and Pattern
  Recognition}, 2020, pp. 8110--8119.

\bibitem{lin2017adversarial}
K.~Lin, D.~Li, X.~He, Z.~Zhang, and M.-T. Sun, ``Adversarial ranking for
  language generation,'' \emph{Advances in Neural Information Processing
  Systems}, 2017.

\bibitem{subramanian2017adversarial}
S.~Subramanian, S.~Rajeswar, F.~Dutil, C.~Pal, and A.~Courville, ``Adversarial
  generation of natural language,'' in \emph{Proceedings of the 2nd Workshop on
  Representation Learning for NLP}, 2017, pp. 241--251.

\bibitem{patki2016synthetic}
N.~Patki, R.~Wedge, and K.~Veeramachaneni, ``The synthetic data vault,'' in
  \emph{2016 IEEE International Conference on Data Science and Advanced
  Analytics (DSAA)}.\hskip 1em plus 0.5em minus 0.4em\relax IEEE, 2016, pp.
  399--410.

\bibitem{li2020sync}
Z.~Li, Y.~Zhao, and J.~Fu, ``Sync: A copula based framework for generating
  synthetic data from aggregated sources,'' in \emph{2020 International
  Conference on Data Mining Workshops (ICDMW)}.\hskip 1em plus 0.5em minus
  0.4em\relax IEEE, 2020, pp. 571--578.

\bibitem{zhang2017privbayes}
J.~Zhang, G.~Cormode, C.~M. Procopiuc, D.~Srivastava, and X.~Xiao, ``Privbayes:
  Private data release via bayesian networks,'' \emph{ACM Transactions on
  Database Systems (TODS)}, vol.~42, no.~4, pp. 1--41, 2017.

\bibitem{chow1968approximating}
C.~Chow and C.~Liu, ``Approximating discrete probability distributions with
  dependence trees,'' \emph{IEEE transactions on Information Theory}, vol.~14,
  no.~3, pp. 462--467, 1968.

\bibitem{goodfellow2014generative}
I.~J. Goodfellow, J.~Pouget-Abadie, M.~Mirza, B.~Xu, D.~Warde-Farley, S.~Ozair,
  A.~Courville, and Y.~Bengio, ``Generative adversarial networks,'' \emph{arXiv
  preprint arXiv:1406.2661}, 2014.

\bibitem{radford2015unsupervised}
A.~Radford, L.~Metz, and S.~Chintala, ``Unsupervised representation learning
  with deep convolutional generative adversarial networks,'' in \emph{4th
  International Conference on Learning Representations, ICLR 2016}, 2016.

\bibitem{arjovsky2017wasserstein}
M.~Arjovsky, S.~Chintala, and L.~Bottou, ``Wasserstein generative adversarial
  networks,'' in \emph{International conference on machine learning}.\hskip 1em
  plus 0.5em minus 0.4em\relax PMLR, 2017, pp. 214--223.

\bibitem{gulrajani2017improved}
I.~Gulrajani, F.~Ahmed, M.~Arjovsky, V.~Dumoulin, and A.~Courville, ``Improved
  training of wasserstein gans,'' in \emph{Proceedings of the 31st
  International Conference on Neural Information Processing Systems}, 2017, pp.
  5769--5779.

\bibitem{bellemare2017cramer}
M.~G. Bellemare, I.~Danihelka, W.~Dabney, S.~Mohamed, B.~Lakshminarayanan,
  S.~Hoyer, and R.~Munos, ``The cramer distance as a solution to biased
  wasserstein gradients,'' 2017.

\bibitem{hjelm2017boundary}
R.~D. Hjelm, A.~P. Jacob, T.~Che, A.~Trischler, K.~Cho, and Y.~Bengio,
  ``Boundary-seeking generative adversarial networks,'' \emph{International
  Conference on Learning Representations}, 2018.

\bibitem{srivastava2017veegan}
A.~Srivastava, L.~Valkov, C.~Russell, M.~U. Gutmann, and C.~Sutton, ``Veegan:
  Reducing mode collapse in gans using implicit variational learning,'' in
  \emph{Proceedings of the 31st International Conference on Neural Information
  Processing Systems}, 2017, pp. 3310--3320.

\bibitem{Kingma2014VAE}
D.~P. Kingma and M.~Welling, ``{Auto-encoding variational bayes},'' \emph{2nd
  International Conference on Learning Representations, ICLR 2014 - Conference
  Track Proceedings}, no.~Ml, pp. 1--14, 2014.

\bibitem{Ma2020VAEM}
C.~Ma, S.~Tschiatschek, R.~Turner, J.~M. Hern{\'a}ndez-Lobato, and C.~Zhang,
  ``Vaem: a deep generative model for heterogeneous mixed type data,''
  \emph{Advances in Neural Information Processing Systems}, vol.~33, 2020.

\bibitem{Xu2019CTGAN}
L.~Xu, M.~Skoularidou, A.~Cuesta-Infante, and K.~Veeramachaneni, ``{Modeling
  tabular data using conditional GAN},'' in \emph{Advances in Neural
  Information Processing Systems}, vol.~33, 2019.

\bibitem{Park2018GAN}
N.~Park, M.~Mohammadi, K.~Gorde, S.~Jajodia, H.~Park, and Y.~Kim, ``{Data
  synthesis based on generative adversarial networks},'' \emph{Proceedings of
  the VLDB Endowment}, vol.~11, no.~10, pp. 1071--1083, 2018.

\bibitem{jang2016categorical}
E.~Jang, S.~Gu, and B.~Poole, ``Categorical reparameterization with
  gumbel-softmax,'' in \emph{International Conference on Learning
  Representations}, 2017.

\bibitem{wen2021causal}
B.~Wen, L.~O. Colon, K.~Subbalakshmi, and R.~Chandramouli, ``Causal-{TGAN}:
  Generating tabular data using causal generative adversarial networks,''
  \emph{arXiv preprint arXiv:2104.10680}, 2021.

\bibitem{jordon2018pate}
J.~Jordon, J.~Yoon, and M.~Van Der~Schaar, ``Pate-gan: Generating synthetic
  data with differential privacy guarantees,'' in \emph{International
  conference on learning representations}, 2018.

\bibitem{Mottini2018GAN}
A.~Mottini, A.~Lheritier, and R.~Acuna-Agost, ``{Airline Passenger Name Record
  Generation using Generative Adversarial Networks},'' \emph{arXiv preprint
  arXiv:1807.06657}, 2018.

\bibitem{camino2018generating}
R.~Camino, C.~Hammerschmidt, and R.~State, ``Generating multi-categorical
  samples with generative adversarial networks,'' \emph{ICML workshop on
  Theoretical Foundations and Applications of Deep Generative Models}, 2018.

\bibitem{Baowaly2019GAN}
M.~K. Baowaly, C.~C. Lin, C.~L. Liu, and K.~T. Chen, ``{Synthesizing electronic
  health records using improved generative adversarial networks},''
  \emph{Journal of the American Medical Informatics Association}, vol.~26,
  no.~3, pp. 228--241, 2019.

\bibitem{vardhan2020generating}
L.~V.~H. Vardhan and S.~Kok, ``Generating privacy-preserving synthetic tabular
  data using oblivious variational autoencoders,'' in \emph{Proceedings of the
  Workshop on Economics of Privacy and Data Labor at the 37 th International
  Conference on Machine Learning}, 2020.

\bibitem{zhao2021ctab}
Z.~Zhao, A.~Kunar, H.~Van~der Scheer, R.~Birke, and L.~Y. Chen, ``Ctab-gan:
  Effective table data synthesizing,'' \emph{arXiv preprint arXiv:2102.08369},
  2021.

\bibitem{massey1951kolmogorov}
F.~J. Massey~Jr, ``The kolmogorov-smirnov test for goodness of fit,''
  \emph{Journal of the American statistical Association}, vol.~46, no. 253, pp.
  68--78, 1951.

\bibitem{guidotti2018survey}
R.~Guidotti, A.~Monreale, S.~Ruggieri, F.~Turini, F.~Giannotti, and
  D.~Pedreschi, ``A survey of methods for explaining black box models,''
  \emph{ACM Comput. Surv.}, vol.~51, no.~5, 2018.

\bibitem{gade2019explainable}
K.~Gade, S.~C. Geyik, K.~Kenthapadi, V.~Mithal, and A.~Taly, ``Explainable ai
  in industry,'' in \emph{Proceedings of the ACM SIGKDD International
  Conference on Knowledge Discovery and Data Mining (KDD)}, 2019.

\bibitem{pawelczyk2021carla}
M.~Pawelczyk, S.~Bielawski, J.~Van~den Heuvel, T.~Richter, and G.~Kasneci,
  ``Carla: A python library to benchmark algorithmic recourse and
  counterfactual explanation algorithms,'' in \emph{Advances in Neural
  Information Processing Systems (NeurIPS) Benchmark and Datasets Track}, 2021.

\bibitem{lou2012intelligible}
Y.~Lou, R.~Caruana, and J.~Gehrke, ``Intelligible models for classification and
  regression,'' in \emph{Proceedings of the ACM SIGKDD International Conference
  on Knowledge Discovery and Data Mining (KDD)}, 2012.

\bibitem{alvarez2018towards}
D.~Alvarez-Melis and T.~S. Jaakkola, ``Towards robust interpretability with
  self-explaining neural networks,'' \emph{NeurIPS}, 2018.

\bibitem{wang2019designing}
D.~Wang, Q.~Yang, A.~Abdul, and B.~Y. Lim, ``Designing theory-driven
  user-centric explainable ai,'' in \emph{CHI}, 2019.

\bibitem{bach2015pixel}
S.~Bach, A.~Binder, G.~Montavon, F.~Klauschen, K.-R. M{\"u}ller, and W.~Samek,
  ``On pixel-wise explanations for non-linear classifier decisions by
  layer-wise relevance propagation,'' \emph{PloS one}, vol.~10, no.~7, p.
  e0130140, 2015.

\bibitem{montavon2019layer}
G.~Montavon, A.~Binder, S.~Lapuschkin, W.~Samek, and K.-R. M{\"u}ller,
  ``Layer-wise relevance propagation: an overview,'' \emph{Explainable AI:
  interpreting, explaining and visualizing deep learning}, pp. 193--209, 2019.

\bibitem{kasneci2016licon}
G.~Kasneci and T.~Gottron, ``Licon: A linear weighting scheme for the
  contribution ofinput variables in deep artificial neural networks,'' in
  \emph{CIKM}, 2016.

\bibitem{sundararajan2017axiomatic}
M.~Sundararajan, A.~Taly, and Q.~Yan, ``Axiomatic attribution for deep
  networks,'' in \emph{International Conference on Machine Learning}.\hskip 1em
  plus 0.5em minus 0.4em\relax PMLR, 2017, pp. 3319--3328.

\bibitem{chattopadhay2018grad}
A.~Chattopadhay, A.~Sarkar, P.~Howlader, and V.~N. Balasubramanian,
  ``Grad-cam++: Generalized gradient-based visual explanations for deep
  convolutional networks,'' in \emph{2018 IEEE Winter Conference on
  Applications of Computer Vision (WACV)}.\hskip 1em plus 0.5em minus
  0.4em\relax IEEE, 2018.

\bibitem{ribeiro2016should}
M.~T. Ribeiro, S.~Singh, and C.~Guestrin, ``" why should i trust you?"
  explaining the predictions of any classifier,'' in \emph{Proceedings of the
  22nd ACM SIGKDD international conference on knowledge discovery and data
  mining}, 2016, pp. 1135--1144.

\bibitem{ribeiro2018anchors}
M.~T. Ribeiro, S.~Singh, and C.Guestrin, ``Anchors: High-precision
  model-agnostic explanations,'' in \emph{AAAI}, 2018.

\bibitem{lundberg2020local}
S.~M. Lundberg, G.~Erion, H.~Chen, A.~DeGrave, J.~M. Prutkin, B.~Nair, R.~Katz,
  J.~Himmelfarb, N.~Bansal, and S.-I. Lee, ``From local explanations to global
  understanding with explainable ai for trees,'' \emph{Nature machine
  intelligence}, 2020.

\bibitem{haug2021baselines}
J.~Haug, S.~Zürn, P.~El-Jiz, and G.~Kasneci, ``On baselines for local feature
  attributions,'' \emph{arXiv: 2101.00905}, 2021.

\bibitem{liu2021synthetic}
Y.~Liu, S.~Khandagale, C.~White, and W.~Neiswanger, ``Synthetic benchmarks for
  scientific research in explainable machine learning,'' in \emph{Advances in
  Neural Information Processing Systems (NeurIPS) Benchmark and Datasets
  Track}, 2021.

\bibitem{wachter2017counterfactual}
S.~Wachter, B.~Mittelstadt, and C.~Russell, ``Counterfactual explanations
  without opening the black box: automated decisions and the gdpr,''
  \emph{Harvard Journal of Law \& Technology}, vol.~31, no.~2, 2018.

\bibitem{ustun2019actionable}
B.~Ustun, A.~Spangher, and Y.~Liu, ``Actionable recourse in linear
  classification,'' in \emph{(FAT*)}, 2019.

\bibitem{russell2019efficient}
C.~Russell, ``Efficient search for diverse coherent explanations,'' in
  \emph{(FAT*)}, 2019.

\bibitem{rawal2020individualized}
K.~Rawal and H.~Lakkaraju, ``Beyond individualized recourse: Interpretable and
  interactive summaries of actionable recourses,'' in \emph{NeurIPS}, 2020.

\bibitem{karimi2019model}
A.-H. Karimi, G.~Barthe, B.~Balle, and I.~Valera, ``Model-agnostic
  counterfactual explanations for consequential decisions,'' in \emph{AISTATS},
  2020.

\bibitem{Dhurandhar2018}
A.~Dhurandhar, P.-Y. Chen, R.~Luss, C.-C. Tu, P.~Ting, K.~Shanmugam, and
  P.~Das, ``Explanations based on the missing: Towards contrastive explanations
  with pertinent negatives,'' in \emph{Advances in Neural Information
  Processing Systems (NeurIPS)}, 2018.

\bibitem{mittelstadt2019explaining}
B.~Mittelstadt, C.~Russell, and S.~Wachter, ``Explaining explanations in ai,''
  in \emph{Proceedings of the conference on fairness, accountability, and
  transparency}, 2019.

\bibitem{mothilal2020fat}
R.~K. Mothilal, A.~Sharma, and C.~Tan, ``Explaining machine learning
  classifiers through diverse counterfactual explanations,'' in \emph{FAT*},
  2020.

\bibitem{pawelczyk2019}
M.~Pawelczyk, K.~Broelemann, and G.~Kasneci, ``Learning model-agnostic
  counterfactual explanations for tabular data,'' in \emph{The Web Conference
  2020 (WWW)}.\hskip 1em plus 0.5em minus 0.4em\relax ACM, 2020.

\bibitem{downs2020interpretable}
M.~Downs, J.~L. Chu, Y.~Yacoby, F.~Doshi-Velez, and W.~Pan, ``Cruds:
  Counterfactual recourse using disentangled subspaces,'' \emph{ICML Workshop
  on Human Interpretability in Machine Learning (WHI 2020)}, 2020.

\bibitem{joshi2019towards}
S.~Joshi, O.~Koyejo, W.~Vijitbenjaronk, B.~Kim, and J.~Ghosh, ``Towards
  realistic individual recourse and actionable explanations in black-box
  decision making systems,'' \emph{arXiv preprint arXiv:1907.09615}, 2019.

\bibitem{mahajan2019preserving}
D.~Mahajan, C.~Tan, and A.~Sharma, ``Preserving causal constraints in
  counterfactual explanations for machine learning classifiers,'' \emph{arXiv
  preprint arXiv:1912.03277}, 2019.

\bibitem{pmlr-v124-pawelczyk20a}
M.~Pawelczyk, K.~Broelemann, and G.~Kasneci, ``On counterfactual explanations
  under predictive multiplicity,'' in \emph{Conference on Uncertainty in
  Artificial Intelligence (UAI)}.\hskip 1em plus 0.5em minus 0.4em\relax PMLR,
  2020.

\bibitem{antoran2020getting}
J.~Antorán, U.~Bhatt, T.~Adel, A.~Weller, and J.~M. Hernández-Lobato,
  ``Getting a clue: A method for explaining uncertainty estimates,''
  \emph{ICLR}, 2021.

\bibitem{nazabal2018handling}
A.~Nazabal, P.~M. Olmos, Z.~Ghahramani, and I.~Valera, ``Handling incomplete
  heterogeneous data using vaes,'' \emph{Pattern Recognition}, 2020.

\bibitem{upadhyay2021robust}
S.~Upadhyay, S.~Joshi, and H.~Lakkaraju, ``Towards robust and reliable
  algorithmic recourse,'' in \emph{Advances in Neural Information Processing
  Systems (NeurIPS)}, vol.~34, 2021.

\bibitem{dominguezolmedo2021adversarial}
R.~Dominguez-Olmedo, A.-H. Karimi, and B.~Schölkopf, ``On the adversarial
  robustness of causal algorithmic recourse,'' in \emph{International
  Conference on Machine Learning (ICML)}, 2022.

\bibitem{pawelczyk2022noisy}
M.~Pawelczyk, T.~Datta, J.~van-den Heuvel, G.~Kasneci, and H.~Lakkaraju,
  ``Algorithmic recourse in the face of noisy human responses,''
  \emph{arXiv:2203.06768}, 2022.

\bibitem{karimi2020survey}
A.-H. Karimi, G.~Barthe, B.~Sch{\"o}lkopf, and I.~Valera, ``A survey of
  algorithmic recourse: definitions, formulations, solutions, and prospects,''
  \emph{arXiv preprint arXiv:2010.04050}, 2020.

\bibitem{verma2020counterfactual}
S.~Verma, J.~Dickerson, and K.~Hines, ``Counterfactual explanations for machine
  learning: A review,'' 2020.

\bibitem{Krizhevsky09learningmultiple}
A.~Krizhevsky, ``Learning multiple layers of features from tiny images,'' 2009.

\bibitem{deng2009imagenet}
J.~Deng, W.~Dong, R.~Socher, L.-J. Li, K.~Li, and L.~Fei-Fei, ``Imagenet: A
  large-scale hierarchical image database,'' in \emph{2009 IEEE conference on
  computer vision and pattern recognition}.\hskip 1em plus 0.5em minus
  0.4em\relax Ieee, 2009, pp. 248--255.

\bibitem{HelocDataset}
FICO, ``Home equity line of credit \text{(HELOC)} dataset,''
  \url{https://community.fico.com/s/explainable-machine-learning-challenge},
  2019 (accessed June 15, 2022).

\bibitem{baldi2014searching}
P.~Baldi, P.~Sadowski, and D.~Whiteson, ``Searching for exotic particles in
  high-energy physics with deep learning,'' \emph{Nature communications},
  vol.~5, no.~1, pp. 1--9, 2014.

\bibitem{mooney1997monte}
C.~Z. Mooney, \emph{Monte carlo simulation}.\hskip 1em plus 0.5em minus
  0.4em\relax Sage, 1997, no. 116.

\bibitem{pace1997sparse}
R.~K. Pace and R.~Barry, ``Sparse spatial autoregressions,'' \emph{Statistics
  \& Probability Letters}, vol.~33, no.~3, pp. 291--297, 1997.

\bibitem{breiman2017classification}
L.~Breiman, J.~H. Friedman, R.~A. Olshen, and C.~J. Stone, \emph{Classification
  and regression trees}.\hskip 1em plus 0.5em minus 0.4em\relax Routledge,
  2017.

\bibitem{breiman2001random}
L.~Breiman, ``Random forests,'' \emph{Machine learning}, vol.~45, no.~1, pp.
  5--32, 2001.

\bibitem{broelemann2018gradient}
K.~Broelemann and G.~Kasneci, ``A gradient-based split criterion for highly
  accurate and transparent model trees,'' in \emph{IJCAI}, 2019.

\bibitem{mcculloch1943logical}
W.~S. McCulloch and W.~Pitts, ``A logical calculus of the ideas immanent in
  nervous activity,'' \emph{The bulletin of mathematical biophysics}, vol.~5,
  no.~4, pp. 115--133, 1943.

\bibitem{icml2020_5085}
Y.~Yamada, O.~Lindenbaum, S.~Negahban, and Y.~Kluger, ``Feature selection using
  stochastic gates,'' in \emph{Proceedings of Machine Learning and Systems
  2020}, 2020, pp. 8952--8963.

\bibitem{agarwal2020neural}
R.~Agarwal, N.~Frosst, X.~Zhang, R.~Caruana, and G.~E. Hinton, ``Neural
  additive models: Interpretable machine learning with neural nets,''
  \emph{arXiv preprint arXiv:2004.13912}, 2020.

\bibitem{optuna_2019}
T.~Akiba, S.~Sano, T.~Yanase, T.~Ohta, and M.~Koyama, ``Optuna: A
  next-generation hyperparameter optimization framework,'' in \emph{Proceedings
  of the 25rd {ACM} {SIGKDD} International Conference on Knowledge Discovery
  and Data Mining}, 2019.

\bibitem{merkel2014docker}
D.~Merkel, ``Docker: lightweight linux containers for consistent development
  and deployment,'' \emph{Linux journal}, vol. 2014, no. 239, p.~2, 2014.

\bibitem{bojer2021kaggle}
C.~S. Bojer and J.~P. Meldgaard, ``Kaggle forecasting competitions: An
  overlooked learning opportunity,'' \emph{International Journal of
  Forecasting}, vol.~37, no.~2, pp. 587--603, 2021.

\bibitem{rong2022evaluating}
Y.~Rong, T.~Leemann, V.~Borisov, G.~Kasneci, and E.~Kasneci, ``A consistent and
  efficient evaluation strategy for feature attribution methods,'' in
  \emph{International Conference on Machine Learning}.\hskip 1em plus 0.5em
  minus 0.4em\relax PMLR, 2022.

\bibitem{tomsett2020sanity}
R.~Tomsett, D.~Harborne, S.~Chakraborty, P.~Gurram, and A.~Preece, ``Sanity
  checks for saliency metrics,'' in \emph{Proceedings of the AAAI Conference on
  Artificial Intelligence}, vol.~34, no.~04, 2020, pp. 6021--6029.

\bibitem{ntoutsi2020bias}
E.~Ntoutsi, P.~Fafalios, U.~Gadiraju, V.~Iosifidis, W.~Nejdl, M.-E. Vidal,
  S.~Ruggieri, F.~Turini, S.~Papadopoulos, E.~Krasanakis \emph{et~al.}, ``Bias
  in data-driven artificial intelligence systems—an introductory survey,''
  \emph{Wiley Interdisciplinary Reviews: Data Mining and Knowledge Discovery},
  vol.~10, no.~3, p. e1356, 2020.

\bibitem{domingos2000mining}
P.~Domingos and G.~Hulten, ``Mining high-speed data streams,'' in
  \emph{Proceedings of the sixth ACM SIGKDD international conference on
  Knowledge discovery and data mining}, 2000, pp. 71--80.

\bibitem{manapragada2018extremely}
C.~Manapragada, G.~I. Webb, and M.~Salehi, ``Extremely fast decision tree,'' in
  \emph{Proceedings of the 24th ACM SIGKDD International Conference on
  Knowledge Discovery \& Data Mining}, 2018, pp. 1953--1962.

\bibitem{balestriero2022effects}
R.~Balestriero, L.~Bottou, and Y.~LeCun, ``The effects of regularization and
  data augmentation are class dependent,'' \emph{arXiv preprint
  arXiv:2204.03632}, 2022.

\bibitem{duda2020training}
P.~Duda, M.~Jaworski, A.~Cader, and L.~Wang, ``On training deep neural networks
  using a streaming approach,'' \emph{Journal of Artificial Intelligence and
  Soft Computing Research}, vol.~10, 2020.

\bibitem{haug2020leveraging}
J.~Haug, M.~Pawelczyk, K.~Broelemann, and G.~Kasneci, ``Leveraging model
  inherent variable importance for stable online feature selection,'' in
  \emph{Proceedings of the 26th ACM SIGKDD International Conference on
  Knowledge Discovery \& Data Mining}, 2020, pp. 1478--1502.

\bibitem{haug2021learning}
J.~Haug and G.~Kasneci, ``Learning parameter distributions to detect concept
  drift in data streams,'' in \emph{2020 25th International Conference on
  Pattern Recognition (ICPR)}.\hskip 1em plus 0.5em minus 0.4em\relax IEEE,
  2021, pp. 9452--9459.

\bibitem{tan2018survey}
C.~Tan, F.~Sun, T.~Kong, W.~Zhang, C.~Yang, and C.~Liu, ``A survey on deep
  transfer learning,'' in \emph{International conference on artificial neural
  networks}.\hskip 1em plus 0.5em minus 0.4em\relax Springer, 2018, pp.
  270--279.

\bibitem{shorten2019surveyaugmentation}
C.~Shorten and T.~M. Khoshgoftaar, ``A survey on image data augmentation for
  deep learning,'' \emph{Journal of Big Data}, vol.~6, no.~1, pp. 1--48, 2019.

\bibitem{chawla2002smote}
N.~V. Chawla, K.~W. Bowyer, L.~O. Hall, and W.~P. Kegelmeyer, ``Smote:
  synthetic minority over-sampling technique,'' \emph{Journal of artificial
  intelligence research}, vol.~16, pp. 321--357, 2002.

\bibitem{jing2020self}
L.~Jing and Y.~Tian, ``Self-supervised visual feature learning with deep neural
  networks: A survey,'' \emph{IEEE transactions on pattern analysis and machine
  intelligence}, 2020.

\bibitem{liu2021self}
X.~Liu, F.~Zhang, Z.~Hou, L.~Mian, Z.~Wang, J.~Zhang, and J.~Tang,
  ``Self-supervised learning: Generative or contrastive,'' \emph{IEEE
  Transactions on Knowledge and Data Engineering}, 2021.

\bibitem{ucar2021subtab}
T.~Ucar, E.~Hajiramezanali, and L.~Edwards, ``Subtab: Subsetting features of
  tabular data for self-supervised representation learning,'' \emph{Advances in
  Neural Information Processing Systems}, vol.~34, 2021.

\end{thebibliography}
